\documentclass[sigconf]{acmart}
\AtBeginDocument{%
  }

\copyrightyear{2026}
\acmYear{2026}
\setcopyright{cc}
\setcctype{by-nc-nd}
\acmConference[MM '26] {Proceedings of the 34th ACM International Conference on Multimedia}{November 10--14, 2026}{Rio de Janeiro, Brazil.}
\acmBooktitle{Proceedings of the 34th ACM International Conference on Multimedia (MM '26), November 10--14, 2026, Rio de Janeiro, Brazil}
\acmISBN{979-8-4007-2213-4/2026/11}
\acmDOI{10.1145/3767308.3835084}
\usepackage{xspace}

\newcommand{\eg}{\textit{e}.\textit{g}.\xspace}
\usepackage{bm}
\usepackage{multirow}
\usepackage[table]{xcolor}
\usepackage{enumitem}
\usepackage{algorithm}
\usepackage{algpseudocode}
\usepackage{colortbl}
\usepackage{arydshln}
\usepackage{stfloats}
\usepackage{placeins}
\usepackage{balance}

\begin{document}

\title[Object-Dependent Concept Brittleness in Diffusion Models]
{Interpreting Object-Dependent Concept Brittleness in Text-to-Image Diffusion Models}

\author{Yifan Yuan}
\authornote{Yifan Yuan and Xiangyu Liu contributed equally to this work.}
\affiliation{%
  \department{School of Artificial Intelligence}
  \institution{Shenzhen University}
  \city{Shenzhen}
  \country{China}}
  \email{yifanyuan@szu.edu.cn}

\author{Xiangyu Liu}
\authornotemark[1]
\affiliation{%
  \department{School of Mathematical Sciences}
  \institution{Shenzhen University}
  \city{Shenzhen}
  \country{China}
}
\email{metecade@gmail.com}

\author{Hongming Shan}
\affiliation{%
\department{Institute of Science and Technology for Brain-inspired Intelligence}
 \institution{Fudan University}
 \city{Shanghai}
 \country{China}}
 \email{hmshan@fudan.edu.cn}

\author{Yu Han}
\affiliation{%
\department{College of Computer Science and Software Engineering}
  \institution{Shenzhen University}
  \city{Shenzhen}
  \country{China}}
\email{2024040042@mails.szu.edu.cn}

\author{Yu Jiang}
\affiliation{%
  \department{Department of Statistics and Data Science}
  \institution{National University of Singapore}
  \city{Singapore}
  \country{Singapore}}
\email{e1348927@u.nus.edu}
  
\author{Hao Tan}
\affiliation{%
 \department{School of Artificial Intelligence}
  \institution{Shenzhen University}
  \city{Shenzhen}
  \country{China}}
\email{2500672006@mails.szu.edu.cn}

\author{Junping Zhang}
\affiliation{%
  \department{College of Computer Science and Artificial Intelligence}
  \institution{Fudan University}
  \city{Shanghai}
  \country{China}}
\email{jpzhang@fudan.edu.cn}

\author{Linlin Shen}
\correspondingauthor
\affiliation{%
  \department{School of Artificial Intelligence}
  \institution{Shenzhen University}
  \city{Shenzhen}
  \country{China}}
\email{llshen@szu.edu.cn}

\renewcommand{\shortauthors}{Yifan Yuan et al.}
\newcommand{\shan}[1]{\textcolor{blue}{#1}}

\begin{abstract}
Although text-to-image diffusion models generally exhibit strong
prompt-following ability, we identify a persistent and previously
underexplored failure pattern in which a small subset of prompts
differing only in the \textit{object} consistently fails to realize the same
target \textit{concept} under identical generation settings. We term this
phenomenon \textbf{object-dependent concept brittleness}.
Such cases suggest systematic internal blind spots rather than random
sampling noise.
In this paper, we present an interpretability-oriented framework to audit
and minimally correct these failures.
Our key idea is to analyze denoising trajectories in a step-wise sparse
autoencoder (SAE) space, where abstract style and attribute concepts
become more separable than in the raw denoising representation.
This sparse space enables us to compare successful and failed generations,
identify concept dimensions whose evidence is missing, weakened, or
temporally delayed, and construct class-level concept prototypes from
reliable class-consistent samples.
Based on this audit process, we introduce a lightweight inference-time
correction strategy that interpolates denoising features toward the
corresponding prototype in SAE space.
Rather than serving as a task-specific retraining method, this intervention
acts as a validation of the diagnosed concept deficiency.
We evaluate the proposed framework on style and attribute failure cases
across multiple diffusion backbones, with significant improvements in
concept consistency, text fidelity, and repair success.
Further analyses show that deeper denoising representations provide clearer
concept structure, while early-stage intervention offers the strongest
correction leverage.
Code is available at
\href{https://github.com/Metecade/Object-Dependent-Concept-Brittleness}
{\texttt{GitHub}}.
\end{abstract}

\ccsdesc[500]{Computing methodologies~Image representations}
\ccsdesc[500]{Computing methodologies~Computer vision}


\keywords{Diffusion models; Interpretability; Sparse autoencoders; Concept brittleness; Concept-level failure diagnosis}


\settopmatter{authorsperrow=4}
\maketitle

\begin{figure*}[t]
\centering
  \includegraphics[width=0.75\textwidth]{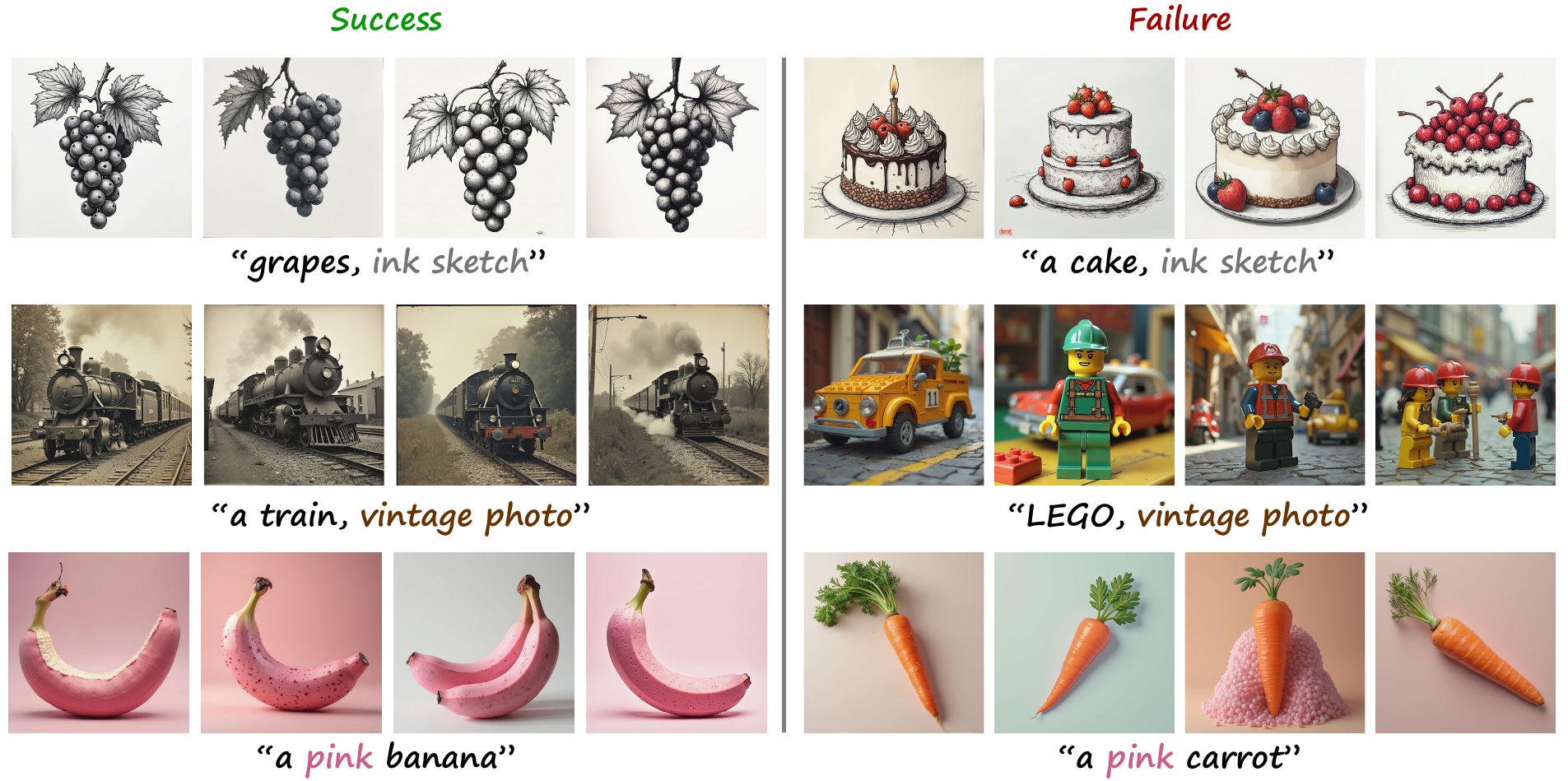}
  \caption{\textbf{Object-Dependent Concept Brittleness} discovered in this work. For a fixed concept, Flux produces correct results for most objects but repeatedly fails on a subset of semantically related objects. Each object is generated with four random seeds.}
  \Description{Examples of successful and failed generations under fixed target concepts. Grapes, a train, and a banana follow the requested concepts, while a cake, LEGO figures, and a carrot fail under the same settings.}
  \label{fig:phe}
\end{figure*}

\section{Introduction}

Text-to-image diffusion models (DMs)~\cite{nichol2022glide,saharia2022photorealistic,betker2023improving} show strong prompt-following ability under diverse style, attribute, and compositional conditions~\cite{cao2025repldm}. Yet we observe a persistent failure pattern that remains underexplored, which we term \textbf{object-dependent concept brittleness}: for a fixed target \textit{concept} (\eg, style, texture, or attribute), the model succeeds on most prompts but repeatedly fails on a small subset of near-identical prompts that differ only in the \textit{object}. Unlike generic prompt-following failure~\cite{ghosh2023geneval,huang2023t2icompbench}, this phenomenon is not caused by prompt complexity or missing concept knowledge; rather, it indicates that the same concept is realized unevenly across objects. For example, under the same setting, Flux~\cite{blackforestlabs2024flux1} can render ``grapes'' in ink sketch style but revert to a colorful style for ``cake''. Similar failures also appear in color, material, texture, and attribute binding~\cite{ghosh2023geneval,huang2023t2icompbench,zhang2025rbind}. Their persistence across multiple random seeds suggests a systematic weakness in how diffusion models internalize abstract concepts across different contents.

These failures are informative because they may expose the boundary of what the model has actually learned: if a concept is realized correctly for most objects but repeatedly breaks on a few nearby cases, then it has likely not been learned in a sufficiently robust and compositional way~\cite{ghosh2023geneval,huang2023t2icompbench}. Diagnosing this phenomenon is difficult because it is \textbf{sparse}, often \textbf{global rather than local}, and fundamentally \textbf{process-dependent}: concept failure may not be visible in attention maps alone and depends on how supporting evidence evolves during denoising~\cite{tang2023daam,hertz2023prompt,chefer2023attend,brack2023sega}.

Prior work related to this problem mainly falls into three directions. Attention attribution and spatial localization methods~\cite{tang2023daam,hertz2023prompt} are effective for local token-region correspondence, but less suited to global failures such as style mismatch or abstract attribute loss. Temporal analyses~\cite{gorgun2026temporal,kwon2023semanticlatent,hertz2023prompt} study when concepts are editable, but usually capture generic stage-wise controllability rather than the difference between successful and failed cases. Studies of diffusion dynamics and representation structure~\cite{chen2024trajectory,kwon2023semanticlatent,hahm2024isometric,yuan2025dissecting} provide broader insights, yet remain too coarse for selective instance-level failure. Overall, existing methods still lack a representation-level framework for content-dependent failure of the same concept.

The above limitations suggest that endpoint inspection or local attention analysis alone is not sufficient for this problem. In raw denoising features, content, style, texture, and attributes are highly entangled, making concept-level diagnosis difficult. We therefore seek a space that preserves denoising information while making concept evidence more separable. Recent work~\cite{gao2025scaling,cywinski2025saeuron,surkov2025onestep} suggests that \textbf{sparse autoencoders (SAEs)} can provide such a representation. Motivated by this, we model denoising features with step-wise SAEs. Our key observation is that, in this timestep-wise sparse space, successful and failed generations become substantially more separable than in raw features, and failed cases show systematic activation differences that are otherwise hard to observe.

Based on this observation, we propose an interpretability-oriented framework for diagnosing and correcting object-dependent concept brittleness in text-to-image diffusion models. We train SAEs at each denoising timestep, compare successful and failed samples in sparse space, construct class-level concept prototypes from reliable class-consistent generations, and use them as \textbf{internal priors} during inference. We then perform lightweight correction by interpolating the current denoising representation toward the prototype in SAE space, without retraining the base diffusion model~\cite{hertz2023prompt,chefer2023attend,brack2023sega}. This intervention serves not only as a plug-and-play utility but also as a \textbf{probe of representational relevance}: if restoring the diagnosed evidence changes the outcome, then the sparse discrepancy is likely mechanistically relevant.  Experiments on style and attribute settings across five diffusion backbones (SD 1.5~\cite{rombach2022latent}, SD 3.5~\cite{stabilityai2024sd35}, SDXL~\cite{podell2023sdxl}, PixArt~\cite{chen2024pixartalpha}, Flux~\cite{blackforestlabs2024flux1}) show consistent improvements in concept consistency and repair success.

Our analysis further reveals two main findings about \textit{where} concept evidence is most legible and \textit{when} it is most correctable. First, deeper denoising features form a clearer sparse concept space and support more effective intervention than earlier representations. Second, correction is strongly \textbf{time-sensitive}: small sparse adjustments in \textbf{early} timesteps can substantially change the final output, whereas later intervention is much weaker, and prolonged strong intervention may damage unrelated content. Together, these results show that tracing systematic failures requires identifying not only which concept evidence is missing, but also where it becomes readable and when it can still be repaired.

In summary, our contributions are as follows:
\begin{itemize}[leftmargin=*, nosep]
    \item \textbf{We define and systematically characterize object-dependent concept brittleness in text-to-image diffusion models.}
       We identify a previously underexplored failure pattern in which the same target concept is realized reliably for most content instances but repeatedly fails on a small subset of near-identical prompts that differ only in content.
    
    \item \textbf{We uncover an interpretable representation pattern underlying concept brittleness.}
       Using step-wise SAEs, we show that successful and failed generations exhibit stable and separable activation differences in sparse representation space, revealing an internal structure that is difficult to observe in raw features.
    
    
    \item \textbf{We validate the framework on two tasks and five widely used diffusion backbones.}
       Across both style and attribute settings, our approach consistently improves concept consistency and repair success, suggesting that the discovered phenomenon and its sparse representation patterns generalize across common diffusion model families (SD 1.5, SD 3.5, SDXL, PixArt, Flux).
\end{itemize}

\section{Related Work}

\subsection{Interpreting Diffusion Model Internals}

Recent work on diffusion interpretability mainly spans three directions. The first uses attention and attribution to localize how textual concepts are expressed in generated images, as exemplified by DAAM~\cite{tang2023daam}, ConceptAttention~\cite{helbling2025conceptattention}, and I2AM~\cite{park2025i2am}. The second moves from localization to mechanism analysis by examining concept-relevant heads or components; representative studies show that cross-attention head patterns can align with human visual concepts and that internal modules can be decomposed into positive and negative contributions to concept expression~\cite{park2025headrelevance,nguyen2025cad}. The third studies concept dynamics along denoising trajectories~\cite{lei2024ddpath}, asking when concepts emerge, stabilize, or remain editable, as in PCI~\cite{gorgun2026temporal}. Collectively, these works reveal interpretable structure across the spatial, modular, and temporal dimensions of diffusion models. Our work builds on this perspective, but targets a different question: why a small subset of near-neighbor prompts systematically fails, and how such failures can be diagnosed as missing or weakened concept evidence within the denoising process.

\subsection{Sparse Autoencoders for Mechanistic Interpretability}

Sparse autoencoders (SAEs) have become an important tool for extracting sparse and semantically meaningful features from dense neural activations~\cite{bricken2023monosemanticity,templeton2024scaling,gao2025scaling}. Recent work has further extended SAE-based analysis to diffusion models~\cite{cywinski2025saeuron,tinaz2025emergence}. For example, SAeUron~\cite{cywinski2025saeuron} uses SAE features for concept-level unlearning, while Tinaz et al.~\cite{tinaz2025emergence} studies concept evolution and stage-dependent steering. These studies establish SAE spaces as useful tools for analyzing and intervening on generative model internals. In contrast, we use step-wise SAEs as a diagnostic representation for comparing successful and failed denoising trajectories. Rather than removing or generically steering concepts, we identify missing or temporally misaligned concept evidence in systematic failures, and test these diagnoses through lightweight prototype-guided correction.

\section{Preliminaries}
\label{sec:pre}
\subsection{Unified Denoising Formulation}
\label{ssec:denoising}
Let $z_t$ denote the sampler latent state at timestep $t$, and let $p$ denote the text prompt. A broad class of diffusion and flow-based generators can be written in the unified form~\cite{ho2020ddpm,song2021ddim,lipman2023flow,liu2023rectifiedflow}:
\begin{equation}
    z_{t-1} = \mathcal U_t\big(z_t,\Phi_\theta(z_t,p,t)\big),
\end{equation}
Here, $\mathcal{U}_t(\cdot)$ denotes the sampler update, and $\Phi_\theta(z_t,p,t)$ denotes the backbone prediction at timestep $t$. For noise-prediction diffusion models~\cite{ho2020ddpm,song2021ddim}, $\Phi_\theta(\cdot)$ is the predicted noise $\epsilon_\theta(z_t,p,t)$. A typical deterministic update can be expressed through
\begin{equation} 
    \hat x_0(z_t,p,t)= \frac{z_t-\sqrt{1-\bar\alpha_t}\,\epsilon_\theta(z_t,p,t)} {\sqrt{\bar\alpha_t}},
\end{equation}
followed by
\begin{equation}
z_{t-1} = \sqrt{\bar\alpha_{t-1}}\,\hat x_0 + \sqrt{1-\bar\alpha_{t-1}}\,\epsilon_\theta(z_t,p,t).
\end{equation}

For flow- or velocity-prediction models~\cite{lipman2023flow,liu2023rectifiedflow}, $\Phi_\theta(\cdot)$ becomes a vector field $v_\theta(z_t,p,t)$, and the update can be written as:
\begin{equation}
    z_{t-\Delta t}=z_t-\Delta t\cdot v_\theta(z_t,p,t).
\end{equation}
Rather than directly analyzing $z_t$, our method operates on the intermediate backbone prediction produced at each timestep. 

\subsection{Sparse Autoencoders}
\label{ssec:sae}
Sparse autoencoders (SAEs)~\cite{makhzani2014ksparse,bricken2023monosemanticity,gao2025scaling} are reconstruction models that map dense activations into sparse latent representations, and are widely used to recover more interpretable internal structure from neural features. In our setting, SAEs serve as a sparse representation tool for denoising features, making condition-related structure easier to analyze than in the original dense feature space.

Given a normalized token feature $x\in\mathbb{R}^{d}$, a standard single-layer ReLU sparse autoencoder is defined as:
\begin{equation}
    z = \operatorname{ReLU}\!\big(W_{\mathrm{enc}}(x-b_{\mathrm{pre}})+b_{\mathrm{enc}}\big),
\end{equation}
\begin{equation}
    \hat{x} = W_{\mathrm{dec}} z + b_{\mathrm{pre}},
\end{equation}
where $W_{\mathrm{enc}}\in\mathbb{R}^{K_d\times d}$ and $W_{\mathrm{dec}}\in\mathbb{R}^{d\times K_d}$ are the encoder and decoder weight matrices respectively, $b_{\mathrm{pre}}\in\mathbb{R}^{d}$ and $b_{\mathrm{enc}}\in\mathbb{R}^{K_d}$ are learnable bias terms, and $z$ denotes the latent feature activations. $K_d$ is the latent dimension of the autoencoder. Typically, $K_d$ is equal to $d$ multiplied by a positive expansion factor.

The SAE is trained to reconstruct the input while encouraging sparsity in the latent space:
\begin{equation}
    \mathcal{L}(x)=\|x-\hat{x}\|_2^2+\lambda \mathcal{L}_{\mathrm{aux}},
\end{equation}
where $\|x-\hat{x}\|_2^2$ is the reconstruction loss and $\mathcal{L}_{\mathrm{aux}}$ is an auxiliary term used to discourage inactive latent units.

In this work, we adopt the Top-K variant of SAE~\cite{makhzani2014ksparse,gao2025scaling}, which retains only the $K$ largest latent activations for each input and sets the rest to zero. The encoder is therefore written as:
\begin{equation}
    z = \operatorname{TopK}\!\big(W_{\mathrm{enc}}(x-b_{\mathrm{pre}})\big),
\end{equation}
where $\operatorname{TopK}(\cdot)$ keeps the $K$ largest entries and zeroes out the rest, with $K \ll K_d$. The decoder remains unchanged.


\section{Method}


\subsection{Problem Setup}
We study \textbf{object-dependent concept brittleness} in the following setting. Let a text prompt be composed of an \emph{object} $o$ and a target \emph{concept} $c$, where $c$ may denote a style, an attribute, or another abstract concept to be realized in generation. Although a text-to-image diffusion model can satisfy $c$ for most prompts that share the same concept, it may repeatedly fail on a small subset of prompts whose objects differ but remain semantically close to the successful ones. Our goal is to diagnose this object-dependent failure at the level of internal denoising representations, and to test whether restoring the missing concept-related evidence can repair the failure without retraining the base model. To this end, we assume access to a reference set:
\begin{equation}
\mathcal{D}_{\mathrm{ref}}=\{(x_i,p_i,c_i)\}_{i=1}^{N},
\end{equation}
where $x_i$ denotes a reference sample, $p_i$ is its text prompt, and $c_i$ is the corresponding target concept class. For each sample $i$ and timestep $t$, we extract the update quantity of the current denoising step and tokenize it as $X_i^{(t)}\in\mathbb{R}^{P\times D}$, where $P$ is the number of tokens and $D$ is the backbone-dependent feature dimension. From these features, we obtain token-level sparse codes $Z_i^{(t)}$ and pooled sample-level embeddings $s_i^{(t)}$. The reference set is used to estimate class-consistent structure in sparse space and to construct class-conditioned sparse priors for later intervention.

\subsection{Why a Sparse Internal Space is Needed}

The central challenge in diagnosing object-dependent concept brittleness is not merely to identify whether a target concept is missing in the final output, but to determine how successful and failed cases differ \emph{inside} the denoising process. If object-dependent concept failure reflects a systematic weakness in concept realization rather than incidental sampling noise, then the relevant discrepancy should be traceable within the denoising process. 

\begin{figure}[ht]
\centering
\begin{minipage}[t]{0.48\linewidth}
    \centering
    \includegraphics[width=\linewidth]{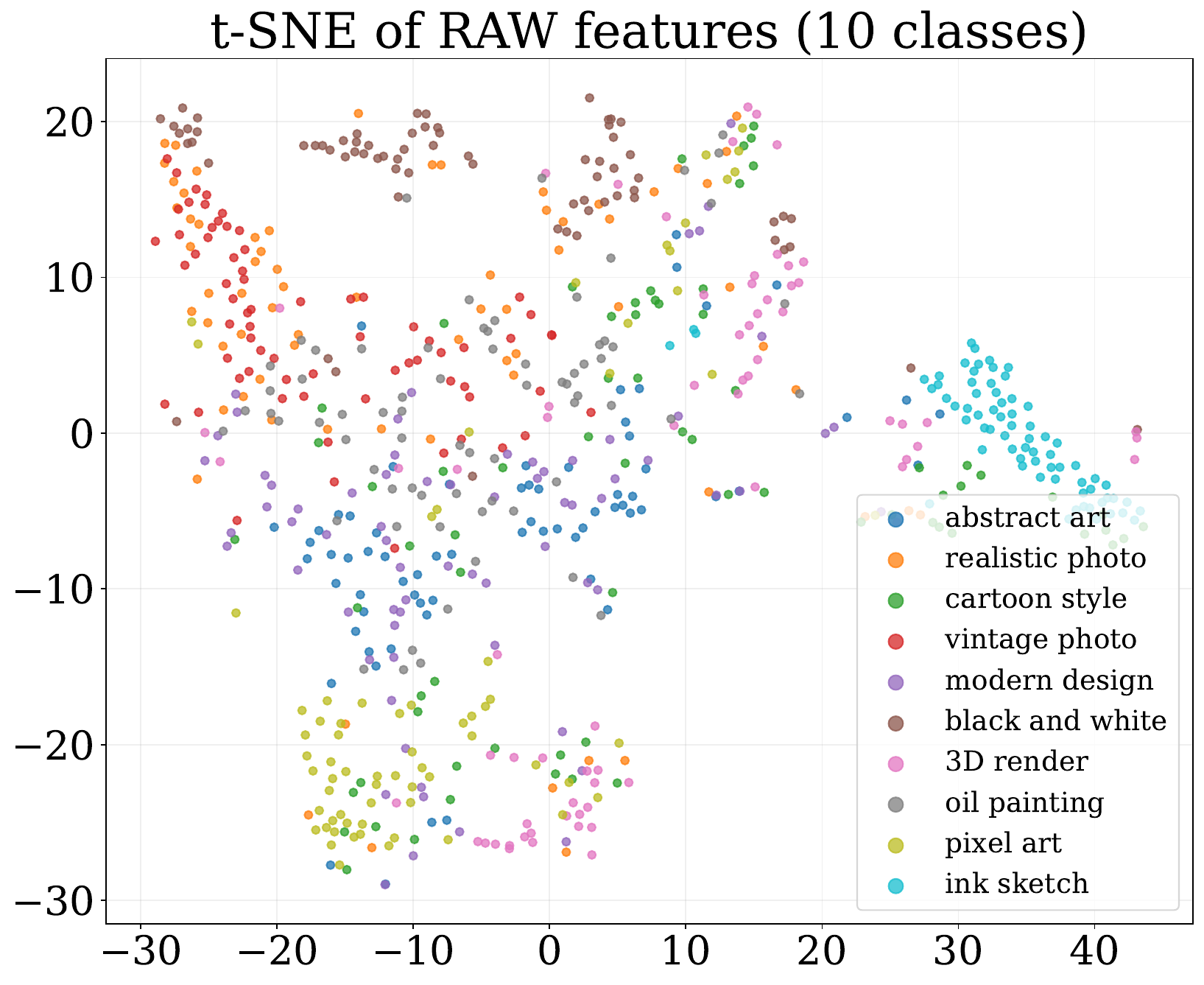}
    
    {\small (a) Noise latent features}
\end{minipage}
\hfill
\begin{minipage}[t]{0.48\linewidth}
    \centering
    \includegraphics[width=\linewidth]{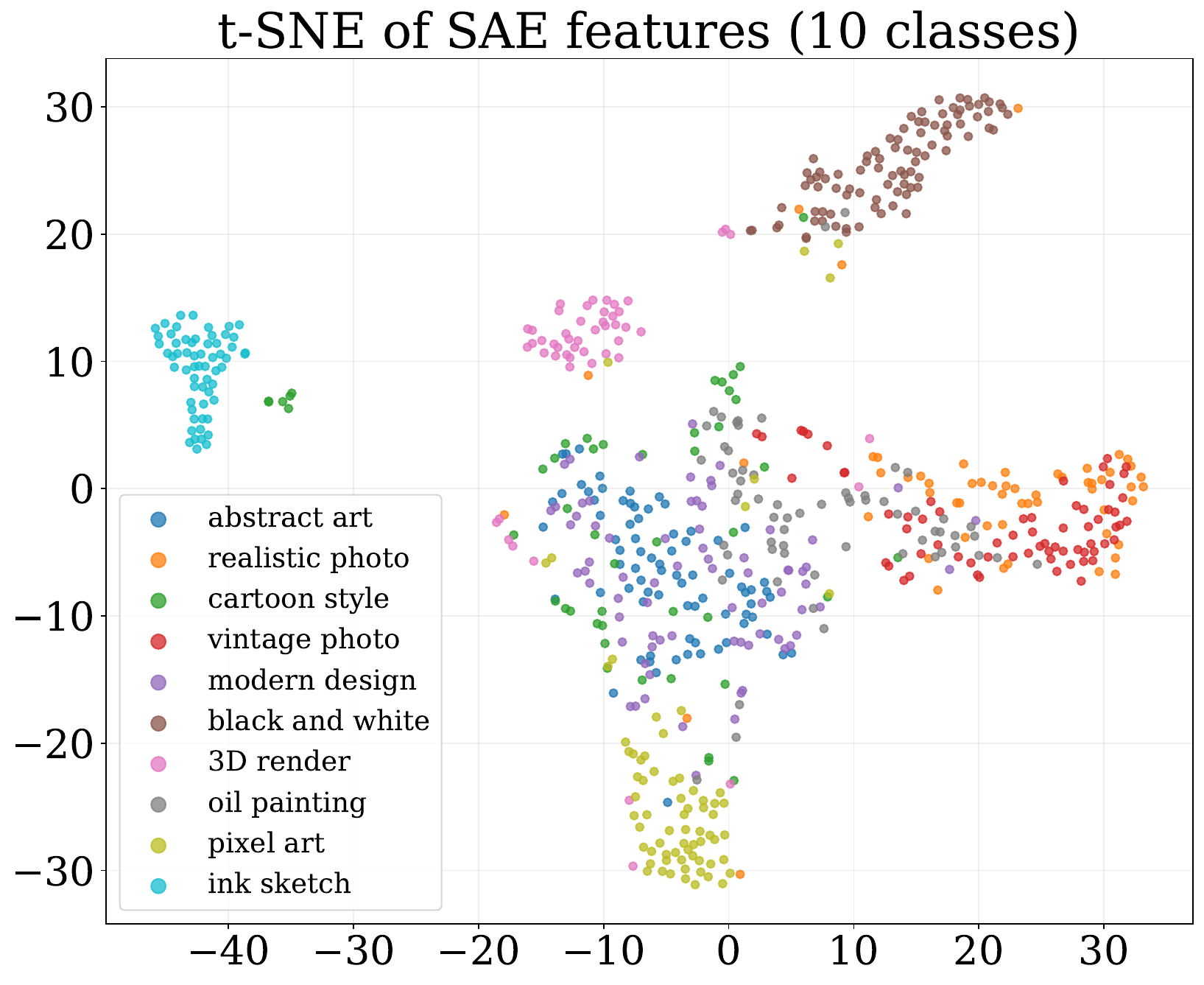}
    
    {\small (b) SAE latent features}
\end{minipage}
\vspace{-8pt}
\caption{T-SNE visualization of raw denoising features and corresponding SAE embeddings for 10 style classes.}
\Description{Two t-SNE scatter plots comparing raw noise-latent features with SAE-latent features across ten style classes. The SAE representation forms more distinct class clusters.}
\label{fig:tsne_comparison}
\vspace{-8pt}
\end{figure}

A first indication comes from the original denoising feature space itself. As shown in Fig.~\ref{fig:tsne_comparison}~(a), when we visualize the pooled raw denoising features of ten style classes with t-SNE, substantial overlap remains across classes. This suggests that the original denoising representation is \textbf{highly entangled}: at each timestep, the backbone prediction simultaneously carries information about content, style, texture, and attributes, and these factors are mixed across spatial tokens and channels. As a result, samples that differ in whether the target concept is successfully realized may still appear close in the raw feature space, while the relevant discrepancy is confined to a small subset of latent directions. This makes it difficult to isolate the internal components associated with successful concept realization from those associated with failure.  




The overlap in Fig.~\ref{fig:tsne_comparison}~(a) also shows that the raw feature space does not provide a stable basis for \textbf{cross-sample comparison}. It is therefore poorly suited to estimating reliable class structure or determining whether a failed sample truly departs from a concept-consistent internal pattern. This motivates the use of a sparse internal space. An appropriate representation should preserve the generative information of the denoising process while making concept-related structure more separable across samples. Sparse autoencoders are well suited to this role, as they transform dense activations into sparse latent codes while approximately preserving the underlying information. In the resulting space, concept-related evidence is concentrated on a smaller set of active dimensions, making deviations from successful patterns easier to detect. 

Based on this motivation, we perform the subsequent analysis in timestep-specific SAE spaces. As shown in Fig.~\ref{fig:tsne_comparison}~(b), after projection into SAE space, the class structure becomes substantially clearer, suggesting that the learned sparse representation is more suitable for subsequent diagnosis than the original dense feature space.



\subsection{Step-wise SAE Representation}

To compare denoising trajectories under the same target condition, we represent the step-update features in a timestep-specific sparse space. Given $X_i^{(t)}\in\mathbb{R}^{P\times D}$, we normalize it using the timestep-specific statistics $(\mu^{(t)},\sigma^{(t)})$ and encode it with a dedicated SAE:
\begin{equation}
Z_i^{(t)}
=
\mathrm{Enc}^{(t)}\!\left(
\frac{X_i^{(t)}-\mu^{(t)}}{\sigma^{(t)}+\varepsilon}
\right),
\end{equation}
where $Z_i^{(t)}\in\mathbb{R}^{P\times K_d}$ is the token-level sparse code and $\varepsilon$ ensures numerical stability. We further obtain the pooled representation
$s_i^{(t)}=\frac{1}{P}\sum_{p=1}^{P}Z_i^{(t)}(p)\in\mathbb{R}^{K_d}$.

We use a separate SAE for \textbf{each timestep} encoder across the full denoising trajectory. This is because denoising features vary substantially across timesteps in both distribution and semantic role: early steps mainly encode coarse global structure, whereas later steps contain more refined condition-specific information. A shared SAE would mix these heterogeneous representations into a single latent space and blur temporally specific structure. In contrast, step-wise SAEs preserve the local statistics of each denoising stage and provide a more suitable basis for cross-sample comparison.

Each SAE is trained with the reconstruction objective in Sec.~\ref{ssec:sae}, combining reconstruction fidelity with sparsity regularization. As a result, the learned sparse space remains faithful to the original denoising features while making condition-related structure more separable. This representation serves as the basis for the subsequent analysis: instead of comparing samples in the raw feature space, we compare them in a timestep-specific sparse space where deviations from successful trajectories can be identified more directly.

In the remainder of the method, we use the step-wise sparse codes $Z_i^{(t)}$ and pooled embeddings $s_i^{(t)}$ as the basic representations for diagnosis and intervention. The former retains token-level structure for later correction, while the latter supports sample-level comparison and class-wise statistics.

\subsection{Diagnosing Concept Brittleness in SAE Space}
\label{ssec:diag}

With the step-wise SAE representation in place, we diagnose concept brittleness by measuring how individual samples depart from the class-consistent sparse structure under the same target condition. The key hypothesis is that, if concept brittleness reflects a systematic failure of internal condition realization rather than incidental sampling noise, then failed generations should exhibit structured departures from the reference profile in SAE space.

\begin{figure}[ht]
\centering

\begin{minipage}[t]{0.8\linewidth}
    \centering
    \includegraphics[width=\linewidth]{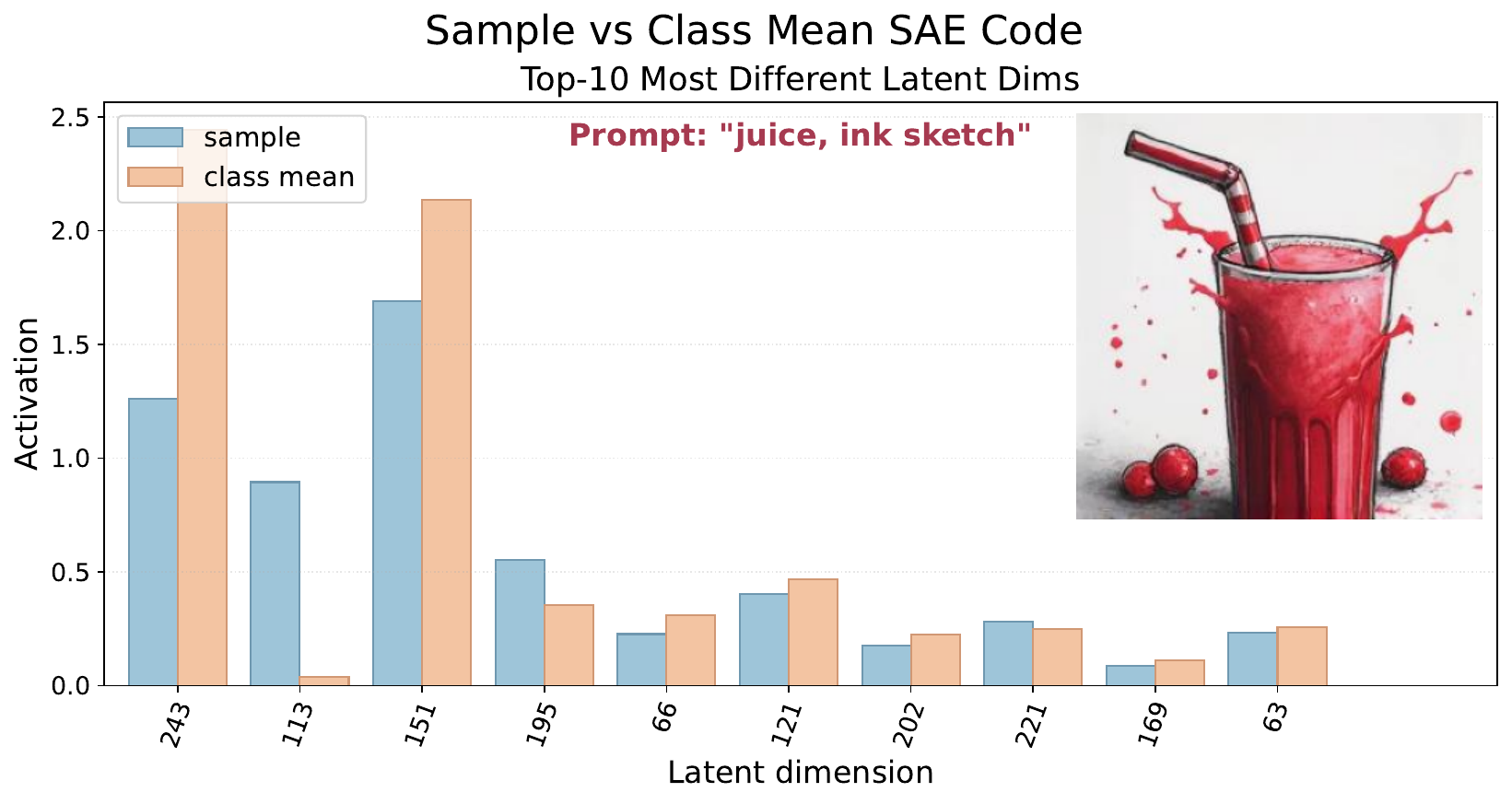}\\[-0.2em]
    {\small (a) Failed sample}
\end{minipage}


\begin{minipage}[t]{0.8\linewidth}
    \centering
    \includegraphics[width=\linewidth]{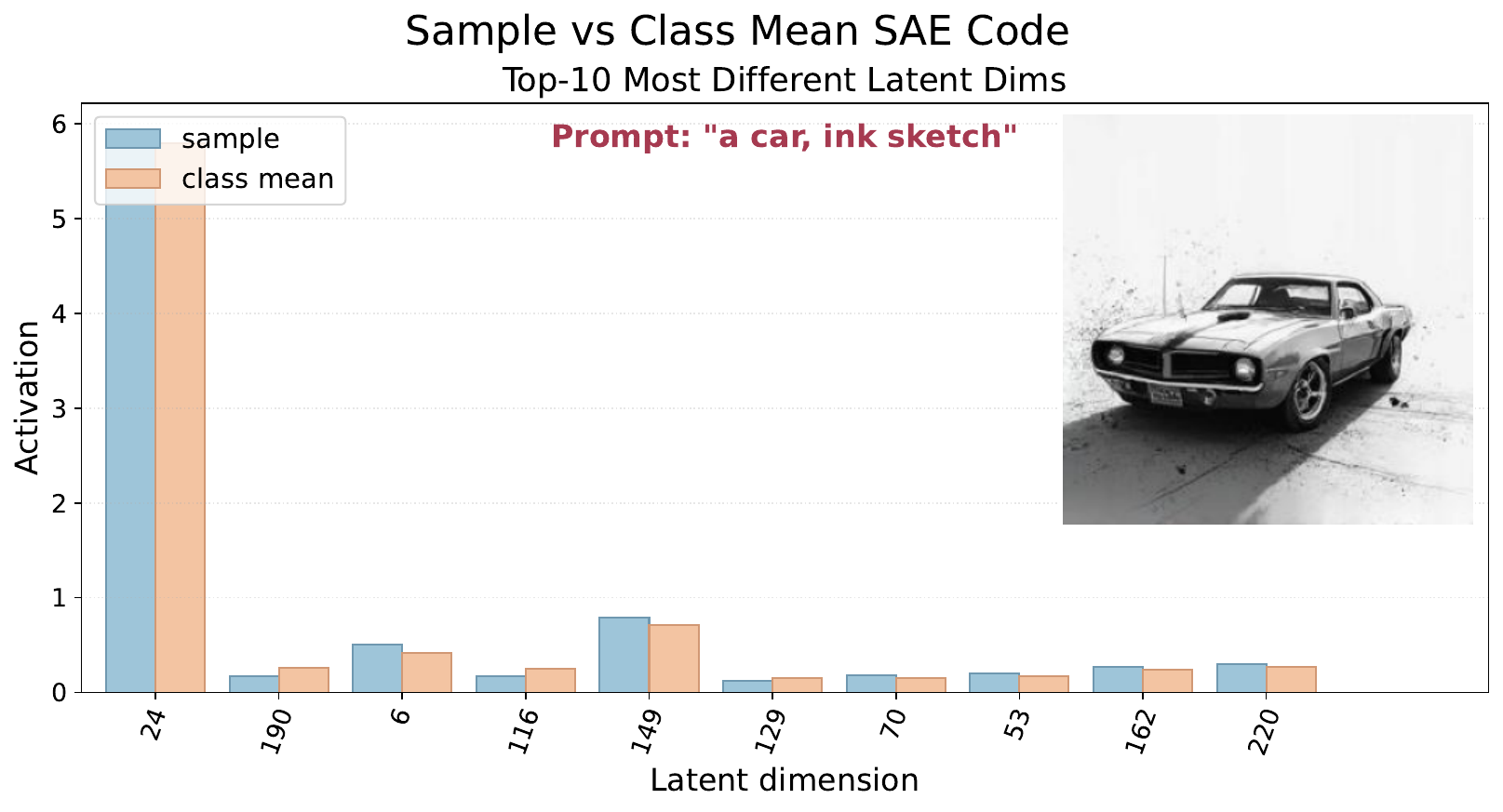}\\[-0.2em]
    {\small (b) Successful sample}
\end{minipage}

\vspace{-0.3em}
\caption{Sample-level SAE activations versus the class mean on the top-10 most deviating latent dimensions. Failed samples (top) exhibit larger sparse-space deviations than successful samples  (bottom).}
\Description{Two bar charts compare a sample's SAE activations with its class mean on the ten most different latent dimensions. The failed sample differs substantially more than the successful sample.}
\label{fig:bad_good}
\vspace{-8pt}
\end{figure}

Let $\mathcal{I}_c=\{i\mid c_i=c\}$ denote the set of reference samples associated with target condition $c$. We first estimate the class-wise center and dispersion in pooled SAE space:
\begin{equation}
\mu_c^{(t)}=
\frac{1}{|\mathcal{I}_c|}
\sum_{i\in\mathcal{I}_c}s_i^{(t)},
\qquad
\sigma_c^{(t)}=
\sqrt{
\frac{1}{|\mathcal{I}_c|}
\sum_{i\in\mathcal{I}_c}
\left(s_i^{(t)}-\mu_c^{(t)}\right)^2
}.
\end{equation}
These statistics define a class-consistent reference profile at timestep $t$. We then quantify how far each sample deviates from this profile using the normalized max-deviation score
\begin{equation}
d_i^{(t,c)}=
\left\|
\frac{s_i^{(t)}-\mu_c^{(t)}}
{\sigma_c^{(t)}+\varepsilon}
\right\|_{\infty}.
\end{equation}
This score measures whether a sample remains consistent with the sparse activation structure of its reference class. As illustrated in Fig.~\ref{fig:bad_good}, visually failed cases tend to exhibit larger deviations from the class mean than successful cases, and the dominant discrepancies are concentrated on a small subset of sparse dimensions rather than distributed uniformly across the representation.

\FloatBarrier

To obtain a reliable class-conditioned reference for later intervention, we retain the prototype-consistent subset
\begin{equation}
\label{eq:tau}
\mathcal{I}_{c,\mathrm{keep}}^{(t)}=
\{\,i\in\mathcal{I}_c\mid d_i^{(t,c)}\le\tau\,\},
\end{equation}
where $\tau$ is a filtering threshold. Samples satisfying this criterion are treated as reliable samples for prototype estimation. This filtering step reduces the influence of atypical or weakly expressed class members and yields a more stable reference subset.

\begin{figure}[ht]
\centering
\includegraphics[width=1.0\linewidth]{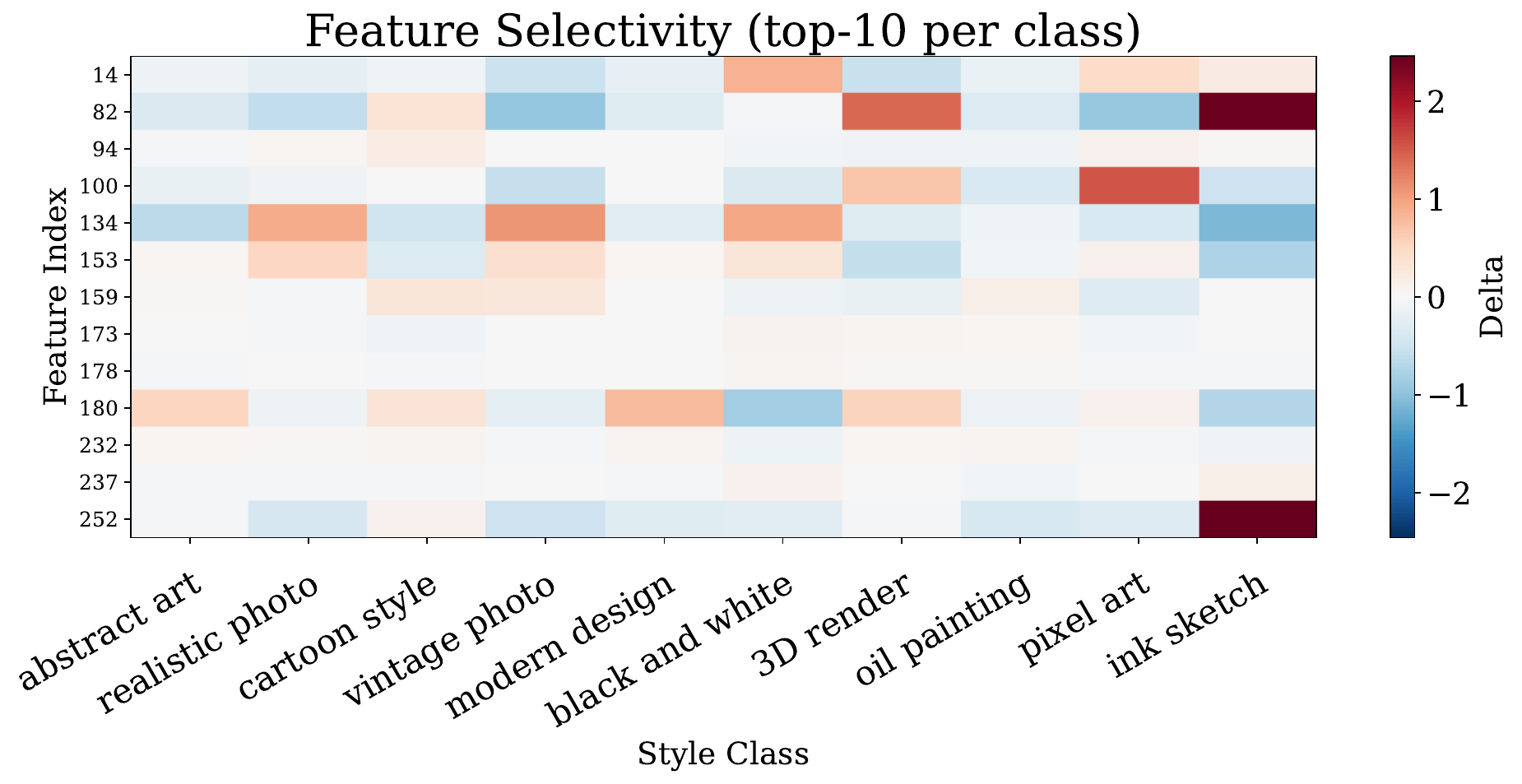}
\vspace{-16pt}
\caption{Class-selective SAE activations after filtering prototype-inconsistent samples. Different style classes activate distinct subsets of sparse dimensions.}
\Description{A heat map of the ten most selective SAE dimensions per style class, showing distinct positive and negative activation patterns across ten styles.}
\label{fig:selectivity}
\vspace{-8pt}
\end{figure}

To make the resulting structure more explicit, we further examine which latent dimensions are most discriminative for each condition class after filtering prototype-inconsistent samples. Figure~\ref{fig:selectivity} shows that these class-selective activations form a structured pattern across conditions: different style classes are associated with distinct subsets of sparse dimensions, and the dominant differences are concentrated on these dimensions rather than spread uniformly over the full code. This observation strengthens the interpretation of concept brittleness as a structured representational mismatch rather than a random perturbation in the denoising feature space.

Once the reliable subset is identified, we turn from sample-level diagnosis to class-level structure. Since pooled embeddings are suitable for filtering but discard token-level information, we construct the final class-conditioned sparse prototype at the token level:
\begin{equation}
R_c^{(t)}(p)=
\frac{1}{|\mathcal{I}_{c,\mathrm{keep}}^{(t)}|}
\sum_{i\in\mathcal{I}_{c,\mathrm{keep}}^{(t)}} Z_i^{(t)}(p),
\qquad p=1,\dots,P,
\end{equation}
which yields $R_c^{(t)}\in\mathbb{R}^{P\times K_d}$. 
Here, $p$ indexes token positions, $\mu_c^{(t)}$ is used as the pooled class center for diagnosis and filtering, and $R_c^{(t)}$ serves as a token-level sparse prototype for subsequent correction.

\begin{figure*}[t]
\centering
\includegraphics[width=0.86\linewidth]{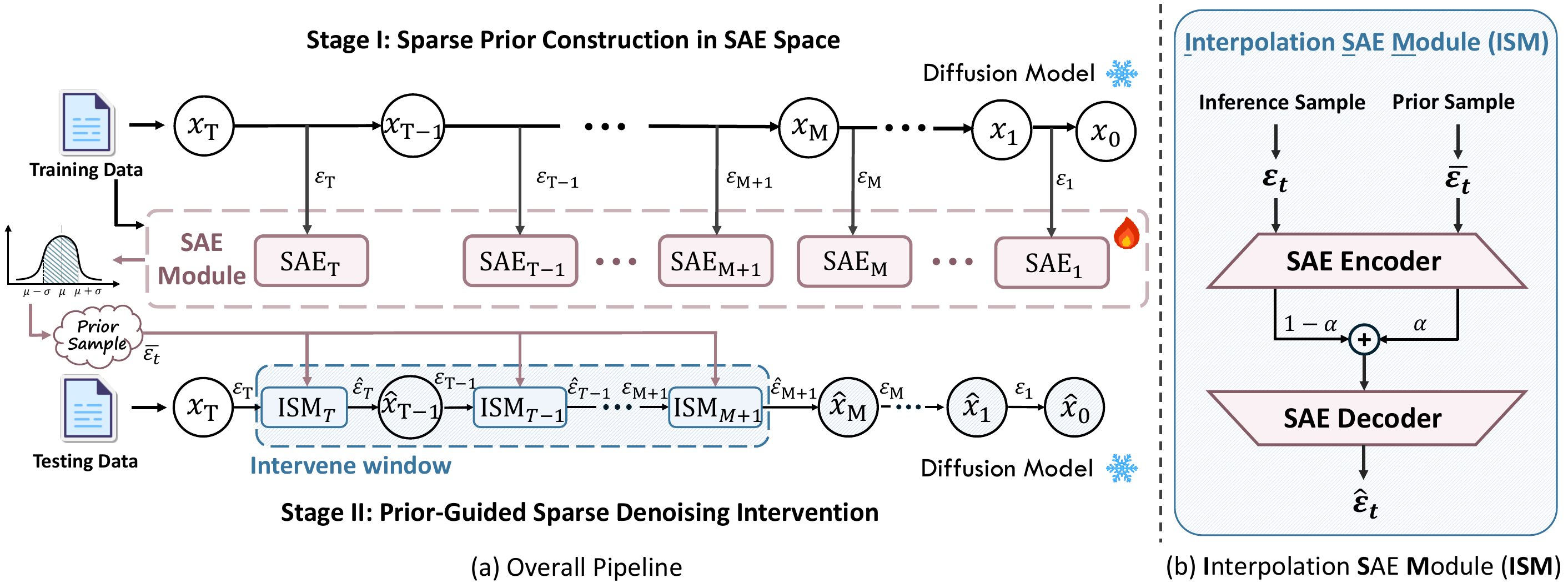}
\vspace{-10pt}
\caption{Overview of the proposed two-stage framework. Stage I learns
timestep-specific SAEs, filters prototype-inconsistent samples, and
constructs class-conditioned sparse prototypes. Stage II performs
prototype-guided interpolation in SAE space at selected denoising
steps and decodes the corrected representation for subsequent sampling.}
\Description{A two-stage pipeline. Stage I trains timestep-specific sparse autoencoders and builds class prototypes. Stage II interpolates test-sample sparse codes with those prototypes during selected early denoising steps.}
\label{fig:framework}
\end{figure*}

This diagnosis procedure plays two roles in the framework as shown in Fig.~\ref{fig:framework}. First, it makes concept brittleness measurable in SAE space by exposing how atypical or failed samples depart from the class-consistent sparse structure. Second, it converts the reliable class structure into a token-level prototype that can later be used as an internal prior during inference. In this way, diagnosis and correction are linked through the same sparse representation: the structure used to identify deviations in Stage I also provides the reference used for intervention in Stage II.

\subsection{Prototype-Guided Correction}
Once the class-conditioned sparse prototypes $R_c^{(t)}$ are available, we use them to correct denoising trajectories during inference. The goal of this step is to test whether restoring the class-consistent sparse structure identified in Sec.~\ref{ssec:diag} is sufficient to alter a failure outcome. In this sense, correction serves both as an intervention mechanism and as a probe of whether the diagnosed representational mismatch is relevant to the final generation.

Given a test prompt with target class $c^\star$, we intervene at selected timesteps $t\in\mathcal{T}_{\mathrm{guide}}$. The current step-update feature $X_{\mathrm{test}}^{(t)}$ is normalized and encoded into the timestep-specific SAE space:
\begin{equation}
Z_{\mathrm{test}}^{(t)}
=
\mathrm{Enc}^{(t)}\!\left(
\frac{X_{\mathrm{test}}^{(t)}-\mu^{(t)}}
{\sigma^{(t)}+\varepsilon}
\right).
\end{equation}
We then interpolate it toward the target-class prototype:
\begin{equation}
\label{eq:interpolate}
\bar{Z}^{(t)}
=
(1-\alpha)Z_{\mathrm{test}}^{(t)}
+\alpha R_{c^\star}^{(t)},
\end{equation}
where $\alpha\in[0,1]$ controls the intervention strength. The corrected code is decoded and restored to the original feature scale:
\begin{equation}
X_{\mathrm{guided}}^{(t)}
=
\mathrm{Dec}^{(t)}\!\left(\bar{Z}^{(t)}\right)
\odot(\sigma^{(t)}+\varepsilon)+\mu^{(t)}.
\end{equation}
The resulting feature is used as the current sampler update, while unguided timesteps retain the original update. Thus, the intervention directly modifies the step-update quantity rather than an arbitrary hidden state.

This design has two advantages. First, it keeps the base diffusion model frozen and performs correction entirely at inference time, avoiding any task-specific finetuning. Second, the intervention is performed in the same timestep-specific sparse space used for diagnosis, so its effect remains directly tied to the class-consistent structure identified in Stage I. If moving a failed trajectory toward this sparse prototype changes the final generation, the diagnosed sparse mismatch is functionally relevant to the failure outcome.

In practice, we do not apply this correction uniformly across all timesteps. As will be shown in Sec.~\ref{ssec:time}, the effect of intervention is strongly time-dependent, with early denoising steps providing substantially larger leverage than middle or late stages. We therefore apply prototype-guided correction only within a selected guidance window, which allows the method to improve condition realization while avoiding unnecessary distortion of unrelated content. Figure~\ref{fig:framework} illustrates this procedure through the interpolation SAE module (ISM), where the current step-update feature is encoded, interpolated with the corresponding sparse prototype, decoded, and then returned to the sampler for continued denoising.

\section{Experiments}
\subsection{Experimental Setup}

\noindent{\textbf{Tasks and Dataset.}}
We evaluate our framework on two held-out tasks: \emph{style generation} and \emph{attribute control}. To support both under a unified protocol, we construct a dataset with separate training and test splits, since existing public datasets do not jointly emphasize controlled content variation, explicit content--condition separation, and matched coverage of style and attribute concepts required for analyzing object-dependent concept brittleness. The training split contains 2,080 samples covering 10 styles and 10 attribute terms, while the test split contains 1,780 samples from more than 150 entity categories and follows the prompt template ``\texttt{[color] [texture] [shape] [entity], [style]}''. The training split is used only for feature extraction, SAE training, sample filtering, and sparse-prior construction; the test split is used only for final evaluation.

\noindent{\textbf{Backbones.}}
We evaluate five diffusion backbones: Stable Diffusion 1.5~\cite{rombach2022latent}, Stable Diffusion 3.5~\cite{stabilityai2024sd35}, Stable Diffusion XL~\cite{podell2023sdxl}, PixArt-Alpha~\cite{chen2024pixartalpha}, and FLUX.1-dev~\cite{blackforestlabs2024flux1}. We follow the official sampler settings of each model: SD 1.5, SDXL, and FLUX.1-dev use 50 denoising steps, SD 3.5 uses 40, and PixArt-Alpha uses 20. All models use their default guidance scales and generate images at $256 \times 256$ resolution. 

\noindent{\textbf{Quantitative Metrics.}}
For style generation, we report \textbf{CLIP-Image (CLIP-I)} for similarity to target-style references, \textbf{CLIP-Text (CLIP-T)} for full-prompt alignment, \textbf{Style Alignment} for target-style consistency, and inference time, while treating style-specific \textbf{FID}~\cite{heusel2017ttur} as a secondary reference metric. For attribute control, we report \textbf{BLIP-VQA}~\cite{li2022blip} scores for \textit{Color}, \textit{Texture}, and \textit{Shape}, together with attribute-conditioned \textbf{CLIP Similarity (CLIPSim)}~\cite{radford2021clip} and inference time. Following T2I-CompBench++~\cite{huang2025t2icompbenchpp}, BLIP-VQA uses attribute-specific yes/no questions, while CLIPSim measures semantic alignment with attribute-focused text.

\noindent{\textbf{Implementation Details.}} 
At each timestep $t$, we extract the tokenized backbone prediction $X^{(t)} \in \mathbb{R}^{P \times D}$ with $P=256$ spatial tokens and $D=64$ channels. We train one timestep-specific SAE per denoising step using flattened token features, with per-step z-score normalization. Unless otherwise stated, we use $d_{\text{model}}=64$, latent width $K_d=256$ ($4\times$ expansion), Top-$K$ sparsity $k=10$, and $\mathrm{auxk}=128$. Training uses 200 epochs, batch size 40{,}960, learning rate $1\times10^{-3}$, auxiliary loss coefficient $1/32$, and dead-feature threshold 50. During inference, prototype-guided correction uses $\alpha=0.8$ and is applied to the earliest 20\% of denoising steps. All quantitative results are obtained with a fixed test seed of 2026 on a single NVIDIA A800 80GB GPU.

\begin{table*}[t]
\caption{Main results on style generation and attribute control tasks compared with five widely used diffusion model backbones.}
\vspace{-10pt}
\label{tab:main_results}
\centering
\setlength{\tabcolsep}{5pt}
\begin{tabular}{lcccccccc}
\toprule
\multirow{2}{*}{Model}
& \multicolumn{3}{c}{Style Generation}
& \multicolumn{4}{c}{Attribute Control}
& \multirow{2}{*}{\shortstack{Infer Time (s)\\(1-sample end-to-end)}} \\
\cmidrule(lr){2-4} \cmidrule(lr){5-8}
& CLIP-I~$\uparrow$ & CLIP-T~$\uparrow$ & Style Align.~$\uparrow$
& Color~$\uparrow$ & Texture~$\uparrow$ & Shape~$\uparrow$ & CLIPSim~$\uparrow$
& \\
\midrule

SD 1.5~\cite{rombach2022latent}
& 0.644 & 0.327 & 0.237
& 0.204 & 0.195 & 0.236 & 0.217
& $1.227{\pm}0.045$ \\

SD 1.5 + Ours
& \textbf{0.659} & \textbf{0.333} & \textbf{0.239}
& \textbf{0.336} & \textbf{0.328} & \textbf{0.330} & \textbf{0.231}
& $1.261{\pm}0.062$ \\

\midrule

SD 3.5~\cite{stabilityai2024sd35}
& 0.725 & 0.261 & 0.220
& 0.341 & 0.302 & 0.451 & 0.245
& $15.407{\pm}0.100$ \\

SD 3.5 + Ours
& \textbf{0.797} & \textbf{0.297} & \textbf{0.221}
& \textbf{0.461} & \textbf{0.437} & \textbf{0.475} & \textbf{0.250}
& $15.701{\pm}0.598$ \\

\midrule

SDXL~\cite{podell2023sdxl}
& 0.630 & 0.341 & 0.238
& 0.353 & 0.276 & 0.300 & 0.235
& $6.388{\pm}0.011$ \\

SDXL + Ours
& \textbf{0.653} & \textbf{0.359} & \textbf{0.243}
& \textbf{0.415} & \textbf{0.361} & \textbf{0.318} & \textbf{0.237}
& $6.393{\pm}0.034$ \\

\midrule

PixArt-Alpha~\cite{chen2024pixartalpha}
& 0.627 & 0.336 & 0.232
& 0.349 & 0.356 & 0.315 & 0.222
& $0.586{\pm}0.013$ \\

PixArt-Alpha + Ours
& \textbf{0.629} & \textbf{0.338} & \textbf{0.239}
& \textbf{0.368} & \textbf{0.366} & \textbf{0.358} & \textbf{0.247}
& $0.609{\pm}0.069$ \\

\midrule

Flux.1-dev~\cite{blackforestlabs2024flux1}
& 0.598 & 0.273 & 0.220
& 0.379 & 0.326 & 0.240 & 0.229
& $4.104{\pm}0.014$ \\

Flux.1-dev + Ours
& \textbf{0.623} & \textbf{0.288} & \textbf{0.230}
& \textbf{0.394} & \textbf{0.346} & \textbf{0.379} & \textbf{0.233}
& $4.149{\pm}0.028$ \\

\bottomrule
\end{tabular}
\vspace{-8pt}
\end{table*}

\subsection{Main Results on Style Generation}
We first examine whether the proposed sparse-prior intervention can improve style generation consistently across different diffusion backbones on held-out prompts. As shown in Table~\ref{tab:main_results}, across all five backbones, our method improves the prompt-based metrics that most directly reflect style expression and text fidelity, while introducing only negligible runtime overhead. The gains are especially pronounced on stronger backbones such as SD 3.5, suggesting that the learned sparse prior functions as a backbone-agnostic mechanism for restoring missing style evidence, rather than merely as a heuristic tailored to a particular model. 

Importantly, our goal is not to match a broad reference distribution for a given style domain, but to improve prompt-level concept expression under held-out prompts. For this reason, we treat CLIP-Image, CLIP-Text, and Style Alignment as the primary evidence in the main text, since they track prompt-conditioned improvement more directly than style-specific FID. The negligible increase in inference time further supports our claim that the proposed module functions as a lightweight plug-in, rather than as a control mechanism that depends on costly optimization during inference.

\begin{figure}[t]
\centering
\includegraphics[width=\linewidth]{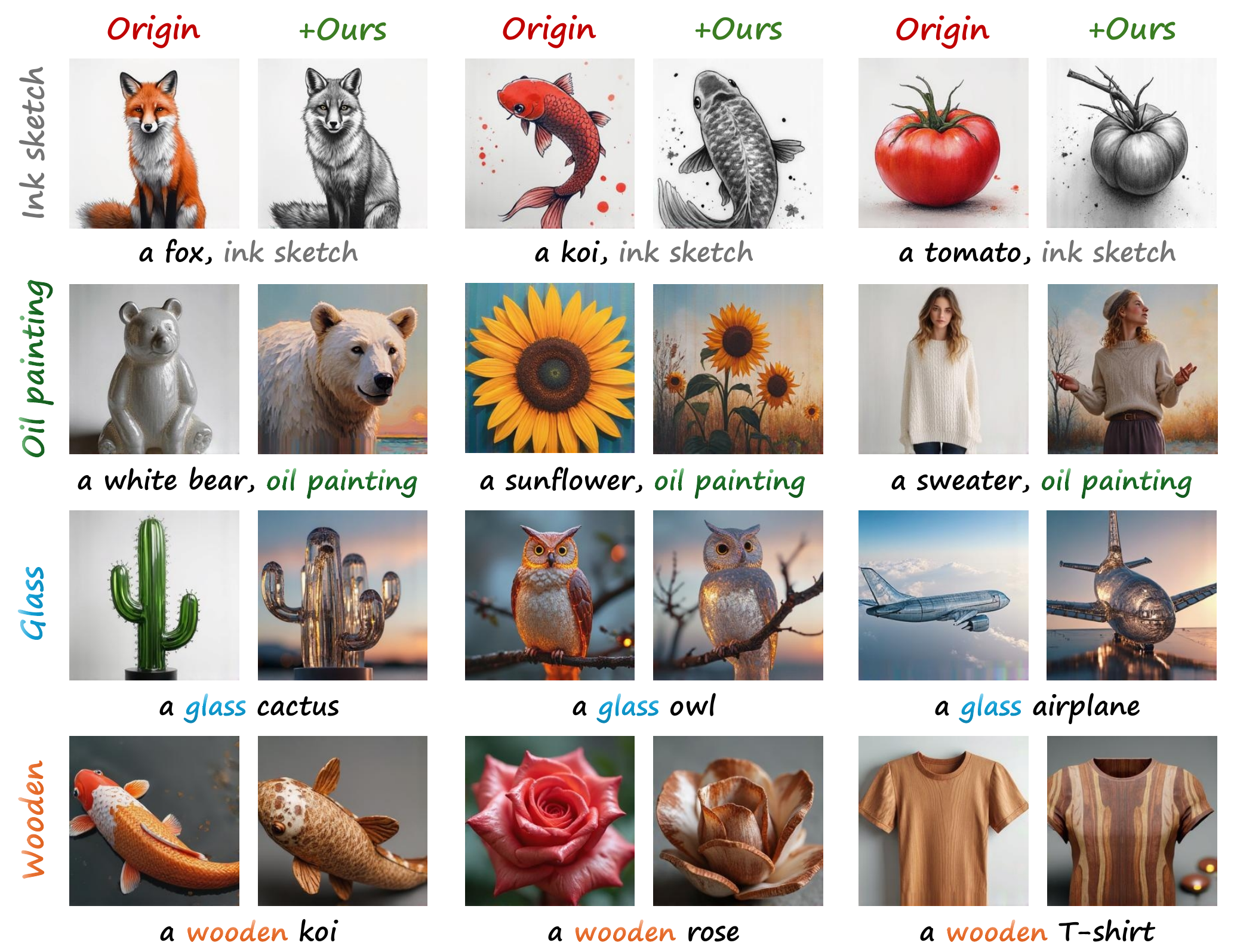}
\vspace{-8pt}
\caption{Qualitative correction results on representative style and attribute failures generated by Flux~\cite{blackforestlabs2024flux1}.}
\Description{Before-and-after examples for ink sketch, oil painting, glass, and wooden concepts. The proposed correction strengthens each requested concept while preserving the depicted object.}
\vspace{-5pt}
\label{fig:sample}
\vspace{-5pt}
\end{figure}

\subsection{Main Results on Attribute Control}
We next explore whether the same mechanism extends beyond global style cues to more fine-grained attribute evidence. As shown in the right half of Table~\ref{tab:main_results}, our proposed method improves most or all of the Color, Texture, and Shape dimensions, while maintaining similarly small runtime overhead. This suggests that the learned sparse prior is not merely a style-specific filter, but instead captures a more general form of concept evidence that also applies to localized and semantically precise attributes.

This result is meaningful because attribute control is qualitatively different from style generation. Whereas style cues are often global, attributes such as color, material, and shape are finer-grained and more difficult to preserve consistently. The fact that the same intervention improves performance in both settings provides strong evidence that the proposed framework generalizes not only across model architectures, but also across concept families, while retaining its lightweight plug-and-play nature.

Qualitative results in Fig.~\ref{fig:sample} further support these findings: our intervention strengthens the target concept while largely preserving object identity and overall content.
\section{Analysis}

\subsection{Where Is Concept Evidence Most Legible?}

We first analyze where concept evidence is most legible inside the diffusion backbone and whether clearer class structure also supports stronger intervention. To isolate representation depth, we compare three FLUX readouts---Last Double DiT Layer, Last Single DiT Layer, and the final Noise Latent---while fixing the training set, SAE procedure, prototype construction, intervention protocol, test set, and random seed. Each SAE uses a dictionary width four times the input dimensionality.

We begin by examining class structure in sparse space. Table~\ref{tab:layer_pre} shows a clear depth-dependent trend: both 5-fold linear-probe accuracy (\textbf{Acc}) and separation ratio (\textbf{Sep}) improve toward the denoising output. Noise Latent achieves the best separability, followed by Last Single DiT Layer, while Last Double DiT Layer performs worst. This indicates that concept evidence becomes progressively more class-selective in later representations.

We next test whether improved legibility translates into stronger intervention. Using the same Stage-II procedure, we apply concept-guided correction with the SAE bank and class priors learned from each layer. Table~\ref{tab:layer} follows the same ordering as Table~\ref{tab:layer_pre}: Noise Latent achieves the best prompt-conditioned performance.
Although FID varies only mildly, the prompt-conditioned metrics consistently favor the denoising output. Thus, better separability reflects actionable evidence rather than merely a visualization artifact.
Later representations are therefore both more legible in SAE space and more effective for concept-level correction. We adopt the denoising output as the default readout because it offers the best trade-off between interpretability and intervention effectiveness.

\begin{table}[t]
\caption{Comparison of different representation layers for concept separation.}
\vspace{-8pt}
\label{tab:layer_pre}
\centering
\small
\setlength{\tabcolsep}{4pt}
\renewcommand{\arraystretch}{1.0}
\begin{tabular}{lcc}
    \toprule
    Representation Layer & Acc~$\uparrow$ & Sep~$\uparrow$ \\
    \midrule
    \rowcolor{blue!10}
    Noise Latent (ours) & \textbf{0.7648} & \textbf{1.8479} \\
    Last Single DiT Layer & \underline{0.7284} & \underline{1.8245} \\
    Last Double DiT Layer & 0.7108 & 1.7120 \\
    \bottomrule
\end{tabular}
\vspace{-7pt}
\end{table}


\begin{table}[t]
\caption{Effect of representation layer on style correction.}
\vspace{-8pt}
\label{tab:layer}
\centering
\resizebox{\columnwidth}{!}{
    \begin{tabular}{lcccc}
        \toprule
        Representation & CLIP-I~$\uparrow$ & CLIP-T~$\uparrow$ & Style Alignment~$\uparrow$ & FID~$\downarrow$ \\
        \midrule
        \rowcolor{blue!10}
        Noise Latent (ours) & \textbf{0.623} & \textbf{0.288} & \textbf{0.230} & 293.3 \\
        Last Single DiT Layer & \underline{0.621} & \underline{0.284} & \underline{0.229} & \underline{293.0} \\
        Last Double DiT Layer & 0.606 & 0.275 & 0.224 & \textbf{291.8} \\
        \bottomrule
    \end{tabular}
}
\vspace{-8pt}
\end{table}
\FloatBarrier

\subsection{When Is Concept Evidence Most Correctable?}
\label{ssec:time}

We next ask when concept evidence remains correctable. Because early updates establish concept evidence while later ones mainly refine committed content, we combine single-step mean ablation to measure timestep leverage with intervention-window ablation to identify the most effective correction period.

\begin{figure}[t]
\centering
\includegraphics[width=0.95\linewidth]{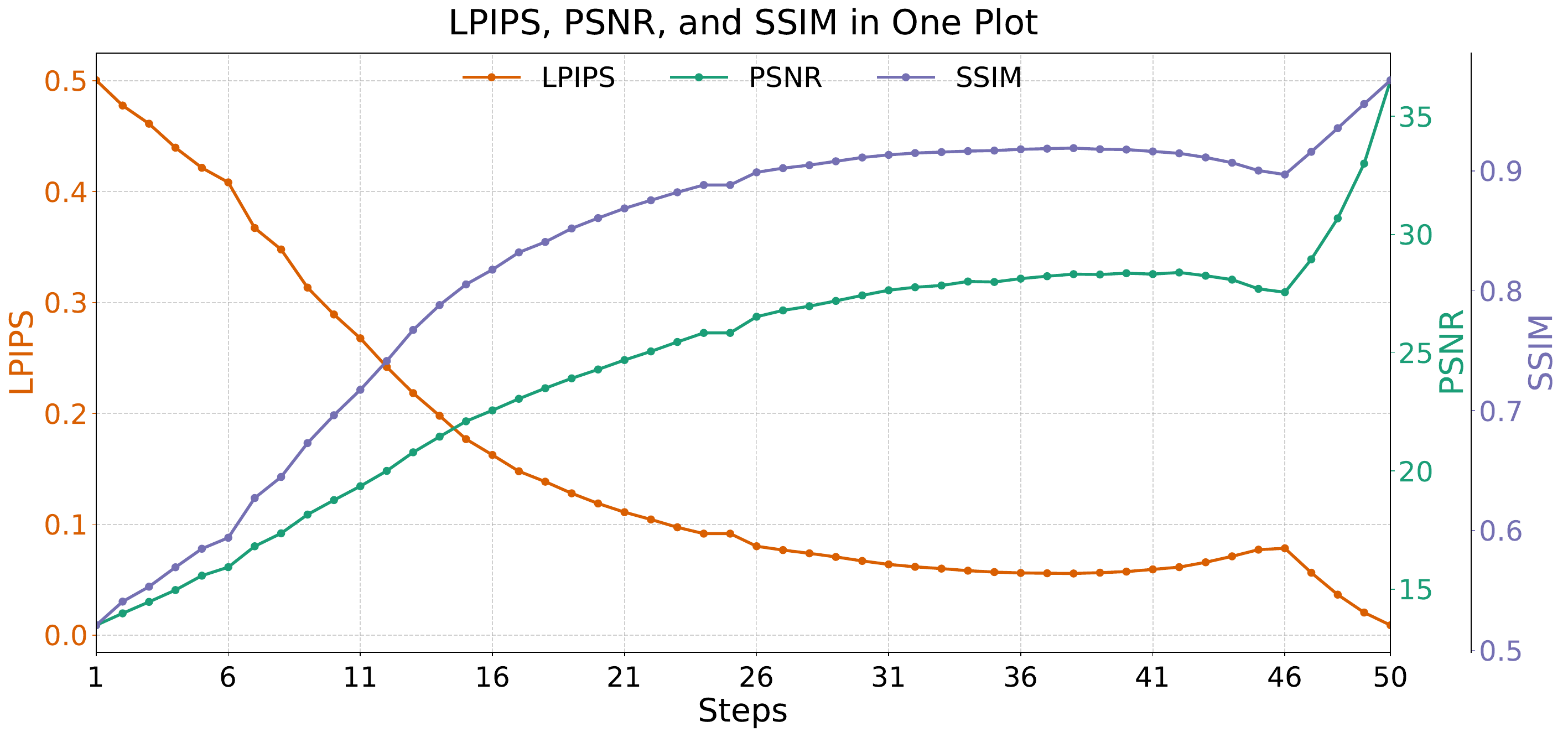}
\vspace{-8pt}
\caption{Effect of single-step mean ablation at different denoising timesteps.}
\Description{Line chart of LPIPS, PSNR, and SSIM over fifty denoising steps. Perturbations at early steps cause the largest output change, while later-step effects are weaker.}
\label{fig:time}
\vspace{-8pt}
\end{figure}

For single-step mean ablation, we replace one update in a normal trajectory with the mean feature of other samples at the same timestep, excluding the current sample. In Fig.~\ref{fig:time}, early perturbations cause the largest deviations in LPIPS, PSNR, and SSIM, while middle and late perturbations are progressively weaker. Because this operation preserves the overall feature scale while removing sample-specific information, it reveals the intrinsic leverage of each timestep.
Thus, early timesteps carry high-leverage concept evidence whose errors propagate most strongly to the final image.

We then compare ``Early Only (0--20\%)'', ``Middle Only (20--60\%)'', ``Late Only (60--100\%)'', ``Early + Middle'', and ``All Steps'' with other settings fixed. Table~\ref{tab:time_} shows that ``Early Only'' performs best overall, whereas ``Middle Only'' and ``Late Only'' are progressively weaker. ``Early + Middle'' slightly raises Style Alignment but lowers CLIP-Text and worsens FID, and ``All Steps'' performs worst. Extending intervention therefore overwrites formed content and leads to overcorrection.

\begin{table}[t]
\caption{Effect of guidance window on style correction.}
\vspace{-8pt}
\label{tab:time_}
\centering
\resizebox{\columnwidth}{!}{
    \begin{tabular}{lcccc}
        \toprule
        Window & CLIP-I~$\uparrow$ & CLIP-T~$\uparrow$ & Style Alignment~$\uparrow$ & FID~$\downarrow$ \\
        \midrule
        \rowcolor{blue!10}
        Early Only (ours)& \textbf{0.623} & \textbf{0.288} & \underline{0.230} & \textbf{293.3} \\
        Middle Only& 0.616 & \underline{0.274} & \underline{0.230} & \underline{327.4} \\
        Late Only& 0.593 & 0.271 & 0.224 & 351.2 \\
        Early + Middle & \underline{0.621} & 0.239 & \textbf{0.233} & 371.9 \\
        All Steps & 0.603 & 0.206 & 0.221 & 481.4 \\
        \bottomrule
    \end{tabular}
}
\vspace{-8pt}
\end{table}

Together, Fig.~\ref{fig:time} and Table~\ref{tab:time_} identify a limited early correction window. Early steps determine whether concept evidence is successfully established, while later steps offer little room for repair and risk overcorrection. This temporal asymmetry explains why our framework restricts intervention to an early subset of timesteps rather than applying it throughout the trajectory.

\section{Conclusion}

In this paper, we study object-dependent concept brittleness in text-to-image diffusion models, where the same target condition succeeds for most objects but fails on a small subset of closely related prompts. To diagnose this phenomenon, we introduce a step-wise SAE framework that maps denoising features into a sparse space where successful and failed trajectories can be compared directly. The resulting structured mismatches in condition-related activations make object-dependent brittleness more interpretable.

Using class-level sparse prototypes from reliable class-consistent samples, we perform lightweight inference-time correction that also probes whether the diagnosed discrepancy is relevant to the failure outcome. Experiments across style and attribute settings on five diffusion backbones show consistent gains in condition consistency and repair success. Our analysis further shows that condition-related information becomes clearer in deeper features, while effective correction is concentrated in early timesteps.
Future work includes extending this analysis to larger diffusion architectures, video generation models, and multimodal agent settings~\cite{wu2026promsa}, and developing more fine-grained interventions and broader benchmarks for content-dependent condition failure.


\bibliographystyle{ACM-Reference-Format}
\balance
\bibliography{ref}


\begin{thebibliography}{42}


\ifx \showCODEN    \undefined \def \showCODEN     #1{\unskip}     \fi
\ifx \showISBNx    \undefined \def \showISBNx     #1{\unskip}     \fi
\ifx \showISBNxiii \undefined \def \showISBNxiii  #1{\unskip}     \fi
\ifx \showISSN     \undefined \def \showISSN      #1{\unskip}     \fi
\ifx \showLCCN     \undefined \def \showLCCN      #1{\unskip}     \fi
\ifx \shownote     \undefined \def \shownote      #1{#1}          \fi
\ifx \showarticletitle \undefined \def \showarticletitle #1{#1}   \fi
\ifx \showURL      \undefined \def \showURL       {\relax}        \fi
\providecommand\bibfield[2]{#2}
\providecommand\bibinfo[2]{#2}
\providecommand\natexlab[1]{#1}
\providecommand\showeprint[2][]{arXiv:#2}

\bibitem[Betker et~al\mbox{.}(2023)]%
        {betker2023improving}
\bibfield{author}{\bibinfo{person}{James Betker}, \bibinfo{person}{Gabriel
  Goh}, \bibinfo{person}{Li Jing}, \bibinfo{person}{Tim Brooks},
  \bibinfo{person}{Jianfeng Wang}, \bibinfo{person}{Linjie Li},
  \bibinfo{person}{Long Ouyang}, \bibinfo{person}{Juntang Zhuang},
  \bibinfo{person}{Joyce Lee}, \bibinfo{person}{Yufei Guo},
  \bibinfo{person}{Wesam Manassra}, \bibinfo{person}{Prafulla Dhariwal},
  \bibinfo{person}{Casey Chu}, \bibinfo{person}{Yunxin Jiao}, {and}
  \bibinfo{person}{Aditya Ramesh}.} \bibinfo{year}{2023}\natexlab{}.
\newblock \bibinfo{title}{Improving Image Generation with Better Captions}.
\newblock
\urldef\tempurl%
\url{https://cdn.openai.com/papers/dall-e-3.pdf}
\showURL{%
\tempurl}


\bibitem[{{Black Forest Labs}}(2024)]%
        {blackforestlabs2024flux1}
\bibfield{author}{\bibinfo{person}{{{Black Forest Labs}}}.}
  \bibinfo{year}{2024}\natexlab{}.
\newblock \bibinfo{title}{{{FLUX.1}}}.
\newblock
  \bibinfo{howpublished}{\url{https://blackforestlabs.ai/announcing-black-forest-labs/}}.
\newblock


\bibitem[Brack et~al\mbox{.}(2023)]%
        {brack2023sega}
\bibfield{author}{\bibinfo{person}{Manuel Brack}, \bibinfo{person}{Felix
  Friedrich}, \bibinfo{person}{Dominik Hintersdorf}, \bibinfo{person}{Lukas
  Struppek}, \bibinfo{person}{Patrick Schramowski}, {and}
  \bibinfo{person}{Kristian Kersting}.} \bibinfo{year}{2023}\natexlab{}.
\newblock \showarticletitle{{SEGA}: Instructing Text-to-Image Models Using
  Semantic Guidance}. In \bibinfo{booktitle}{\emph{Advances in Neural
  Information Processing Systems}}, \bibfield{editor}{\bibinfo{person}{A.~Oh},
  \bibinfo{person}{T.~Naumann}, \bibinfo{person}{A.~Globerson},
  \bibinfo{person}{K.~Saenko}, \bibinfo{person}{M.~Hardt}, {and}
  \bibinfo{person}{S.~Levine}} (Eds.), Vol.~\bibinfo{volume}{36}.
  \bibinfo{publisher}{Curran Associates, Inc.}, \bibinfo{address}{Red Hook, NY,
  USA}, \bibinfo{pages}{25365--25389}.
\newblock
\href{https://doi.org/10.52202/075280-1102}{doi:\nolinkurl{10.52202/075280-1102}}


\bibitem[Bricken et~al\mbox{.}(2023)]%
        {bricken2023monosemanticity}
\bibfield{author}{\bibinfo{person}{Trenton Bricken}, \bibinfo{person}{Adly
  Templeton}, \bibinfo{person}{Joshua Batson}, \bibinfo{person}{Brian Chen},
  \bibinfo{person}{Adam Jermyn}, \bibinfo{person}{Tom Conerly},
  \bibinfo{person}{Nicholas~L. Turner}, \bibinfo{person}{Cem Anil},
  \bibinfo{person}{Carson Denison}, \bibinfo{person}{Amanda Askell},
  \bibinfo{person}{Robert Lasenby}, \bibinfo{person}{Yifan Wu},
  \bibinfo{person}{Shauna Kravec}, \bibinfo{person}{Nicholas Schiefer},
  \bibinfo{person}{Tim Maxwell}, \bibinfo{person}{Nicholas Joseph},
  \bibinfo{person}{Alex Tamkin}, \bibinfo{person}{Karina Nguyen},
  \bibinfo{person}{Brayden McLean}, \bibinfo{person}{Josiah~E. Burke},
  \bibinfo{person}{Tristan Hume}, \bibinfo{person}{Shan Carter},
  \bibinfo{person}{Tom Henighan}, {and} \bibinfo{person}{Chris Olah}.}
  \bibinfo{year}{2023}\natexlab{}.
\newblock \bibinfo{title}{Towards Monosemanticity: Decomposing Language Models
  With Dictionary Learning}.
\newblock \bibinfo{howpublished}{Transformer Circuits Thread}.
\newblock
\urldef\tempurl%
\url{https://transformer-circuits.pub/2023/monosemantic-features/index.html}
\showURL{%
\tempurl}


\bibitem[Cao et~al\mbox{.}(2025)]%
        {cao2025repldm}
\bibfield{author}{\bibinfo{person}{Boyuan Cao}, \bibinfo{person}{Jiaxin Ye},
  \bibinfo{person}{Yujie Wei}, {and} \bibinfo{person}{Hongming Shan}.}
  \bibinfo{year}{2025}\natexlab{}.
\newblock \showarticletitle{{RepLDM}: Reprogramming Pretrained Latent Diffusion
  Models for High-Quality, High-Efficiency, High-Resolution Image Generation}.
  In \bibinfo{booktitle}{\emph{Advances in Neural Information Processing
  Systems}}, \bibfield{editor}{\bibinfo{person}{D.~Belgrave},
  \bibinfo{person}{C.~Zhang}, \bibinfo{person}{H.~Lin},
  \bibinfo{person}{R.~Pascanu}, \bibinfo{person}{P.~Koniusz},
  \bibinfo{person}{M.~Ghassemi}, {and} \bibinfo{person}{N.~Chen}} (Eds.),
  Vol.~\bibinfo{volume}{38}. \bibinfo{publisher}{Curran Associates, Inc.},
  \bibinfo{address}{Red Hook, NY, USA}, \bibinfo{pages}{61316--61351}.
\newblock
\href{https://doi.org/10.52202/085713-2049}{doi:\nolinkurl{10.52202/085713-2049}}


\bibitem[Chefer et~al\mbox{.}(2023)]%
        {chefer2023attend}
\bibfield{author}{\bibinfo{person}{Hila Chefer}, \bibinfo{person}{Yuval
  Alaluf}, \bibinfo{person}{Yael Vinker}, \bibinfo{person}{Lior Wolf}, {and}
  \bibinfo{person}{Daniel Cohen-Or}.} \bibinfo{year}{2023}\natexlab{}.
\newblock \showarticletitle{Attend-and-Excite: Attention-Based Semantic
  Guidance for Text-to-Image Diffusion Models}.
\newblock \bibinfo{journal}{\emph{ACM Transactions on Graphics}}
  \bibinfo{volume}{42}, \bibinfo{number}{4} (\bibinfo{year}{2023}),
  \bibinfo{pages}{148:1--148:10}.
\newblock
\href{https://doi.org/10.1145/3592116}{doi:\nolinkurl{10.1145/3592116}}


\bibitem[Chen et~al\mbox{.}(2024b)]%
        {chen2024trajectory}
\bibfield{author}{\bibinfo{person}{Defang Chen}, \bibinfo{person}{Zhenyu Zhou},
  \bibinfo{person}{Can Wang}, \bibinfo{person}{Chunhua Shen}, {and}
  \bibinfo{person}{Siwei Lyu}.} \bibinfo{year}{2024}\natexlab{b}.
\newblock \showarticletitle{On the Trajectory Regularity of {{ODE}}-Based
  Diffusion Sampling}. In \bibinfo{booktitle}{\emph{Proceedings of the 41st
  International Conference on Machine Learning}}
  \emph{(\bibinfo{series}{Proceedings of Machine Learning Research},
  Vol.~\bibinfo{volume}{235})}, \bibfield{editor}{\bibinfo{person}{Ruslan
  Salakhutdinov}, \bibinfo{person}{Zico Kolter}, \bibinfo{person}{Katherine
  Heller}, \bibinfo{person}{Adrian Weller}, \bibinfo{person}{Nuria Oliver},
  \bibinfo{person}{Jonathan Scarlett}, {and} \bibinfo{person}{Felix
  Berkenkamp}} (Eds.). \bibinfo{publisher}{PMLR}, \bibinfo{address}{Vienna,
  Austria}, \bibinfo{pages}{7905--7934}.
\newblock
\urldef\tempurl%
\url{https://proceedings.mlr.press/v235/chen24bm.html}
\showURL{%
\tempurl}


\bibitem[Chen et~al\mbox{.}(2024a)]%
        {chen2024pixartalpha}
\bibfield{author}{\bibinfo{person}{Junsong Chen}, \bibinfo{person}{Jincheng
  Yu}, \bibinfo{person}{Chongjian Ge}, \bibinfo{person}{Lewei Yao},
  \bibinfo{person}{Enze Xie}, \bibinfo{person}{Zhongdao Wang},
  \bibinfo{person}{James Kwok}, \bibinfo{person}{Ping Luo},
  \bibinfo{person}{Huchuan Lu}, {and} \bibinfo{person}{Zhenguo Li}.}
  \bibinfo{year}{2024}\natexlab{a}.
\newblock \showarticletitle{{PixArt}-$\alpha$: Fast Training of Diffusion
  Transformer for Photorealistic Text-to-Image Synthesis}. In
  \bibinfo{booktitle}{\emph{The Twelfth International Conference on Learning
  Representations}}. \bibinfo{publisher}{OpenReview.net},
  \bibinfo{address}{Vienna, Austria}.
\newblock
\urldef\tempurl%
\url{https://openreview.net/forum?id=eAKmQPe3m1}
\showURL{%
\tempurl}


\bibitem[Cywi{\'n}ski and Deja(2025)]%
        {cywinski2025saeuron}
\bibfield{author}{\bibinfo{person}{Bartosz Cywi{\'n}ski} {and}
  \bibinfo{person}{Kamil Deja}.} \bibinfo{year}{2025}\natexlab{}.
\newblock \showarticletitle{{SAeUron}: Interpretable Concept Unlearning in
  Diffusion Models with Sparse Autoencoders}. In
  \bibinfo{booktitle}{\emph{Proceedings of the 42nd International Conference on
  Machine Learning}} \emph{(\bibinfo{series}{Proceedings of Machine Learning
  Research}, Vol.~\bibinfo{volume}{267})},
  \bibfield{editor}{\bibinfo{person}{Aarti Singh}, \bibinfo{person}{Maryam
  Fazel}, \bibinfo{person}{Daniel Hsu}, \bibinfo{person}{Simon Lacoste-Julien},
  \bibinfo{person}{Felix Berkenkamp}, \bibinfo{person}{Tegan Maharaj},
  \bibinfo{person}{Kiri Wagstaff}, {and} \bibinfo{person}{Jerry Zhu}} (Eds.).
  \bibinfo{publisher}{PMLR}, \bibinfo{address}{Vancouver, Canada},
  \bibinfo{pages}{11738--11775}.
\newblock
\urldef\tempurl%
\url{https://proceedings.mlr.press/v267/cywinski25a.html}
\showURL{%
\tempurl}


\bibitem[Gao et~al\mbox{.}(2025)]%
        {gao2025scaling}
\bibfield{author}{\bibinfo{person}{Leo Gao}, \bibinfo{person}{Tom Dupr{\'e}~la
  Tour}, \bibinfo{person}{Henk Tillman}, \bibinfo{person}{Gabriel Goh},
  \bibinfo{person}{Rajan Troll}, \bibinfo{person}{Alec Radford},
  \bibinfo{person}{Ilya Sutskever}, \bibinfo{person}{Jan Leike}, {and}
  \bibinfo{person}{Jeffrey Wu}.} \bibinfo{year}{2025}\natexlab{}.
\newblock \showarticletitle{Scaling and Evaluating Sparse Autoencoders}. In
  \bibinfo{booktitle}{\emph{The Thirteenth International Conference on Learning
  Representations}}. \bibinfo{publisher}{OpenReview.net},
  \bibinfo{address}{Singapore}.
\newblock
\urldef\tempurl%
\url{https://openreview.net/forum?id=tcsZt9ZNKD}
\showURL{%
\tempurl}


\bibitem[Ghosh et~al\mbox{.}(2023)]%
        {ghosh2023geneval}
\bibfield{author}{\bibinfo{person}{Dhruba Ghosh}, \bibinfo{person}{Hannaneh
  Hajishirzi}, {and} \bibinfo{person}{Ludwig Schmidt}.}
  \bibinfo{year}{2023}\natexlab{}.
\newblock \showarticletitle{{GenEval}: An Object-Focused Framework for
  Evaluating Text-to-Image Alignment}. In \bibinfo{booktitle}{\emph{Advances in
  Neural Information Processing Systems}},
  \bibfield{editor}{\bibinfo{person}{A.~Oh}, \bibinfo{person}{T.~Naumann},
  \bibinfo{person}{A.~Globerson}, \bibinfo{person}{K.~Saenko},
  \bibinfo{person}{M.~Hardt}, {and} \bibinfo{person}{S.~Levine}} (Eds.),
  Vol.~\bibinfo{volume}{36}. \bibinfo{publisher}{Curran Associates, Inc.},
  \bibinfo{address}{Red Hook, NY, USA}, \bibinfo{pages}{52132--52152}.
\newblock
\href{https://doi.org/10.52202/075280-2270}{doi:\nolinkurl{10.52202/075280-2270}}


\bibitem[G{\"o}rg{\"u}n et~al\mbox{.}(2026)]%
        {gorgun2026temporal}
\bibfield{author}{\bibinfo{person}{Ada G{\"o}rg{\"u}n}, \bibinfo{person}{Fawaz
  Sammani}, \bibinfo{person}{Nikos Deligiannis}, \bibinfo{person}{Bernt
  Schiele}, {and} \bibinfo{person}{Jonas Fischer}.}
  \bibinfo{year}{2026}\natexlab{}.
\newblock \showarticletitle{Temporal Concept Dynamics in Diffusion Models via
  Prompt-Conditioned Interventions}. In \bibinfo{booktitle}{\emph{The
  Fourteenth International Conference on Learning Representations}}.
  \bibinfo{publisher}{OpenReview.net}, \bibinfo{address}{Rio de Janeiro,
  Brazil}.
\newblock
\urldef\tempurl%
\url{https://openreview.net/forum?id=ABjaSsrYPD}
\showURL{%
\tempurl}


\bibitem[Hahm et~al\mbox{.}(2024)]%
        {hahm2024isometric}
\bibfield{author}{\bibinfo{person}{Jaehoon Hahm}, \bibinfo{person}{Junho Lee},
  \bibinfo{person}{Sunghyun Kim}, {and} \bibinfo{person}{Joonseok Lee}.}
  \bibinfo{year}{2024}\natexlab{}.
\newblock \showarticletitle{Isometric Representation Learning for Disentangled
  Latent Space of Diffusion Models}. In \bibinfo{booktitle}{\emph{Proceedings
  of the 41st International Conference on Machine Learning}}
  \emph{(\bibinfo{series}{Proceedings of Machine Learning Research},
  Vol.~\bibinfo{volume}{235})}, \bibfield{editor}{\bibinfo{person}{Ruslan
  Salakhutdinov}, \bibinfo{person}{Zico Kolter}, \bibinfo{person}{Katherine
  Heller}, \bibinfo{person}{Adrian Weller}, \bibinfo{person}{Nuria Oliver},
  \bibinfo{person}{Jonathan Scarlett}, {and} \bibinfo{person}{Felix
  Berkenkamp}} (Eds.). \bibinfo{publisher}{PMLR}, \bibinfo{address}{Vienna,
  Austria}, \bibinfo{pages}{17224--17245}.
\newblock
\urldef\tempurl%
\url{https://proceedings.mlr.press/v235/hahm24a.html}
\showURL{%
\tempurl}


\bibitem[Helbling et~al\mbox{.}(2025)]%
        {helbling2025conceptattention}
\bibfield{author}{\bibinfo{person}{Alec Helbling}, \bibinfo{person}{Tuna
  Han~Salih Meral}, \bibinfo{person}{Benjamin Hoover}, \bibinfo{person}{Pinar
  Yanardag}, {and} \bibinfo{person}{Duen~Horng Chau}.}
  \bibinfo{year}{2025}\natexlab{}.
\newblock \showarticletitle{ConceptAttention: Diffusion Transformers Learn
  Highly Interpretable Features}. In \bibinfo{booktitle}{\emph{Proceedings of
  the 42nd International Conference on Machine Learning}}
  \emph{(\bibinfo{series}{Proceedings of Machine Learning Research},
  Vol.~\bibinfo{volume}{267})}, \bibfield{editor}{\bibinfo{person}{Aarti
  Singh}, \bibinfo{person}{Maryam Fazel}, \bibinfo{person}{Daniel Hsu},
  \bibinfo{person}{Simon Lacoste-Julien}, \bibinfo{person}{Felix Berkenkamp},
  \bibinfo{person}{Tegan Maharaj}, \bibinfo{person}{Kiri Wagstaff}, {and}
  \bibinfo{person}{Jerry Zhu}} (Eds.). \bibinfo{publisher}{PMLR},
  \bibinfo{address}{Vancouver, Canada}, \bibinfo{pages}{22946--22963}.
\newblock
\urldef\tempurl%
\url{https://proceedings.mlr.press/v267/helbling25a.html}
\showURL{%
\tempurl}


\bibitem[Hertz et~al\mbox{.}(2023)]%
        {hertz2023prompt}
\bibfield{author}{\bibinfo{person}{Amir Hertz}, \bibinfo{person}{Ron Mokady},
  \bibinfo{person}{Jay Tenenbaum}, \bibinfo{person}{Kfir Aberman},
  \bibinfo{person}{Yael Pritch}, {and} \bibinfo{person}{Daniel Cohen-Or}.}
  \bibinfo{year}{2023}\natexlab{}.
\newblock \showarticletitle{Prompt-to-Prompt Image Editing with Cross-Attention
  Control}. In \bibinfo{booktitle}{\emph{The Eleventh International Conference
  on Learning Representations}}. \bibinfo{publisher}{OpenReview.net},
  \bibinfo{address}{Kigali, Rwanda}.
\newblock
\urldef\tempurl%
\url{https://openreview.net/forum?id=_CDixzkzeyb}
\showURL{%
\tempurl}


\bibitem[Heusel et~al\mbox{.}(2017)]%
        {heusel2017ttur}
\bibfield{author}{\bibinfo{person}{Martin Heusel}, \bibinfo{person}{Hubert
  Ramsauer}, \bibinfo{person}{Thomas Unterthiner}, \bibinfo{person}{Bernhard
  Nessler}, {and} \bibinfo{person}{Sepp Hochreiter}.}
  \bibinfo{year}{2017}\natexlab{}.
\newblock \showarticletitle{GANs Trained by a Two Time-Scale Update Rule
  Converge to a Local Nash Equilibrium}. In \bibinfo{booktitle}{\emph{Advances
  in Neural Information Processing Systems}},
  \bibfield{editor}{\bibinfo{person}{I.~Guyon}, \bibinfo{person}{U.~von
  Luxburg}, \bibinfo{person}{S.~Bengio}, \bibinfo{person}{H.~Wallach},
  \bibinfo{person}{R.~Fergus}, \bibinfo{person}{S.~Vishwanathan}, {and}
  \bibinfo{person}{R.~Garnett}} (Eds.), Vol.~\bibinfo{volume}{30}.
  \bibinfo{publisher}{Curran Associates, Inc.}, \bibinfo{address}{Red Hook, NY,
  USA}, \bibinfo{pages}{6626--6637}.
\newblock
\urldef\tempurl%
\url{https://proceedings.neurips.cc/paper/2017/hash/8a1d694707eb0fefe65871369074926d-Abstract.html}
\showURL{%
\tempurl}


\bibitem[Ho et~al\mbox{.}(2020)]%
        {ho2020ddpm}
\bibfield{author}{\bibinfo{person}{Jonathan Ho}, \bibinfo{person}{Ajay Jain},
  {and} \bibinfo{person}{Pieter Abbeel}.} \bibinfo{year}{2020}\natexlab{}.
\newblock \showarticletitle{Denoising Diffusion Probabilistic Models}. In
  \bibinfo{booktitle}{\emph{Advances in Neural Information Processing
  Systems}}, \bibfield{editor}{\bibinfo{person}{H.~Larochelle},
  \bibinfo{person}{M.~Ranzato}, \bibinfo{person}{R.~Hadsell},
  \bibinfo{person}{M.~F. Balcan}, {and} \bibinfo{person}{H.~Lin}} (Eds.),
  Vol.~\bibinfo{volume}{33}. \bibinfo{publisher}{Curran Associates, Inc.},
  \bibinfo{address}{Red Hook, NY, USA}, \bibinfo{pages}{6840--6851}.
\newblock
\urldef\tempurl%
\url{https://proceedings.neurips.cc/paper/2020/hash/4c5bcfec8584af0d967f1ab10179ca4b-Abstract.html}
\showURL{%
\tempurl}


\bibitem[Huang et~al\mbox{.}(2025)]%
        {huang2025t2icompbenchpp}
\bibfield{author}{\bibinfo{person}{Kaiyi Huang}, \bibinfo{person}{Chengqi
  Duan}, \bibinfo{person}{Kaiyue Sun}, \bibinfo{person}{Enze Xie},
  \bibinfo{person}{Zhenguo Li}, {and} \bibinfo{person}{Xihui Liu}.}
  \bibinfo{year}{2025}\natexlab{}.
\newblock \showarticletitle{{T2I-CompBench}++: An Enhanced and Comprehensive
  Benchmark for Compositional Text-to-Image Generation}.
\newblock \bibinfo{journal}{\emph{IEEE Transactions on Pattern Analysis and
  Machine Intelligence}} \bibinfo{volume}{47}, \bibinfo{number}{5}
  (\bibinfo{year}{2025}), \bibinfo{pages}{3563--3579}.
\newblock
\href{https://doi.org/10.1109/TPAMI.2025.3531907}{doi:\nolinkurl{10.1109/TPAMI.2025.3531907}}


\bibitem[Huang et~al\mbox{.}(2023)]%
        {huang2023t2icompbench}
\bibfield{author}{\bibinfo{person}{Kaiyi Huang}, \bibinfo{person}{Kaiyue Sun},
  \bibinfo{person}{Enze Xie}, \bibinfo{person}{Zhenguo Li}, {and}
  \bibinfo{person}{Xihui Liu}.} \bibinfo{year}{2023}\natexlab{}.
\newblock \showarticletitle{{T2I-CompBench}: A Comprehensive Benchmark for
  Open-World Compositional Text-to-Image Generation}. In
  \bibinfo{booktitle}{\emph{Advances in Neural Information Processing
  Systems}}, \bibfield{editor}{\bibinfo{person}{A.~Oh},
  \bibinfo{person}{T.~Naumann}, \bibinfo{person}{A.~Globerson},
  \bibinfo{person}{K.~Saenko}, \bibinfo{person}{M.~Hardt}, {and}
  \bibinfo{person}{S.~Levine}} (Eds.), Vol.~\bibinfo{volume}{36}.
  \bibinfo{publisher}{Curran Associates, Inc.}, \bibinfo{address}{Red Hook, NY,
  USA}, \bibinfo{pages}{78723--78747}.
\newblock
\href{https://doi.org/10.52202/075280-3443}{doi:\nolinkurl{10.52202/075280-3443}}


\bibitem[Kwon et~al\mbox{.}(2023)]%
        {kwon2023semanticlatent}
\bibfield{author}{\bibinfo{person}{Mingi Kwon}, \bibinfo{person}{Jaeseok
  Jeong}, {and} \bibinfo{person}{Youngjung Uh}.}
  \bibinfo{year}{2023}\natexlab{}.
\newblock \showarticletitle{Diffusion Models Already Have a Semantic Latent
  Space}. In \bibinfo{booktitle}{\emph{The Eleventh International Conference on
  Learning Representations}}. \bibinfo{publisher}{OpenReview.net},
  \bibinfo{address}{Kigali, Rwanda}.
\newblock
\urldef\tempurl%
\url{https://openreview.net/forum?id=pd1P2eUBVfq}
\showURL{%
\tempurl}


\bibitem[Lei et~al\mbox{.}(2024)]%
        {lei2024ddpath}
\bibfield{author}{\bibinfo{person}{Yiming Lei}, \bibinfo{person}{Zilong Li},
  \bibinfo{person}{Junping Zhang}, {and} \bibinfo{person}{Hongming Shan}.}
  \bibinfo{year}{2024}\natexlab{}.
\newblock \showarticletitle{Denoising Diffusion Path: Attribution Noise
  Reduction with An Auxiliary Diffusion Model}. In
  \bibinfo{booktitle}{\emph{Advances in Neural Information Processing
  Systems}}, \bibfield{editor}{\bibinfo{person}{A.~Globerson},
  \bibinfo{person}{L.~Mackey}, \bibinfo{person}{D.~Belgrave},
  \bibinfo{person}{A.~Fan}, \bibinfo{person}{U.~Paquet},
  \bibinfo{person}{J.~Tomczak}, {and} \bibinfo{person}{C.~Zhang}} (Eds.),
  Vol.~\bibinfo{volume}{37}. \bibinfo{publisher}{Curran Associates, Inc.},
  \bibinfo{address}{Red Hook, NY, USA}, \bibinfo{pages}{54003--54025}.
\newblock
\href{https://doi.org/10.52202/079017-1710}{doi:\nolinkurl{10.52202/079017-1710}}


\bibitem[Li et~al\mbox{.}(2022)]%
        {li2022blip}
\bibfield{author}{\bibinfo{person}{Junnan Li}, \bibinfo{person}{Dongxu Li},
  \bibinfo{person}{Caiming Xiong}, {and} \bibinfo{person}{Steven Hoi}.}
  \bibinfo{year}{2022}\natexlab{}.
\newblock \showarticletitle{{BLIP}: Bootstrapping Language-Image Pre-training
  for Unified Vision-Language Understanding and Generation}. In
  \bibinfo{booktitle}{\emph{Proceedings of the 39th International Conference on
  Machine Learning}} \emph{(\bibinfo{series}{Proceedings of Machine Learning
  Research}, Vol.~\bibinfo{volume}{162})},
  \bibfield{editor}{\bibinfo{person}{Kamalika Chaudhuri},
  \bibinfo{person}{Stefanie Jegelka}, \bibinfo{person}{Le~Song},
  \bibinfo{person}{Csaba Szepesvari}, \bibinfo{person}{Gang Niu}, {and}
  \bibinfo{person}{Sivan Sabato}} (Eds.). \bibinfo{publisher}{PMLR},
  \bibinfo{address}{Baltimore, Maryland, USA}, \bibinfo{pages}{12888--12900}.
\newblock
\urldef\tempurl%
\url{https://proceedings.mlr.press/v162/li22n.html}
\showURL{%
\tempurl}


\bibitem[Lipman et~al\mbox{.}(2023)]%
        {lipman2023flow}
\bibfield{author}{\bibinfo{person}{Yaron Lipman}, \bibinfo{person}{Ricky T.~Q.
  Chen}, \bibinfo{person}{Heli Ben-Hamu}, \bibinfo{person}{Maximilian Nickel},
  {and} \bibinfo{person}{Matthew Le}.} \bibinfo{year}{2023}\natexlab{}.
\newblock \showarticletitle{Flow Matching for Generative Modeling}. In
  \bibinfo{booktitle}{\emph{The Eleventh International Conference on Learning
  Representations}}. \bibinfo{publisher}{OpenReview.net},
  \bibinfo{address}{Kigali, Rwanda}.
\newblock
\urldef\tempurl%
\url{https://openreview.net/forum?id=PqvMRDCJT9t}
\showURL{%
\tempurl}


\bibitem[Liu et~al\mbox{.}(2023)]%
        {liu2023rectifiedflow}
\bibfield{author}{\bibinfo{person}{Xingchao Liu}, \bibinfo{person}{Chengyue
  Gong}, {and} \bibinfo{person}{Qiang Liu}.} \bibinfo{year}{2023}\natexlab{}.
\newblock \showarticletitle{Flow Straight and Fast: Learning to Generate and
  Transfer Data with Rectified Flow}. In \bibinfo{booktitle}{\emph{The Eleventh
  International Conference on Learning Representations}}.
  \bibinfo{publisher}{OpenReview.net}, \bibinfo{address}{Kigali, Rwanda}.
\newblock
\urldef\tempurl%
\url{https://openreview.net/forum?id=XVjTT1nw5z}
\showURL{%
\tempurl}


\bibitem[Makhzani and Frey(2014)]%
        {makhzani2014ksparse}
\bibfield{author}{\bibinfo{person}{Alireza Makhzani} {and}
  \bibinfo{person}{Brendan Frey}.} \bibinfo{year}{2014}\natexlab{}.
\newblock \showarticletitle{k-Sparse Autoencoders}. In
  \bibinfo{booktitle}{\emph{The Second International Conference on Learning
  Representations}}. \bibinfo{publisher}{OpenReview.net},
  \bibinfo{address}{Banff, Canada}.
\newblock
\urldef\tempurl%
\url{https://openreview.net/forum?id=QDm4QXNOsuQVE}
\showURL{%
\tempurl}


\bibitem[Nguyen et~al\mbox{.}(2025)]%
        {nguyen2025cad}
\bibfield{author}{\bibinfo{person}{Hung-Quang Nguyen}, \bibinfo{person}{Hoang
  Phan}, {and} \bibinfo{person}{Khoa~D. Doan}.}
  \bibinfo{year}{2025}\natexlab{}.
\newblock \showarticletitle{Unveiling Concept Attribution in Diffusion Models}.
  In \bibinfo{booktitle}{\emph{Advances in Neural Information Processing
  Systems}}, \bibfield{editor}{\bibinfo{person}{D.~Belgrave},
  \bibinfo{person}{C.~Zhang}, \bibinfo{person}{H.~Lin},
  \bibinfo{person}{R.~Pascanu}, \bibinfo{person}{P.~Koniusz},
  \bibinfo{person}{M.~Ghassemi}, {and} \bibinfo{person}{N.~Chen}} (Eds.),
  Vol.~\bibinfo{volume}{38}. \bibinfo{publisher}{Curran Associates, Inc.},
  \bibinfo{address}{Red Hook, NY, USA}, \bibinfo{pages}{28596--28622}.
\newblock
\href{https://doi.org/10.52202/085713-0962}{doi:\nolinkurl{10.52202/085713-0962}}


\bibitem[Nichol et~al\mbox{.}(2022)]%
        {nichol2022glide}
\bibfield{author}{\bibinfo{person}{Alexander~Quinn Nichol},
  \bibinfo{person}{Prafulla Dhariwal}, \bibinfo{person}{Aditya Ramesh},
  \bibinfo{person}{Pranav Shyam}, \bibinfo{person}{Pamela Mishkin},
  \bibinfo{person}{Bob McGrew}, \bibinfo{person}{Ilya Sutskever}, {and}
  \bibinfo{person}{Mark Chen}.} \bibinfo{year}{2022}\natexlab{}.
\newblock \showarticletitle{{GLIDE}: Towards Photorealistic Image Generation
  and Editing with Text-Guided Diffusion Models}. In
  \bibinfo{booktitle}{\emph{Proceedings of the 39th International Conference on
  Machine Learning}} \emph{(\bibinfo{series}{Proceedings of Machine Learning
  Research}, Vol.~\bibinfo{volume}{162})},
  \bibfield{editor}{\bibinfo{person}{Kamalika Chaudhuri},
  \bibinfo{person}{Stefanie Jegelka}, \bibinfo{person}{Le~Song},
  \bibinfo{person}{Csaba Szepesvari}, \bibinfo{person}{Gang Niu}, {and}
  \bibinfo{person}{Sivan Sabato}} (Eds.). \bibinfo{publisher}{PMLR},
  \bibinfo{address}{Baltimore, Maryland, USA}, \bibinfo{pages}{16784--16804}.
\newblock
\urldef\tempurl%
\url{https://proceedings.mlr.press/v162/nichol22a.html}
\showURL{%
\tempurl}


\bibitem[Park and Jang(2025)]%
        {park2025i2am}
\bibfield{author}{\bibinfo{person}{Junseo Park} {and} \bibinfo{person}{Hyeryung
  Jang}.} \bibinfo{year}{2025}\natexlab{}.
\newblock \showarticletitle{$\text{I}^2\text{AM}$: Interpreting Image-to-Image
  Latent Diffusion Models via Bi-Attribution Maps}. In
  \bibinfo{booktitle}{\emph{The Thirteenth International Conference on Learning
  Representations}}. \bibinfo{publisher}{OpenReview.net},
  \bibinfo{address}{Singapore}.
\newblock
\urldef\tempurl%
\url{https://openreview.net/forum?id=bBNUiErs26}
\showURL{%
\tempurl}


\bibitem[Park et~al\mbox{.}(2025)]%
        {park2025headrelevance}
\bibfield{author}{\bibinfo{person}{Jungwon Park}, \bibinfo{person}{Jungmin Ko},
  \bibinfo{person}{Dongnam Byun}, \bibinfo{person}{Jangwon Suh}, {and}
  \bibinfo{person}{Wonjong Rhee}.} \bibinfo{year}{2025}\natexlab{}.
\newblock \showarticletitle{Cross-Attention Head Position Patterns Can Align
  with Human Visual Concepts in Text-to-Image Generative Models}. In
  \bibinfo{booktitle}{\emph{The Thirteenth International Conference on Learning
  Representations}}. \bibinfo{publisher}{OpenReview.net},
  \bibinfo{address}{Singapore}.
\newblock
\urldef\tempurl%
\url{https://openreview.net/forum?id=1vggIT5vvj}
\showURL{%
\tempurl}


\bibitem[Podell et~al\mbox{.}(2023)]%
        {podell2023sdxl}
\bibfield{author}{\bibinfo{person}{Dustin Podell}, \bibinfo{person}{Zion
  English}, \bibinfo{person}{Kyle Lacey}, \bibinfo{person}{Andreas Blattmann},
  \bibinfo{person}{Tim Dockhorn}, \bibinfo{person}{Jonas M{\"u}ller},
  \bibinfo{person}{Joe Penna}, {and} \bibinfo{person}{Robin Rombach}.}
  \bibinfo{year}{2023}\natexlab{}.
\newblock \bibinfo{title}{{SDXL}: Improving Latent Diffusion Models for
  High-Resolution Image Synthesis}.
\newblock
\showeprint[arxiv]{2307.01952}~[cs.CV]
\href{https://doi.org/10.48550/arXiv.2307.01952}{doi:\nolinkurl{10.48550/arXiv.2307.01952}}


\bibitem[Radford et~al\mbox{.}(2021)]%
        {radford2021clip}
\bibfield{author}{\bibinfo{person}{Alec Radford}, \bibinfo{person}{Jong~Wook
  Kim}, \bibinfo{person}{Chris Hallacy}, \bibinfo{person}{Aditya Ramesh},
  \bibinfo{person}{Gabriel Goh}, \bibinfo{person}{Sandhini Agarwal},
  \bibinfo{person}{Girish Sastry}, \bibinfo{person}{Amanda Askell},
  \bibinfo{person}{Pamela Mishkin}, \bibinfo{person}{Jack Clark},
  \bibinfo{person}{Gretchen Krueger}, {and} \bibinfo{person}{Ilya Sutskever}.}
  \bibinfo{year}{2021}\natexlab{}.
\newblock \showarticletitle{Learning Transferable Visual Models From Natural
  Language Supervision}. In \bibinfo{booktitle}{\emph{Proceedings of the 38th
  International Conference on Machine Learning}}
  \emph{(\bibinfo{series}{Proceedings of Machine Learning Research},
  Vol.~\bibinfo{volume}{139})}, \bibfield{editor}{\bibinfo{person}{Marina
  Meila} {and} \bibinfo{person}{Tong Zhang}} (Eds.). \bibinfo{publisher}{PMLR},
  \bibinfo{address}{Virtual}, \bibinfo{pages}{8748--8763}.
\newblock
\urldef\tempurl%
\url{https://proceedings.mlr.press/v139/radford21a.html}
\showURL{%
\tempurl}


\bibitem[Rombach et~al\mbox{.}(2022)]%
        {rombach2022latent}
\bibfield{author}{\bibinfo{person}{Robin Rombach}, \bibinfo{person}{Andreas
  Blattmann}, \bibinfo{person}{Dominik Lorenz}, \bibinfo{person}{Patrick
  Esser}, {and} \bibinfo{person}{Bj{\"o}rn Ommer}.}
  \bibinfo{year}{2022}\natexlab{}.
\newblock \showarticletitle{High-Resolution Image Synthesis with Latent
  Diffusion Models}. In \bibinfo{booktitle}{\emph{Proceedings of the IEEE/CVF
  Conference on Computer Vision and Pattern Recognition}}.
  \bibinfo{publisher}{IEEE Computer Society}, \bibinfo{address}{Los Alamitos,
  CA, USA}, \bibinfo{pages}{10684--10695}.
\newblock
\href{https://doi.org/10.1109/CVPR52688.2022.01042}{doi:\nolinkurl{10.1109/CVPR52688.2022.01042}}


\bibitem[Saharia et~al\mbox{.}(2022)]%
        {saharia2022photorealistic}
\bibfield{author}{\bibinfo{person}{Chitwan Saharia}, \bibinfo{person}{William
  Chan}, \bibinfo{person}{Saurabh Saxena}, \bibinfo{person}{Lala Li},
  \bibinfo{person}{Jay Whang}, \bibinfo{person}{Emily Denton},
  \bibinfo{person}{Seyed Kamyar~Seyed Ghasemipour},
  \bibinfo{person}{Burcu~Karagol Ayan}, \bibinfo{person}{S.~Sara Mahdavi},
  \bibinfo{person}{Rapha~Gontijo Lopes}, \bibinfo{person}{Tim Salimans},
  \bibinfo{person}{Jonathan Ho}, \bibinfo{person}{David~J. Fleet}, {and}
  \bibinfo{person}{Mohammad Norouzi}.} \bibinfo{year}{2022}\natexlab{}.
\newblock \showarticletitle{Photorealistic Text-to-Image Diffusion Models with
  Deep Language Understanding}. In \bibinfo{booktitle}{\emph{Advances in Neural
  Information Processing Systems}},
  \bibfield{editor}{\bibinfo{person}{S.~Koyejo}, \bibinfo{person}{S.~Mohamed},
  \bibinfo{person}{A.~Agarwal}, \bibinfo{person}{D.~Belgrave},
  \bibinfo{person}{K.~Cho}, {and} \bibinfo{person}{A.~Oh}} (Eds.),
  Vol.~\bibinfo{volume}{35}. \bibinfo{publisher}{Curran Associates, Inc.},
  \bibinfo{address}{Red Hook, NY, USA}, \bibinfo{pages}{36479--36494}.
\newblock
\href{https://doi.org/10.52202/068431-2643}{doi:\nolinkurl{10.52202/068431-2643}}


\bibitem[Song et~al\mbox{.}(2021)]%
        {song2021ddim}
\bibfield{author}{\bibinfo{person}{Jiaming Song}, \bibinfo{person}{Chenlin
  Meng}, {and} \bibinfo{person}{Stefano Ermon}.}
  \bibinfo{year}{2021}\natexlab{}.
\newblock \showarticletitle{Denoising Diffusion Implicit Models}. In
  \bibinfo{booktitle}{\emph{The Ninth International Conference on Learning
  Representations}}. \bibinfo{publisher}{OpenReview.net},
  \bibinfo{address}{Virtual}.
\newblock
\urldef\tempurl%
\url{https://openreview.net/forum?id=St1giarCHLP}
\showURL{%
\tempurl}


\bibitem[{Stability AI}(2024)]%
        {stabilityai2024sd35}
\bibfield{author}{\bibinfo{person}{{Stability AI}}.}
  \bibinfo{year}{2024}\natexlab{}.
\newblock \bibinfo{title}{Introducing Stable Diffusion 3.5}.
\newblock
\urldef\tempurl%
\url{https://stability.ai/news/introducing-stable-diffusion-3-5}
\showURL{%
\tempurl}


\bibitem[Surkov et~al\mbox{.}(2025)]%
        {surkov2025onestep}
\bibfield{author}{\bibinfo{person}{Viacheslav Surkov}, \bibinfo{person}{Chris
  Wendler}, \bibinfo{person}{Antonio Mari}, \bibinfo{person}{Mikhail Terekhov},
  \bibinfo{person}{Justin Deschenaux}, \bibinfo{person}{Robert West},
  \bibinfo{person}{Caglar Gulcehre}, {and} \bibinfo{person}{David Bau}.}
  \bibinfo{year}{2025}\natexlab{}.
\newblock \showarticletitle{One-Step is Enough: Sparse Autoencoders for
  Text-to-Image Diffusion Models}. In \bibinfo{booktitle}{\emph{Advances in
  Neural Information Processing Systems}},
  \bibfield{editor}{\bibinfo{person}{D.~Belgrave}, \bibinfo{person}{C.~Zhang},
  \bibinfo{person}{H.~Lin}, \bibinfo{person}{R.~Pascanu},
  \bibinfo{person}{P.~Koniusz}, \bibinfo{person}{M.~Ghassemi}, {and}
  \bibinfo{person}{N.~Chen}} (Eds.), Vol.~\bibinfo{volume}{38}.
  \bibinfo{publisher}{Curran Associates, Inc.}, \bibinfo{address}{Red Hook, NY,
  USA}, \bibinfo{pages}{99956--100027}.
\newblock
\href{https://doi.org/10.52202/085713-3342}{doi:\nolinkurl{10.52202/085713-3342}}


\bibitem[Tang et~al\mbox{.}(2023)]%
        {tang2023daam}
\bibfield{author}{\bibinfo{person}{Raphael Tang}, \bibinfo{person}{Linqing
  Liu}, \bibinfo{person}{Akshat Pandey}, \bibinfo{person}{Zhiying Jiang},
  \bibinfo{person}{Gefei Yang}, \bibinfo{person}{Karun Kumar},
  \bibinfo{person}{Pontus Stenetorp}, \bibinfo{person}{Jimmy Lin}, {and}
  \bibinfo{person}{Ferhan Ture}.} \bibinfo{year}{2023}\natexlab{}.
\newblock \showarticletitle{What the {{DAAM}}: Interpreting Stable Diffusion
  Using Cross Attention}. In \bibinfo{booktitle}{\emph{Proceedings of the 61st
  Annual Meeting of the Association for Computational Linguistics (Volume 1:
  Long Papers)}}. \bibinfo{publisher}{Association for Computational
  Linguistics}, \bibinfo{address}{Toronto, Canada},
  \bibinfo{pages}{5644--5659}.
\newblock
\href{https://doi.org/10.18653/v1/2023.acl-long.310}{doi:\nolinkurl{10.18653/v1/2023.acl-long.310}}


\bibitem[Templeton et~al\mbox{.}(2024)]%
        {templeton2024scaling}
\bibfield{author}{\bibinfo{person}{Adly Templeton}, \bibinfo{person}{Tom
  Conerly}, \bibinfo{person}{Jonathan Marcus}, \bibinfo{person}{Jack Lindsey},
  \bibinfo{person}{Trenton Bricken}, \bibinfo{person}{Brian Chen},
  \bibinfo{person}{Adam Pearce}, \bibinfo{person}{Craig Citro},
  \bibinfo{person}{Emmanuel Ameisen}, \bibinfo{person}{Andy Jones},
  \bibinfo{person}{Hoagy Cunningham}, \bibinfo{person}{Nicholas~L. Turner},
  \bibinfo{person}{Callum McDougall}, \bibinfo{person}{Monte MacDiarmid},
  \bibinfo{person}{Alex Tamkin}, \bibinfo{person}{Esin Durmus},
  \bibinfo{person}{Tristan Hume}, \bibinfo{person}{Francesco Mosconi},
  \bibinfo{person}{C.~Daniel Freeman}, \bibinfo{person}{Theodore~R. Sumers},
  \bibinfo{person}{Edward Rees}, \bibinfo{person}{Joshua Batson},
  \bibinfo{person}{Adam Jermyn}, \bibinfo{person}{Shan Carter},
  \bibinfo{person}{Chris Olah}, {and} \bibinfo{person}{Tom Henighan}.}
  \bibinfo{year}{2024}\natexlab{}.
\newblock \bibinfo{title}{Scaling Monosemanticity: Extracting Interpretable
  Features from Claude 3 Sonnet}.
\newblock \bibinfo{howpublished}{Transformer Circuits Thread}.
\newblock
\urldef\tempurl%
\url{https://transformer-circuits.pub/2024/scaling-monosemanticity/index.html}
\showURL{%
\tempurl}


\bibitem[Tinaz et~al\mbox{.}(2025)]%
        {tinaz2025emergence}
\bibfield{author}{\bibinfo{person}{Berk Tinaz}, \bibinfo{person}{Zalan Fabian},
  {and} \bibinfo{person}{Mahdi Soltanolkotabi}.}
  \bibinfo{year}{2025}\natexlab{}.
\newblock \showarticletitle{Emergence and Evolution of Interpretable Concepts
  in Diffusion Models}. In \bibinfo{booktitle}{\emph{Advances in Neural
  Information Processing Systems}},
  \bibfield{editor}{\bibinfo{person}{D.~Belgrave}, \bibinfo{person}{C.~Zhang},
  \bibinfo{person}{H.~Lin}, \bibinfo{person}{R.~Pascanu},
  \bibinfo{person}{P.~Koniusz}, \bibinfo{person}{M.~Ghassemi}, {and}
  \bibinfo{person}{N.~Chen}} (Eds.), Vol.~\bibinfo{volume}{38}.
  \bibinfo{publisher}{Curran Associates, Inc.}, \bibinfo{address}{Red Hook, NY,
  USA}, \bibinfo{pages}{166943--166986}.
\newblock
\href{https://doi.org/10.52202/085713-5566}{doi:\nolinkurl{10.52202/085713-5566}}


\bibitem[Wu et~al\mbox{.}(2026)]%
        {wu2026promsa}
\bibfield{author}{\bibinfo{person}{ZhengXian Wu}, \bibinfo{person}{Hangrui Xu},
  \bibinfo{person}{Kai Shi}, \bibinfo{person}{Zhuohong Chen},
  \bibinfo{person}{Yunyao Yu}, \bibinfo{person}{Chuanrui Zhang},
  \bibinfo{person}{Zirui Liao}, \bibinfo{person}{Jun Yang},
  \bibinfo{person}{Zhenyu Yang}, \bibinfo{person}{Haonan Lu}, {and}
  \bibinfo{person}{Haoqian Wang}.} \bibinfo{year}{2026}\natexlab{}.
\newblock \bibinfo{title}{{ProMSA}: Progressive Multimodal Search Agents for
  Knowledge-Based Visual Question Answering}.
\newblock
\shownote{Accepted to ECCV 2026}.
\newblock
\showeprint[arxiv]{2606.27974}~[cs.CV]
\href{https://doi.org/10.48550/arXiv.2606.27974}{doi:\nolinkurl{10.48550/arXiv.2606.27974}}


\bibitem[Yuan et~al\mbox{.}(2025)]%
        {yuan2025dissecting}
\bibfield{author}{\bibinfo{person}{Yifan Yuan}, \bibinfo{person}{Guanqun Yang},
  \bibinfo{person}{James~Z. Wang}, \bibinfo{person}{Hui Zhang},
  \bibinfo{person}{Hongming Shan}, \bibinfo{person}{Fei-Yue Wang}, {and}
  \bibinfo{person}{Junping Zhang}.} \bibinfo{year}{2025}\natexlab{}.
\newblock \showarticletitle{Dissecting and Mitigating Semantic Discrepancy in
  Stable Diffusion for Image-to-Image Translation}.
\newblock \bibinfo{journal}{\emph{IEEE/CAA Journal of Automatica Sinica}}
  \bibinfo{volume}{12}, \bibinfo{number}{4} (\bibinfo{year}{2025}),
  \bibinfo{pages}{705--718}.
\newblock
\href{https://doi.org/10.1109/JAS.2024.124800}{doi:\nolinkurl{10.1109/JAS.2024.124800}}


\bibitem[Zhang and Wan(2025)]%
        {zhang2025rbind}
\bibfield{author}{\bibinfo{person}{Huixuan Zhang} {and}
  \bibinfo{person}{Xiaojun Wan}.} \bibinfo{year}{2025}\natexlab{}.
\newblock \showarticletitle{{R-Bind}: Unified Enhancement of Attribute and
  Relation Binding in Text-to-Image Diffusion Models}. In
  \bibinfo{booktitle}{\emph{Proceedings of the 2025 Conference on Empirical
  Methods in Natural Language Processing}}. \bibinfo{publisher}{Association for
  Computational Linguistics}, \bibinfo{address}{Suzhou, China},
  \bibinfo{pages}{6856--6870}.
\newblock
\href{https://doi.org/10.18653/v1/2025.emnlp-main.349}{doi:\nolinkurl{10.18653/v1/2025.emnlp-main.349}}


\end{thebibliography}

\def\maketitlesupplementary
   {
   \newpage
       \twocolumn[
        \centering
        \LARGE
        \vspace{0.5em}\textbf{Appendix} \\
        \vspace{2.0em}
       ] 
   }

\maketitlesupplementary

\appendix
\setcounter{page}{1}

This appendix provides supplementary details and analyses that complement the main text. Specifically, it expands the description of dataset construction, feature extraction, and the two-stage implementation pipeline; gives explicit definitions of all quantitative metrics; reports additional baseline comparisons and more qualitative examples; and includes sanity checks, ablation studies, and robustness analyses. Together, these materials are intended to improve reproducibility and to provide further evidence for the interpretability and intervention claims of the proposed framework.

\section*{Appendix Contents}
\begin{itemize}[leftmargin=1.5em]
    \item \hyperref[app:data]{Dataset Construction}
    \item \hyperref[app:algo]{Detailed Algorithm and Pipeline}
    \item \hyperref[app:metric]{Quantitative Metric Details}
    \item \hyperref[app:show]{More Qualitative Results}
    \item \hyperref[app:check]{Intervening in the Raw Feature Space versus in SAE Space}
    \item \hyperref[app:ablation]{Ablation Analyses}
    \item \hyperref[app:robust]{Robustness Analyses}
\end{itemize}

\section{Dataset Construction}
\label{app:data}

\subsection{Concept Vocabulary}
\label{app:data_vocab}

As stated in the main text, the training split covers 10 styles and 10 attribute terms, while the held-out test split follows the prompt template
\texttt{[color] [texture] [shape] [entity], [style]}.
Below, we specify the concept vocabulary explicitly.

\noindent{\bf{Training styles.}}
The style-training prompts follow the template
\texttt{[entity], [style]}.
The 10 styles used in the training split are
\begin{quote}
\texttt{abstract art},
\texttt{modern art},
\texttt{ink sketch},
\texttt{pixel art},
\texttt{cartoon style},
\texttt{oil painting},
\texttt{3D render},
\texttt{realistic photo},
\texttt{black and white photo}, and
\texttt{vintage photo}.
\end{quote}

These styles cover several distinct visual regimes. In particular, the vocabulary includes highly abstract styles such as \texttt{abstract art} and \texttt{modern art}, medium-constrained styles such as \texttt{3D render}, \texttt{oil painting}, \texttt{pixel art}, and \texttt{ink sketch}, as well as photography-related styles such as \texttt{realistic photo}, \texttt{black and white photo}, and \texttt{vintage photo}. This breadth is useful in our setting because it spans both high-level stylistic abstraction and visually concrete rendering constraints.

\noindent{\bf{Training attribute terms.}}
The attribute-training prompts follow the template
\texttt{[attribute] [entity]}.
The 10 attribute terms used in the training split are
\begin{quote}
\texttt{gold},
\texttt{pink},
\texttt{silver},
\texttt{metallic},
\texttt{rubber},
\texttt{wooden},
\texttt{glass},
\texttt{cylindrical},
\texttt{teardrop}, and
\texttt{pyramidal}.
\end{quote}

These terms can be grouped into three appearance-oriented categories. The color-related terms are \texttt{gold}, \texttt{pink}, and \texttt{silver}. The texture-related terms are \texttt{metallic}, \texttt{rubber}, \texttt{wooden}, and \texttt{glass}. The shape-related terms are \texttt{cylindrical}, \texttt{pyramidal}, and \texttt{teardrop}. Each attribute term appears 208 times in the training prompt file.

\noindent{\bf{Test-time color / texture / shape vocabulary.}}
The held-out test split uses a richer closed-set vocabulary over color, texture, and shape within the unified template
\texttt{[color] [texture] [shape] [entity], [style]}.

The color vocabulary is
\begin{quote}
\texttt{black},
\texttt{blue},
\texttt{brown},
\texttt{gold},
\texttt{green},
\texttt{orange},
\texttt{pink},
\texttt{purple},
\texttt{red},
\texttt{silver},
\texttt{white}, and
\texttt{yellow}.
\end{quote}

The texture vocabulary is
\begin{quote}
\texttt{fabric},
\texttt{fluffy},
\texttt{glass},
\texttt{leather},
\texttt{metallic},
\texttt{plastic},
\texttt{rubber}, and
\texttt{wooden}.
\end{quote}

The shape vocabulary is
\begin{quote}
\texttt{conical},
\texttt{cylindrical},
\texttt{diamond},
\texttt{pentagonal},
\texttt{pyramidal},
\texttt{rectangular},
\texttt{round},
\texttt{spherical},
\texttt{square},
\texttt{teardrop}, and
\texttt{triangular}.
\end{quote}

\noindent{\bf{Why these concepts.}}
The prompt files show directly that the training split covers 10 styles and 10 attribute terms, and that the test split follows the explicit protocol
\texttt{[color] [texture] [shape] [entity], [style]}.
Beyond these directly observable facts, the overall concept design follows a deliberate coverage strategy.

First, the style vocabulary spans both abstract and medium-specific conditions. Second, the appearance vocabulary focuses on color, texture, and shape, which are visually salient and also common sources of generation failure. Third, the training attribute space is intentionally compact, while the held-out test split expands this space into a richer closed-set composition over color, texture, and shape. This design indicates that the benchmark is intended to test controlled generalization under concept composition rather than simple memorization of training prompts.

\subsection{Entity Coverage and Split Characteristics}
\label{app:data_entity}

As summarized in the main text, the held-out test split contains prompts from more than 150 entity categories. Inspection of the prompt files further shows that the entity pool spans many semantic clusters rather than a narrow set of object nouns.

Representative categories include the following:
\begin{quote}
Animals: \texttt{dog}, \texttt{cat}, \texttt{elephant}, \texttt{lion}, \texttt{owl}, \texttt{whale}\\
Plants: \texttt{tree}, \texttt{flower}, \texttt{oak tree}, \texttt{tulip}, \texttt{rose}\\
People: \texttt{child}, \texttt{elderly person}, \texttt{doctor}, \texttt{teacher}, \texttt{tourist}
\end{quote}
as well as vehicles, buildings, furniture, electronics, clothing, accessories, tools, toys, food, books, sports equipment, instruments, household items, and scene-related categories.

\noindent{\bf{Coverage of semantically related objects.}}
The prompt files do not provide an explicit formal coverage rule such as WordNet-based sampling or embedding-based sampling. However, the observed vocabulary structure suggests that semantically related objects are mainly covered through manual enumeration within semantic clusters. For example, related animal categories such as 
\begin{quote}
\texttt{dog}, \texttt{cat}, \texttt{lion}, \texttt{fox}, and \texttt{bear} co-occur; furniture categories such as \texttt{chair}, \texttt{table}, \texttt{sofa}, \texttt{bed}, and \texttt{bookshelf} appear in groups; transportation categories such as \texttt{car}, \texttt{bus}, \texttt{truck}, \texttt{train}, \texttt{boat}, \texttt{airplane}, and \texttt{helicopter} are also grouped; and cultural objects such as \texttt{novel}, \texttt{dictionary}, \texttt{textbook}, \texttt{magazine}, \texttt{newspaper}, \texttt{journal}, and \texttt{notebook} 
\end{quote}
are jointly included. This clustered organization is important in our setting because object-dependent brittleness is most visible when the same concept is evaluated on semantically nearby content instances.

\noindent{\bf{Observed split properties and partial cleaning.}}
The style-training prompt file contains 208 normalized unique entities, with no exact duplicates after normalization. The attribute-training prompt file contains 208 prompts for each attribute term, but only 180 unique entities in each attribute block; the remaining 28 entries are repeated entities used to fill the block to 208 examples. In addition, the attribute prompt file appears to exclude many plural nouns, mass nouns, and some pattern-like terms. Taken together, these observations suggest that the dataset undergoes partial manual cleaning to remove entities that are awkward under direct attribute-prefix templates, although the resulting prompt space is not fully linguistically sanitized. We therefore view the dataset as a controlled prompt benchmark rather than as a natural-language benchmark.

\subsection{Prompt Generation Rules}
\label{app:data_prompt}

Although the style and attribute tasks do not use the same literal prompt string, they follow a unified high-level protocol: the entity content and the condition concept are explicitly separated, and the test split recombines them under a more compositional template.

\noindent{\bf{Style training prompts.}}
The style-training split uses the template
\texttt{[entity], [style]}.
It contains 208 entities and 10 styles, which results in 2080 prompts in total. In practice, this is close to a Cartesian product of entity and style.

\noindent{\bf{Attribute training prompts.}}
The attribute-training split uses the template
\texttt{[attribute] [entity]}.
It contains 10 attribute terms, with 208 prompts for each term, again yielding 2080 prompts in total. However, this is not a strict Cartesian product over unique entities, because each attribute block contains 180 unique entities together with 28 repeated fillers.

\noindent{\bf{Held-out joint test prompts.}}
The held-out test split uses the template
\texttt{[color] [texture] [shape] [entity], [style]}.
Here, the color, texture, and shape slots are optional, but each prompt contains at least one attribute term. The total size of the test split is 1780 prompts, and the 10 styles are evenly distributed, with 178 prompts for each style.

The number of attribute terms in each test prompt is distributed as follows: 565 prompts contain one attribute term, 599 prompts contain two attribute terms, and 616 prompts contain three attribute terms. A finer breakdown shows that there are 188 prompts with color only, 199 with texture only, 178 with shape only, 177 with color + texture, 214 with color + shape, 208 with texture + shape, and 616 with color + texture + shape.

\noindent{\bf{Relationship between style and attribute tasks.}}
The style and attribute tasks should therefore be understood as two task-specific subtemplates within one shared protocol rather than as a single literal prompt template. Style training uses
\texttt{[entity], [style]},
attribute training uses
\texttt{[attribute] [entity]},
and the held-out test split combines both content and condition cues through
\texttt{[color] [texture] [shape] [entity], [style]}.
This design explicitly separates content from condition during training while reserving richer compositional combinations for final evaluation.

\noindent{\bf{Practical implication for our setting.}}
This dataset design is particularly suitable for analyzing object-dependent concept brittleness. Because the entity token is explicitly exposed and the condition token or tokens are controlled independently, one can compare prompts that differ primarily in content while keeping the target concept fixed. One can also test whether a sparse prior learned from a simplified training condition space transfers to a more compositional held-out test space.

\noindent{\bf{Representative prompt examples.}}
To make the construction of the dataset more concrete, we provide a small set of representative prompts drawn from the training and held-out splits. These examples are not intended to be exhaustive. Rather, they are selected to illustrate the grammatical form, semantic coverage, and compositional structure of the benchmark.

For the style-training split, prompts follow the simple template \texttt{[entity], [style]}. Representative examples include:
\begin{quote}
\texttt{"a dog, abstract art"}\\
\texttt{"an elephant, abstract art"}\\
\texttt{"a car, abstract art"}\\
\texttt{"a dog, ink sketch"}\\
\texttt{"an owl, realistic photo"}
\end{quote}

These examples show that the same entity space is reused across different styles and spans multiple semantic categories, including animals, vehicles, and everyday objects. This design makes it possible to examine whether the same target concept is expressed consistently when the object token changes.

For the held-out test split, prompts follow the richer compositional template \texttt{[color] [texture] [shape] [entity], [style]}. Representative examples include:
\begin{quote}
\texttt{"a gold leather dog, abstract art"}\\
\texttt{"a white car, cartoon style"}\\
\texttt{"a wooden watch, realistic photo"}\\
\texttt{"a leather horse, ink sketch"}\\
\texttt{"a red fabric triangular pizza, vintage photo"}\\
\texttt{"a pink glass cylindrical traffic light, 3D render"}
\end{quote}

These examples illustrate two important properties of the benchmark. First, the test split does not merely replace the object token under a fixed style condition, but composes multiple controllable appearance cues, including color, material or texture, and shape, within a unified prompt structure. Second, the entity vocabulary is intentionally broad and covers not only common animals and artifacts, but also people, food items, and small man-made objects. This diversity makes the benchmark suitable for evaluating whether a method generalizes beyond a narrow object domain and remains stable under more complex concept combinations.


\section{Detailed Algorithm and Pipeline}
\label{app:algo}

Algorithms~\ref{alg:stage1} and~\ref{alg:stage2} summarize the full two-stage procedure used in our implementation. In this appendix, we further clarify the practical feature-extraction protocol, the training details of the timestep-wise SAE bank, the construction of class prototypes, and the behavior of the inference-time intervention.

\subsection{Stage I: Sparse Prior Construction}
Given a reference set $\mathcal{D}_{\mathrm{ref}}=\{(x_i,p_i,c_i)\}_{i=1}^{N}$, we first extract a denoising feature tensor from each timestep and train one SAE for each timestep. Concretely, for every $t\in\{0,\dots,T-1\}$, we collect tokenized features
\[
X_i^{(t)} \in \mathbb{R}^{P\times D},
\]
compute timestep-specific normalization statistics $(\mu^{(t)},\sigma^{(t)})$, normalize all token features with z-score normalization, and train a dedicated Top-$K$ SAE
\[
\bigl(\mathrm{Enc}^{(t)},\mathrm{Dec}^{(t)}\bigr).
\]
The resulting token-level sparse code is denoted by
\[
Z_i^{(t)}=\mathrm{Enc}^{(t)}\!\left(\frac{X_i^{(t)}-\mu^{(t)}}{\sigma^{(t)}+\varepsilon}\right)\in\mathbb{R}^{P\times K_d}.
\]
For sample-level comparison, we apply mean pooling over tokens and obtain
\[
s_i^{(t)}=\frac{1}{P}\sum_{p=1}^{P} Z_i^{(t)}(p)\in\mathbb{R}^{K_d}.
\]

For each concept class $c$, we then estimate the class center and class dispersion in pooled SAE space. In the implementation used for prototype construction, we measure the deviation of each pooled code from the class-consistent sparse profile with the normalized max-deviation score
\[
d_i^{(t,c)}
=
\left\|
\frac{s_i^{(t)}-\mu_c^{(t)}}{\sigma_c^{(t)}+\varepsilon}
\right\|_\infty ,
\]
where token aggregation is performed by mean pooling. We then retain the reliable subset as
\[
I_{c,\mathrm{keep}}^{(t)}
=
\left\{
i \in I_c \;\middle|\; d_i^{(t,c)} \le \tau
\right\}.
\]
Samples satisfying this criterion are treated as prototype-consistent samples, where the default threshold is $\tau = 1.0$. In practice, this filtering step removes roughly one third of the initial class members and retains a more stable subset for prototype estimation. We then construct the class prototype in token-level SAE space:
\[
R_c^{(t)}(p)=
\frac{1}{|I_{c,\mathrm{keep}}^{(t)}|}
\sum_{i\in I_{c,\mathrm{keep}}^{(t)}} Z_i^{(t)}(p),
\qquad p=1,\dots,P.
\]
Thus, each timestep and each class are associated with a token-level sparse prior
\[
R_c^{(t)}\in\mathbb{R}^{P\times K_d}.
\]

\begin{algorithm}[!htbp]
\caption{Stage I: Sparse Prior Construction in SAE Space}
\label{alg:stage1}
\begin{algorithmic}[1]
\Require Reference dataset $\mathcal{D}_{\mathrm{ref}}=\{(x_i,p_i,c_i)\}_{i=1}^{N}$, backbone predictor $\Phi_\theta$, timesteps $\{0,\dots,T-1\}$, dictionary width $K_d$, sparsity level $K$, threshold $\tau$
\Ensure SAE bank $\{(\mathrm{Enc}^{(t)},\mathrm{Dec}^{(t)})\}$, normalization statistics $\{(\mu^{(t)},\sigma^{(t)})\}$, sparse priors $\{R_c^{(t)}\}$

\State $\mathcal{B}_{\mathrm{SAE}},\mathcal{B}_{\mathrm{norm}},\mathcal{B}_{\mathrm{prior}} \gets \emptyset$
\For{$t=0$ to $T-1$}
    \State $X_i^{(t)} \gets \mathrm{Tokenize}(\Phi_\theta(\cdot,p_i,t)),\ \forall i\in\{1,\dots,N\}$
    \State $\mathcal{S}^{(t)} \gets \{x_{i,t,p}\mid i=1,\dots,N,\ p=1,\dots,P\}$
    \State $\mu^{(t)} \gets \mathrm{Mean}(\mathcal{S}^{(t)}),\ \sigma^{(t)} \gets \mathrm{Std}(\mathcal{S}^{(t)})$
    \State $(\mathrm{Enc}^{(t)},\mathrm{Dec}^{(t)}) \gets \mathrm{TrainSAE}\!\left(\dfrac{\mathcal{S}^{(t)}-\mu^{(t)}}{\sigma^{(t)}+\varepsilon}, K_d, K\right)$
    \Comment{Train timestep-specific SAE}
    \State $Z_i^{(t)} \gets \mathrm{Enc}^{(t)}\!\left(\dfrac{X_i^{(t)}-\mu^{(t)}}{\sigma^{(t)}+\varepsilon}\right),\ \forall i$
    \State $s_i^{(t)} \gets \dfrac{1}{P}\sum_{p=1}^{P} Z_i^{(t)}(p),\ \forall i$
    \ForAll{class $c$}
        \State $I_c \gets \{i\mid c_i=c\}$
        \State $\mu_c^{(t)} \gets \dfrac{1}{|I_c|}\sum_{i\in I_c}s_i^{(t)}$
        \State $\sigma_c^{(t)} \gets \sqrt{\dfrac{1}{|I_c|}\sum_{i\in I_c}(s_i^{(t)}-\mu_c^{(t)})^2}$
        \State $I_{c,\mathrm{keep}}^{(t)} \gets \left\{ i\in I_c\;\middle|\;
        \left\|\dfrac{s_i^{(t)}-\mu_c^{(t)}}{\sigma_c^{(t)}+\varepsilon}\right\|_\infty \le \tau \right\}$
        \Comment{Filter abnormal samples in pooled SAE space}
        \State $R_c^{(t)}(p) \gets \dfrac{1}{|I_{c,\mathrm{keep}}^{(t)}|}\sum_{i\in I_{c,\mathrm{keep}}^{(t)}} Z_i^{(t)}(p),\ \forall p\in\{1,\dots,P\}$
    \EndFor
\EndFor
\State \Return $\{(\mathrm{Enc}^{(t)},\mathrm{Dec}^{(t)}),(\mu^{(t)},\sigma^{(t)}),\{R_c^{(t)}\}_c\}_{t=0}^{T-1}$
\end{algorithmic}
\end{algorithm}

\subsection{Stage II: Prior-Guided Sparse Denoising Intervention}
During inference, we run the original sampler and intervene only at the selected timesteps $\mathcal{T}_{\mathrm{guide}}$. For the current denoising feature $X_{\mathrm{test}}^{(t)}$, we first normalize it with the saved timestep-specific statistics, encode it into sparse space, and linearly interpolate it with the target prototype:
\[
\bar Z^{(t)}=(1-\alpha) Z_{\mathrm{test}}^{(t)}+\alpha R_{c^\star}^{(t)}.
\]
The interpolated sparse code is then decoded back into the original feature space, unnormalized, and converted into the feature tensor required by the sampler update. Importantly, this intervention is inserted \emph{before} the current sampler state update is written back, while the intervention target itself is the step-update delta of the current denoising step. Therefore, the method does not modify an arbitrary hidden state. Instead, it directly adjusts the update quantity used by the sampler at that timestep.

\begin{algorithm}[t]
\caption{Stage II: Prior-Guided Sparse Denoising Intervention}
\label{alg:stage2}
\begin{algorithmic}[1]
\Require Initial latent $z_T$, prompt $p$, target class $c^\star$, guidance steps $\mathcal{T}_{\mathrm{guide}}\subseteq\{0,\dots,T-1\}$, interpolation weight $\alpha$, backbone predictor $\Phi_\theta$, sampler $\{U_t\}_{t=0}^{T-1}$, SAE bank $\{(\mathrm{Enc}^{(t)},\mathrm{Dec}^{(t)})\}$, normalization statistics $\{(\mu^{(t)},\sigma^{(t)})\}$, sparse priors $\{R_c^{(t)}\}$
\Ensure Final image $x_0$

\State $z \gets z_T$

\For{$t=T-1$ downto $0$}
    \State $\Delta^{(t)} \gets \Phi_\theta(z,p,t),\quad X_{\mathrm{test}}^{(t)} \gets \mathrm{Tokenize}(\Delta^{(t)})$
    \If{$t \in \mathcal{T}_{\mathrm{guide}}$}
        \State $\widetilde{X}_{\mathrm{test}}^{(t)} \gets \dfrac{X_{\mathrm{test}}^{(t)}-\mu^{(t)}}{\sigma^{(t)}+\varepsilon}$
        \State $Z_{\mathrm{test}}^{(t)} \gets \mathrm{Enc}^{(t)}(\widetilde{X}_{\mathrm{test}}^{(t)})$
        \State $\bar Z^{(t)} \gets (1-\alpha) Z_{\mathrm{test}}^{(t)} + \alpha R_{c^\star}^{(t)}$
        \State $\widehat{X}^{(t)} \gets \mathrm{Dec}^{(t)}(\bar Z^{(t)})$
        \State $X_{\mathrm{guided}}^{(t)} \gets \widehat{X}^{(t)} \odot (\sigma^{(t)}+\varepsilon) + \mu^{(t)}$
        \State $\Delta_{\mathrm{guided}}^{(t)} \gets \mathrm{UnTokenize}(X_{\mathrm{guided}}^{(t)})$
        \Comment{Inject target-class sparse prior at selected steps}
    \Else
        \State $\Delta_{\mathrm{guided}}^{(t)} \gets \Delta^{(t)}$
    \EndIf
    \State $z \gets U_t(z,\Delta_{\mathrm{guided}}^{(t)})$
\EndFor

\State $x_0 \gets \mathrm{Decode}(z)$
\State \Return $x_0$
\end{algorithmic}
\end{algorithm}
\vspace{-5pt}

\subsection{Backbone-Specific Feature Extraction}
Although the framework is unified in Algorithms~\ref{alg:stage1} and~\ref{alg:stage2}, the exact feature tensor saved at each step depends on the backbone.

For \textbf{FLUX.1-dev}, we save the latent update delta of the current step rather than an intermediate hidden state. The latent is first packed into a token sequence, which yields features of shape $[B,P,64]$. Therefore, the extracted feature corresponds to the update of the current step in packed latent-token space.

For \textbf{PixArt-Alpha}, we save the one-step solver update delta after classifier-free guidance. The latent map is patchified with patch size $2$, which yields tokenized features of shape $[B,P,16]$.

For \textbf{SD 3.5}, we likewise use the step-update delta after classifier-free guidance and sampler transformation. After patchifying the latent map with patch size $2$, the saved feature has shape $[B,P,64]$.

For \textbf{SDXL} and \textbf{SD 1.5}, we extract the DDIM step-update delta after classifier-free guidance. Since these backbones naturally produce latent updates of shape $(B,C,H,W)$, we patchify them into token sequences for SAE training and unpatchify them back into the original tensor form before writing the guided update to the sampler.

In all experiments reported in the main text, images are generated at resolution $256\times256$, so the number of tokens is consistently $P=256$ across all backbones. We then apply mean pooling over the token dimension whenever a pooled sample representation is required.

\subsection{Timestep-Wise SAE Bank}
We train one SAE for each timestep and do not share parameters across timesteps. This design follows from the fact that denoising features at different timesteps have different distributions and different semantic roles. Each timestep-specific SAE is trained only on normalized token features from that timestep, and each timestep stores its own normalization statistics $(\mu^{(t)},\sigma^{(t)})$.

Our SAE adopts a Top-$K$ sparse autoencoder of the form
\[
z=\mathrm{TopK}\!\bigl(W_{\mathrm{enc}}(x-b_{\mathrm{pre}})+b_{\mathrm{enc}}\bigr), \qquad
\hat x = W_{\mathrm{dec}} z + b_{\mathrm{pre}}.
\]
In practice, the decoder weights are unit-normalized, and an AuxK residual reconstruction term is added during training to alleviate dead features. The total loss therefore consists of the standard reconstruction loss and the AuxK residual reconstruction loss.

For the FLUX checkpoints used in the main experiments, we set $d_{\mathrm{model}}=64$, dictionary width $K_d=256$, $k=10$, $\mathrm{auxk}=128$, and the dead-feature threshold to $50$. For PixArt-Alpha, whose token dimension is $16$, we set $d_{\mathrm{model}}=16$, $K_d=64$, $k=10$, and $\mathrm{auxk}=32$. Unless otherwise specified, the remaining training hyperparameters are as follows: 200 epochs, batch size $40960$, learning rate $10^{-3}$, and AuxK coefficient $1/32$. We initialize the SAE pre-bias $b_{\mathrm{pre}}$ with the geometric median of a subset of normalized training tokens rather than with zero initialization.

\subsection{Intervention Granularity and Coefficient Sharing}
The interpolation in sparse space is applied to the \emph{entire} token-feature tensor rather than to selected spatial positions or selected latent dimensions. That is, at every guided timestep, we interpolate all tokens and all sparse features with the same elementwise linear rule in Eq.~\eqref{eq:interpolate}. The interpolation weight $\alpha$ remains fixed throughout sampling. In all main experiments, we use a constant $\alpha=0.8$, shared across timesteps, tokens, and latent dimensions.

\subsection{Guidance Window Across Backbones}
The main text states that intervention is restricted to the earliest portion of the denoising process. In implementation, this corresponds to the following backbone-specific step windows: steps 1--10 for FLUX.1-dev, SDXL, and SD1.5; steps 1--8 for SD3.5; and steps 1--5 for PixArt-Alpha. These windows approximately match the earliest $20\%$ of the denoising trajectory for each backbone.

\subsection{Reference Partition and Practical Label Construction}
For practical prototype construction, the success/failure partition used in the code is derived directly from the pooled SAE-space score above rather than from a separate VLM-based or human-annotation pipeline. Specifically, samples with $\mathrm{d}_i^{(t,c)}\le\tau$ are treated as reliable, reference-consistent samples, while those with $\mathrm{d}_i^{(t,c)}>\tau$ are treated as outliers or failed samples for prototype estimation. The same rule is used for both style and attribute settings.

\subsection{Multi-Condition Prompts}
The current implementation does not perform concept-level routing within the prompt. Instead, the target concept is specified externally by a single class identifier $c^\star$, and the corresponding prototype $R_{c^\star}^{(t)}$ is loaded and applied to the full token tensor. Therefore, when a prompt contains multiple conditions, the method does not parse which token span should align with which concept prototype. It uses one manually specified target class and applies the corresponding prototype uniformly to all token positions. This simplification is sufficient for the controlled setting studied in this work, but it is also an important limitation of the current implementation.

\section{Quantitative Metric Details}
\label{app:metric}

To make the quantitative results in the main text fully explicit, we describe the computation of all evaluation metrics used in this work. We group them into three categories: style-generation metrics, attribute-control metrics, and structural metrics used in the representation-layer analysis.

\subsection{Style Generation Metrics}

Suppose that the test set contains $N$ generated images. Let $x_i$ denote the generated image of sample $i$, let $p_i$ denote the full prompt, and let $c_i$ denote the target style. Let $f_{\mathrm{img}}(\cdot)$ and $f_{\mathrm{text}}(\cdot)$ denote the CLIP image encoder and text encoder, respectively. We use cosine similarity throughout:
\[
\mathrm{sim}(u,v)=\frac{u^\top v}{\|u\|_2\|v\|_2}.
\]

For each style $c$, we manually collect a small reference set $\mathcal{R}_c=\{r_1,\dots,r_M\}$ of visually representative images. CLIP-Image measures the average similarity between $x_i$ and the reference images of the target style:
\[
\mathrm{clip\mbox{-}I}(x_i,c_i)=\frac{1}{|\mathcal{R}_{c_i}|}\sum_{r\in\mathcal{R}_{c_i}}\mathrm{sim}\!\left(f_{\mathrm{img}}(x_i),f_{\mathrm{img}}(r)\right),
\]
and the final score is
\[
\mathrm{CLIP\mbox{-}Image}=\frac{1}{N}\sum_{i=1}^{N}\mathrm{clip\mbox{-}I}(x_i,c_i).
\]
This metric emphasizes whether the generated result approaches representative visual examples of the target style.

CLIP-Text measures alignment with the full prompt:
\[
\begin{aligned}
\mathrm{clip\mbox{-}T}(x_i,p_i)=\mathrm{sim}\!\left(f_{\mathrm{img}}(x_i),f_{\mathrm{text}}(p_i)\right),\\
\mathrm{CLIP\mbox{-}Text}=\frac{1}{N}\sum_{i=1}^{N}\mathrm{clip\mbox{-}T}(x_i,p_i).
\end{aligned}
\]
Because the full prompt contains both object content and condition cues, this metric reflects overall prompt adherence rather than style alone.

Style Alignment isolates the target style term and measures
\[
\begin{aligned}
\mathrm{SA}(x_i,c_i)=\mathrm{sim}\!\left(f_{\mathrm{img}}(x_i),f_{\mathrm{text}}(c_i)\right),\\
\mathrm{Style\ Alignment}=\frac{1}{N}\sum_{i=1}^{N}\mathrm{SA}(x_i,c_i).
\end{aligned}
\]
Compared with CLIP-Text, this metric more directly reflects whether the intended style is expressed.

Style-specific FID evaluates distributional proximity between generated samples and a style reference set. For each style $c$, let $\mathcal{G}_c$ denote the generated images under style $c$, and let $\mathcal{R}_c$ denote the corresponding reference set. After extracting Inception features, we estimate Gaussian statistics $(\mu_g,\Sigma_g)$ and $(\mu_r,\Sigma_r)$ for $\mathcal{G}_c$ and $\mathcal{R}_c$, and compute
\[
\mathrm{FID}(\mathcal{G}_c,\mathcal{R}_c)=\|\mu_g-\mu_r\|_2^2+\mathrm{Tr}\!\left(\Sigma_g+\Sigma_r-2(\Sigma_g\Sigma_r)^{1/2}\right).
\]
We report either the per-style values or their mean across styles. Since this metric measures proximity to a style domain rather than prompt-level concept realization, we use it as a secondary reference rather than as the main evidence in the paper.

\subsection{Attribute-Control Metrics}

In the attribute-control task, we evaluate whether color, texture, and shape are correctly bound to the target object. Let $a_i^{\mathrm{col}}$, $a_i^{\mathrm{tex}}$, $a_i^{\mathrm{sha}}$, and $o_i$ denote the target color, texture, shape, and object of sample $i$.

Following T2I-CompBench++~\cite{huang2025t2icompbenchpp}, we use BLIP-VQA to evaluate each attribute dimension with attribute-specific yes/no questions. For color, the question takes the form ``Is the \texttt{[object]} \texttt{[color]}?'' For texture, we ask ``Does the \texttt{[object]} have a \texttt{[texture]} texture?'' For shape, we ask ``Is the shape of the \texttt{[object]} \texttt{[shape]}?'' Although the wording is simple, it directly tests whether the target attribute is realized on the intended object. Let $\hat y_i^{\mathrm{attr}}\in\{0,1\}$ denote whether BLIP-VQA answers the corresponding question correctly, where $\mathrm{attr}\in\{\mathrm{Color},\mathrm{Texture},\mathrm{Shape}\}$. The score for each attribute is
\[
\mathrm{Score}_{\mathrm{attr}}=\frac{1}{N}\sum_{i=1}^{N}\mathbf{1}\!\left[\hat y_i^{\mathrm{attr}}=1\right].
\]
Accordingly, we report
\[
\begin{aligned}
\mathrm{Color}=\frac{1}{N}\sum_{i=1}^{N}\mathbf{1}\!\left[\hat y_i^{\mathrm{col}}=1\right],\\
\mathrm{Texture}=\frac{1}{N}\sum_{i=1}^{N}\mathbf{1}\!\left[\hat y_i^{\mathrm{tex}}=1\right],\\
\mathrm{Shape}=\frac{1}{N}\sum_{i=1}^{N}\mathbf{1}\!\left[\hat y_i^{\mathrm{sha}}=1\right].
\end{aligned}
\]
These three scores reflect attribute-binding accuracy along the three dimensions.

We also report CLIPSim as a continuous semantic-similarity metric between the generated image and an attribute-focused text description. Let $t_i^{\mathrm{attr}}$ denote the attribute text used for sample $i$, which can be either a single attribute token or a short descriptive phrase. We compute
\[
\mathrm{CLIPSim}(x_i,t_i^{\mathrm{attr}})=\mathrm{sim}\!\left(f_{\mathrm{img}}(x_i),f_{\mathrm{text}}(t_i^{\mathrm{attr}})\right),
\]
and average it over the test set:
\[
\mathrm{CLIPSim}=\frac{1}{N}\sum_{i=1}^{N}\mathrm{CLIPSim}(x_i,t_i^{\mathrm{attr}}).
\]
This metric complements BLIP-VQA by providing a continuous estimate of whether the target attribute is expressed.

\subsection{Representation-Layer Metrics}

For the representation-layer analysis, we ask whether prompt-level representations extracted from different layers are easier to classify and better separated in feature space. Let
\[
\mathcal{D}=\{(s_i,y_i)\}_{i=1}^{N}
\]
denote the pooled representation $s_i\in\mathbb{R}^{d}$ and its class label $y_i$.

Acc is computed by fitting a multiclass logistic regression classifier with $5$-fold cross-validation on $\mathcal{D}$. If $\mathcal{V}_m$ denotes the validation split of fold $m$ and $\hat y_i^{(m)}$ denotes the predicted label, then
\[
\mathrm{Acc}=\frac{1}{5}\sum_{m=1}^{5}\frac{1}{|\mathcal{V}_m|}\sum_{(s_i,y_i)\in\mathcal{V}_m}\mathbf{1}\!\left[\hat y_i^{(m)}=y_i\right].
\]
Higher Acc indicates that class information is more legible to a simple linear classifier.

Sep measures geometric class separation directly in representation space. We sample pairs of representations and compute Euclidean distances. The within-class and between-class distances are
\[
\begin{aligned}
\mathrm{Intra}=\mathbb{E}\!\left[\|s_i-s_j\|_2 \mid y_i=y_j\right],\\
\mathrm{Inter}=\mathbb{E}\!\left[\|s_i-s_j\|_2 \mid y_i\neq y_j\right].
\end{aligned}
\]
We then define
\[
\mathrm{Sep}=\frac{\mathrm{Inter}}{\mathrm{Intra}+\varepsilon},
\]
where $\varepsilon$ is a small constant for numerical stability. A larger Sep indicates that samples from different classes are farther apart while samples from the same class remain more compact.

Taken together, these metrics evaluate complementary aspects of the framework. For style generation, CLIP-Image emphasizes visual proximity to representative style examples, CLIP-Text evaluates global adherence to the full prompt, Style Alignment isolates target-style expression, and style-specific FID reflects distributional proximity to a style domain. For attribute control, BLIP-VQA provides explicit attribute-binding accuracy, while CLIPSim provides a continuous semantic measure. For the representation-layer study, Acc measures classifier readability and Sep measures intrinsic geometric separation. Their joint use provides a more complete view of where concept evidence becomes most legible within the diffusion model.

\begin{figure*}[!b]
\centering
\includegraphics[width=1.0\linewidth]{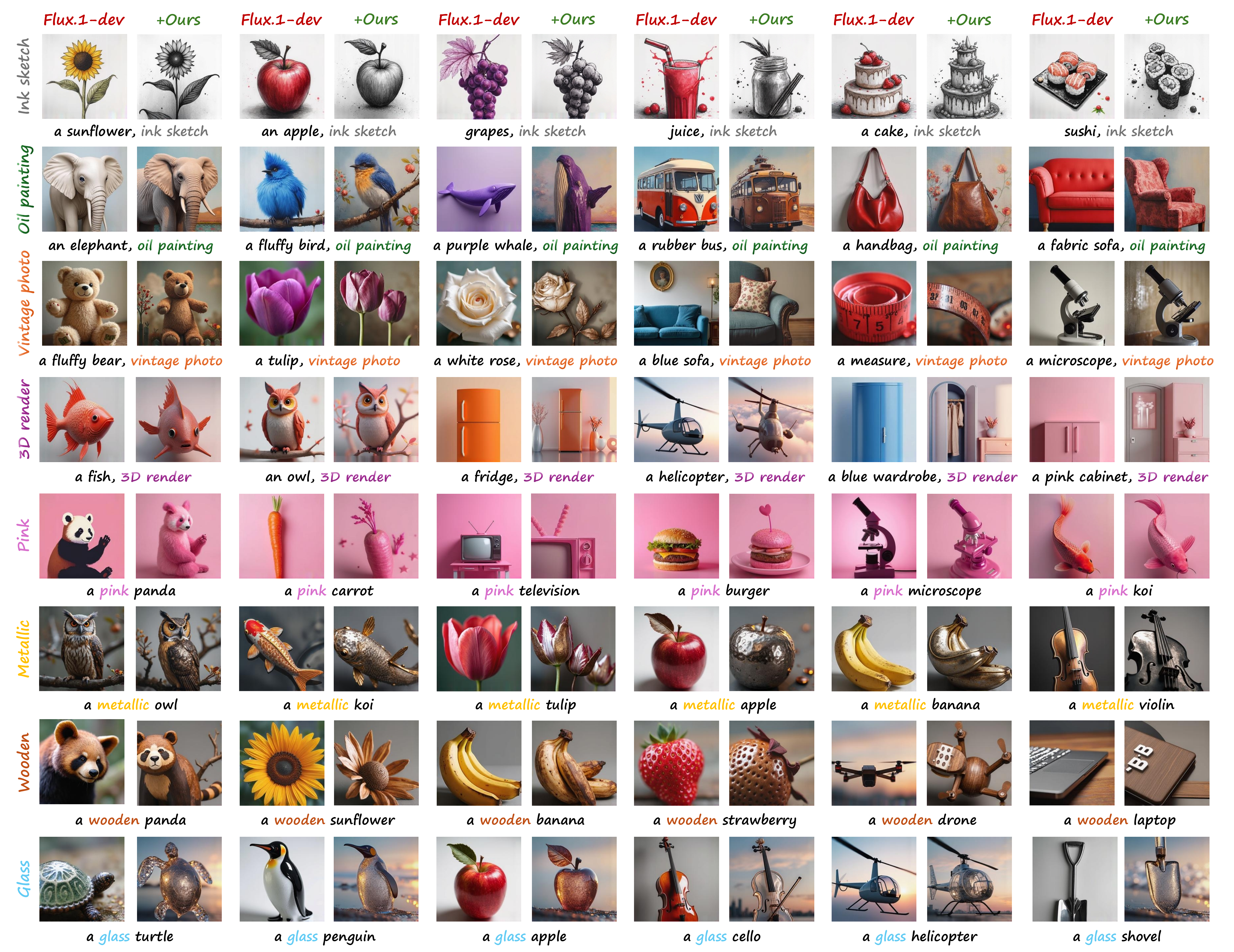}
\caption{Further qualitative comparisons for representative failure cases. The corrected results exhibit clearer target-style or target-attribute expression than the original outputs, which supports the claim that the sparse discrepancy identified in SAE space is functionally relevant to the final generation outcome.}
\label{fig:more1}
\end{figure*}
\vspace{-1em}
\section{More Qualitative Results}
\label{app:show}

\begin{figure*}[t]
\centering
\includegraphics[width=1.0\linewidth]{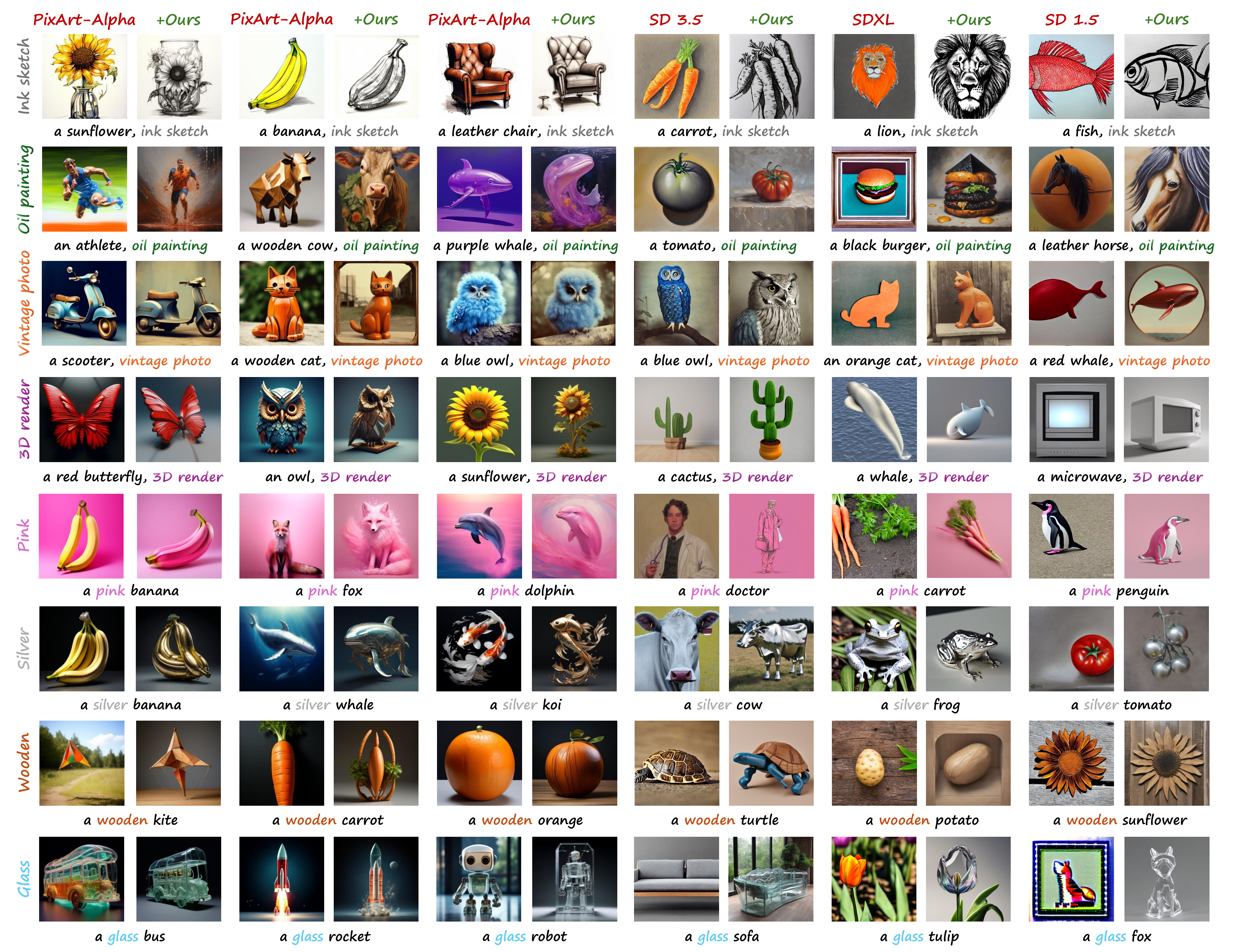}
\caption{Additional qualitative results on held-out prompts. Each group compares the original generation with the result after sparse-prior intervention. The examples show that the proposed method improves concept realization across diverse object categories and prompt types, while largely preserving the main content and layout of the original generation.}
\label{fig:more2}
\end{figure*}

We provide additional qualitative examples to complement the quantitative results in the main text. These examples are intended to illustrate three aspects of the proposed method more directly. First, the sparse-prior intervention improves concept realization across a broad range of objects rather than only on a few isolated prompts. Second, the improvement is observed for both global style conditions and more localized attribute conditions. Third, in most cases, the intervention strengthens the target concept while preserving the main object identity and overall scene structure.

Figures~\ref{fig:more1} and~\ref{fig:more2} present additional side-by-side comparisons between the original generation and the corrected result. Across these examples, the proposed method consistently reduces missing-style and weak-attribute failures that are difficult to repair from the final image alone. The examples also show that the effect is not limited to a single semantic category, but extends to animals, daily objects, vehicles, food items, and scene-related concepts.

Taken together, these examples provide qualitative support for the main claim of the paper. The proposed intervention does not merely produce isolated visual changes, but systematically restores missing or weakened concept evidence across a wide range of prompts. This behavior is consistent with our interpretation that \textbf{\textit{object-dependent concept brittleness arises from an internal mismatch in concept-related sparse activations rather than from purely random sampling variation.}}

\section{Intervening in the Raw Feature Space versus in SAE Space}
\label{app:check}

To clarify that the benefit of the proposed method arises from the intervention space itself rather than from merely inserting an additional interpolation step, we compare two intervention variants. In the first variant, we directly interpolate in the raw denoising feature space. In the second variant, we first map the current denoising feature into SAE space, interpolate it with the target class prototype in sparse space, and then decode it back into the original space for the subsequent sampling step. In this comparison, both variants use the same intervention strength $\alpha=0.8$, and all other sampling settings remain unchanged, so that the effect of the intervention space can be isolated as clearly as possible.

\begin{figure*}[t]
\centering
\includegraphics[width=0.8\linewidth]{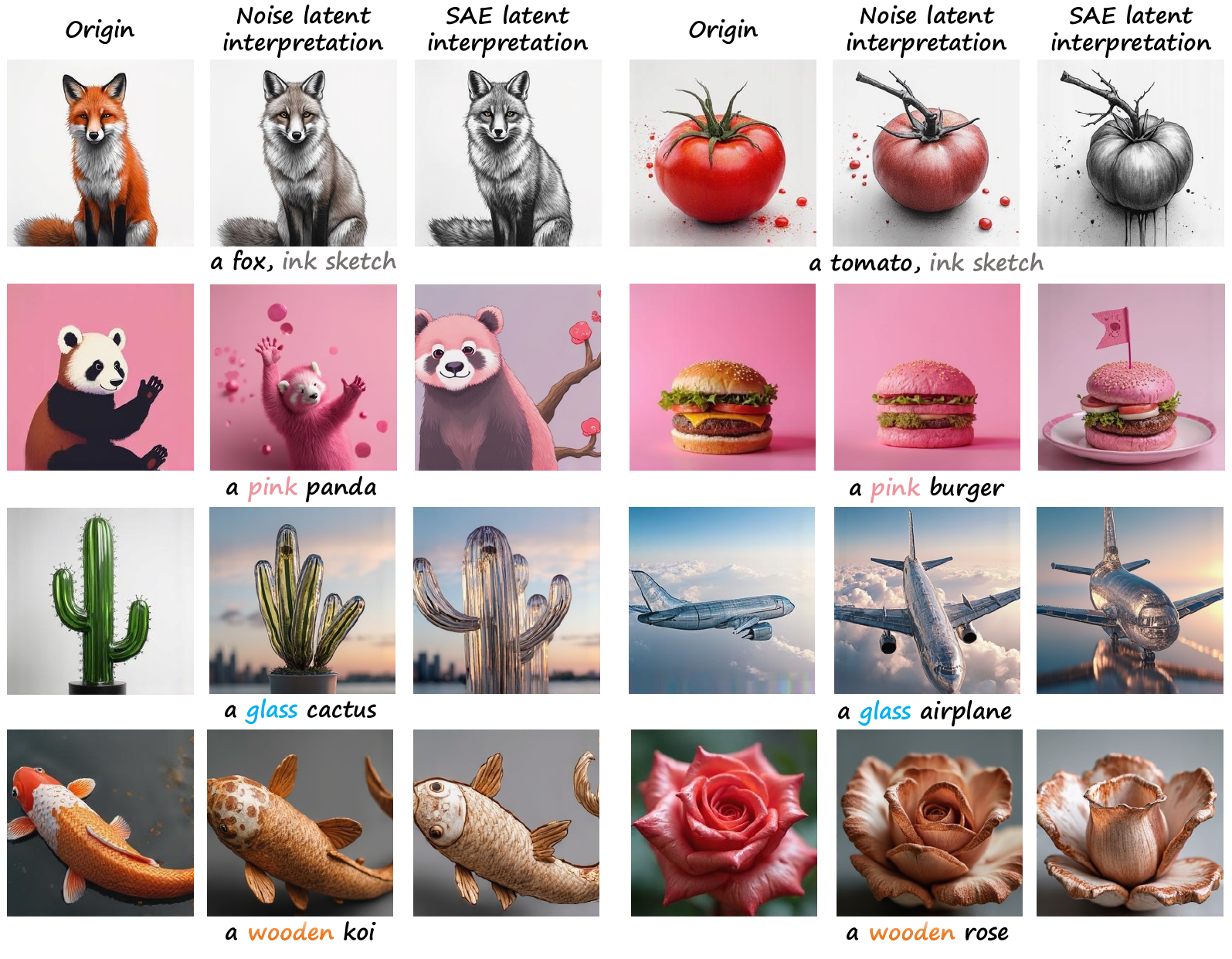}
\vspace{-8pt}
\caption{Comparison between intervention in the raw denoising feature space and intervention in SAE space. For each example, the three columns show the original generation, the result after raw-space intervention on the noise latent, and the result after SAE-space intervention. Raw-space intervention is generally more conservative and better preserves already-formed visual structure, but it often provides only partial correction of the target concept. By contrast, SAE-space intervention more reliably restores the intended style or attribute, which supports the view that concept repair is more effective when the intervention is performed in a sparse representation space where concept evidence is more separable.}
\label{fig:con}
\end{figure*}

Figure~\ref{fig:con} provides representative comparisons for both style and attribute failures. Each triplet shows the original generation, the result obtained by intervention in the raw noise-latent space, and the result obtained by intervention in SAE space. The examples cover several typical failure types, including missing sketch style in ``a fox, ink sketch'' and ``a tomato, ink sketch'', as well as attribute-binding failures such as ``a pink panda'', ``a pink burger'', ``a glass cactus'', ``a glass airplane'', ``a wooden koi'' and ``a wooden rose''. These examples are chosen to span different semantic categories and to test whether the two intervention spaces behave differently across both global style conditions and more localized attribute conditions.

The comparison reveals a consistent difference in behavior. Intervention in the raw denoising feature space is visually more conservative: it often preserves coarse object identity, background layout, and already-formed appearance cues, but its corrective effect is frequently limited. This can be observed, for example, in the sketch examples, where raw-space intervention moves the result toward a weaker grayscale rendering but often fails to impose a clear sketch-like appearance. A similar pattern appears in the oil-painting and 3D-render examples, where the raw-space result partially shifts the image toward the target concept yet still retains a substantial amount of the original appearance. This behavior is consistent with the fact that content, style, and attribute information remain highly entangled in the original feature space, so the intervention signal is diluted by the dominant content structure already present in the denoising trajectory.

By contrast, intervention in SAE space yields a substantially stronger and more stable correction effect. In Figure~\ref{fig:con}, the SAE-space result more reliably restores the intended sketch style for the fox and tomato. The same tendency is visible in the attribute examples, where the target color or material cue becomes more pronounced for the panda, burger, cactus, airplane, koi, and rose. In summary, SAE-space intervention can repair the missing or weakened concept evidence more consistently while still preserving the main semantic identity of the object. These observations support the main claim of the paper: \textit{\textbf{the effectiveness of the proposed intervention does not come simply from modifying the trajectory, but from performing that modification in a representation space where concept-relevant evidence is more explicitly separated from the surrounding content structure.}}

\section{Ablation Analyses}
\label{app:ablation}

\subsection{Sensitivity to the SAE Sparsity Hyperparameter $k$}
We examine the sensitivity of the SAE sparsity level $k$, which controls how many latent dimensions remain active after the Top-$K$ operator. Since the main purpose of the SAE in our framework is to expose concept-relevant structure more clearly than the raw feature space, we evaluate different values of $k$ through the two structural metrics used in the representation-layer analysis, namely Acc and Sep.

As shown in Table~\ref{tab:topk_sensitivity}, both Acc and Sep improve substantially when $k$ increases from a very small value, and then become relatively stable once the sparse code becomes moderately wide. This trend suggests that, beyond a certain point, the SAE already retains enough active dimensions to encode the concept information needed for our analysis. Although slightly larger values of $k$ may improve one metric in isolation, the gains are small and are no longer systematic. We therefore use $k=10$ in the main paper because it provides strong class structure while keeping the sparse representation compact and the subsequent intervention simple.

\begin{table}[!t]
\centering

\caption{Sensitivity to the SAE sparsity level $k$ in the representation-layer analysis.}
\vspace{-8pt}
\label{tab:topk_sensitivity}
\begin{tabular}{lccccc}
\toprule
$k$ & 4 & 6 & 10 & 15 & 20 \\
\midrule
Acc $\uparrow$ & 0.7269 & 0.7709 & 0.7648 & 0.7618 & 0.7800 \\
Sep $\uparrow$ & 1.7967 & 1.8094 & 1.8479 & 1.8796 & 1.8559 \\
\bottomrule
\end{tabular}
\end{table}

\begin{figure*}[!p]
\centering
\includegraphics[width=0.9\linewidth]{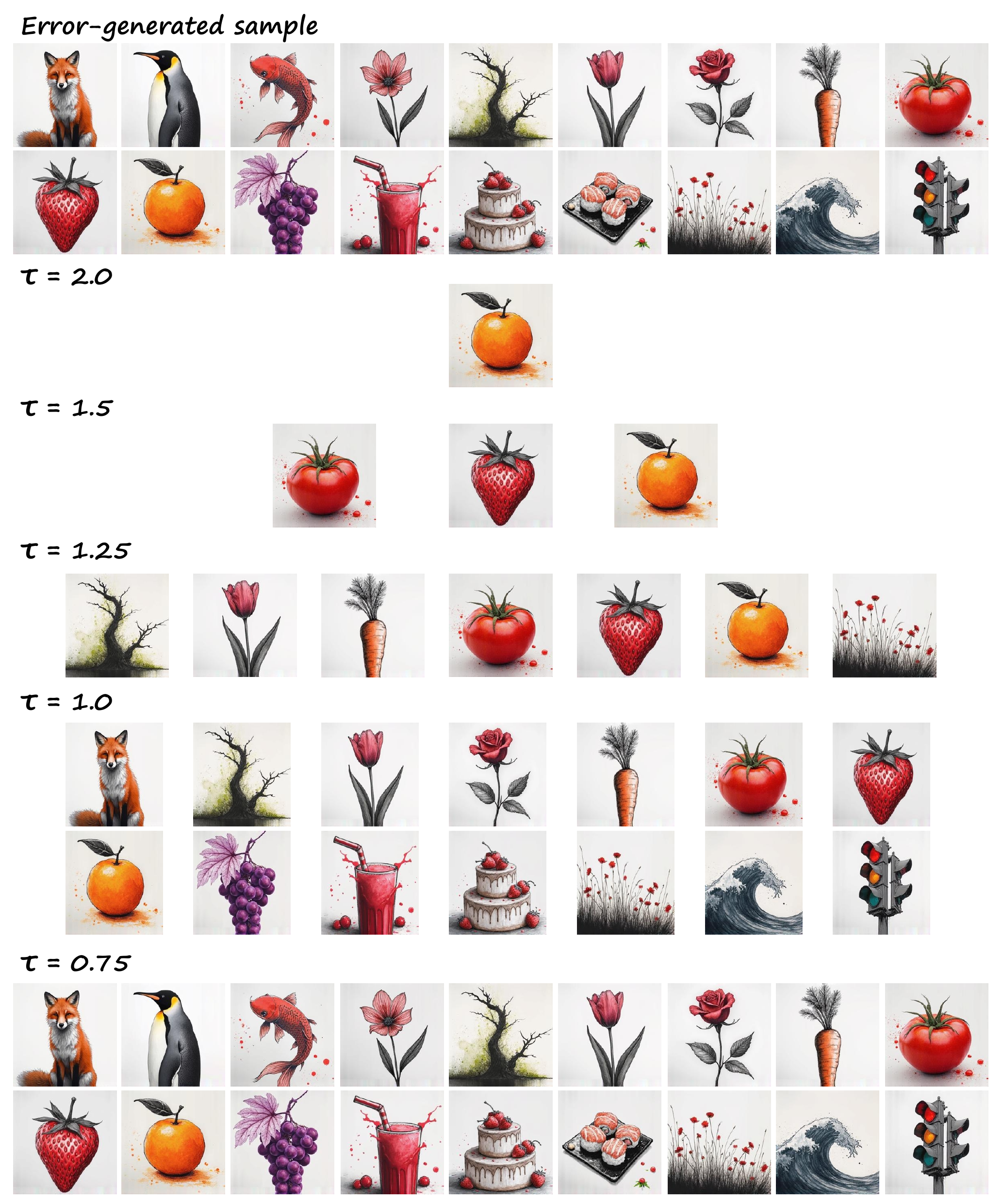}
\caption{Qualitative illustration of the filtering behavior under different thresholds $\bm{\tau}$ for the ink sketch style. The top row shows a manually collected set of visually obvious error-generated samples. The rows below show the subsets rejected by the pooled SAE-space score under different values of $\bm{\tau}$. As $\bm{\tau}$ decreases, the filtering becomes progressively stricter and captures a larger portion of the obvious failures. This visual trend is consistent with the quantitative statistics reported in the text: smaller thresholds improve prototype purity, but they also remove substantially more training samples and thus reduce the diversity of the retained reference subset.}
\label{fig:tau}
\end{figure*}
\subsection{Sensitivity to the Filtering Threshold $\tau$}

We further analyze the sensitivity of the filtering threshold $\tau$ in Eq.~\eqref{eq:tau}. In our framework, $\tau$ controls the strictness of the prototype-construction stage. A smaller value removes more samples that deviate from the class-consistent sparse pattern and therefore yields a cleaner but potentially less diverse reference subset. A larger value preserves more samples, but it may also retain unstable or weakly expressed cases. Since the training split is used to construct class-conditioned sparse priors, the choice of $\tau$ directly determines the trade-off between prototype reliability and intra-class diversity.

To examine this trade-off, we conduct a targeted analysis on the ink sketch style, where severe failure cases are visually clear and relatively easy to identify. We first manually collect a small set of obvious failure examples and use them only as a sanity-check set rather than as an additional supervision signal. We then vary $\tau$ and examine two quantities simultaneously: how many training samples are filtered out by Eq.~\eqref{eq:tau}, and whether the rejected subset overlaps with the manually identified severe failures. Figure~\ref{fig:tau} provides a visual illustration of this process. The top row shows the manually collected error-generated samples, while the rows below show the subsets rejected under different values of $\tau$. Although this figure is intended mainly as a qualitative illustration, it makes the filtering trend more transparent: as $\tau$ decreases, the rejected subset expands and progressively covers a larger portion of the visually obvious failures.

The resulting trend is monotonic. When $\tau$ is large, filtering is loose and removes only a few samples, but many visually obvious failures remain in the retained subset. As $\tau$ decreases, the filter becomes progressively stricter and captures a larger portion of the severe failures, but it also removes substantially more training data. Concretely, the number of excluded samples is $23/2080$ for $\tau=2.0$, $119/2080$ for $\tau=1.5$, $324/2080$ for $\tau=1.25$, $746/2080$ for $\tau=1.0$, and $1421/2080$ for $\tau=0.75$.

The qualitative evidence in Figure~\ref{fig:tau} is consistent with this quantitative trend. Under relatively loose thresholds such as $\tau=2.0$ and $\tau=1.5$, only a small portion of the obvious failures is identified. As the threshold becomes stricter, more of the error-generated samples are captured, and at $\tau=0.75$ the rejected subset already overlaps with nearly the full manually collected set. This behavior supports our interpretation that the pooled SAE-space score provides a meaningful proxy for sample reliability, even though the final choice of $\tau$ must still be determined by the balance between purity and diversity.

Among these settings, $\tau=0.75$ is the most aggressive and excludes nearly all manually identified severe failures. However, it also removes $1421$ out of $2080$ training samples, which is too destructive for the subsequent prototype-estimation stage. Such large-scale filtering substantially reduces the diversity of the retained reference pool and may weaken the robustness of the class-conditioned prior used for guidance. By contrast, $\tau=1.0$ already captures most of the manually identified severe failures while preserving a much larger fraction of the training data. We therefore treat $\tau=1.0$ as a more suitable operating point because it provides a better balance between removing unreliable samples and retaining sufficient intra-class diversity for sparse-prior construction.

Overall, this analysis suggests that the role of $\tau$ is not simply to maximize the number of discarded failure samples. Its real function is to balance prototype purity against prototype diversity. Based on this trade-off, we use $\tau=1.0$ as the default setting in all experiments.

\begin{figure*}[!htbp]
\centering
\includegraphics[width=0.8\linewidth]{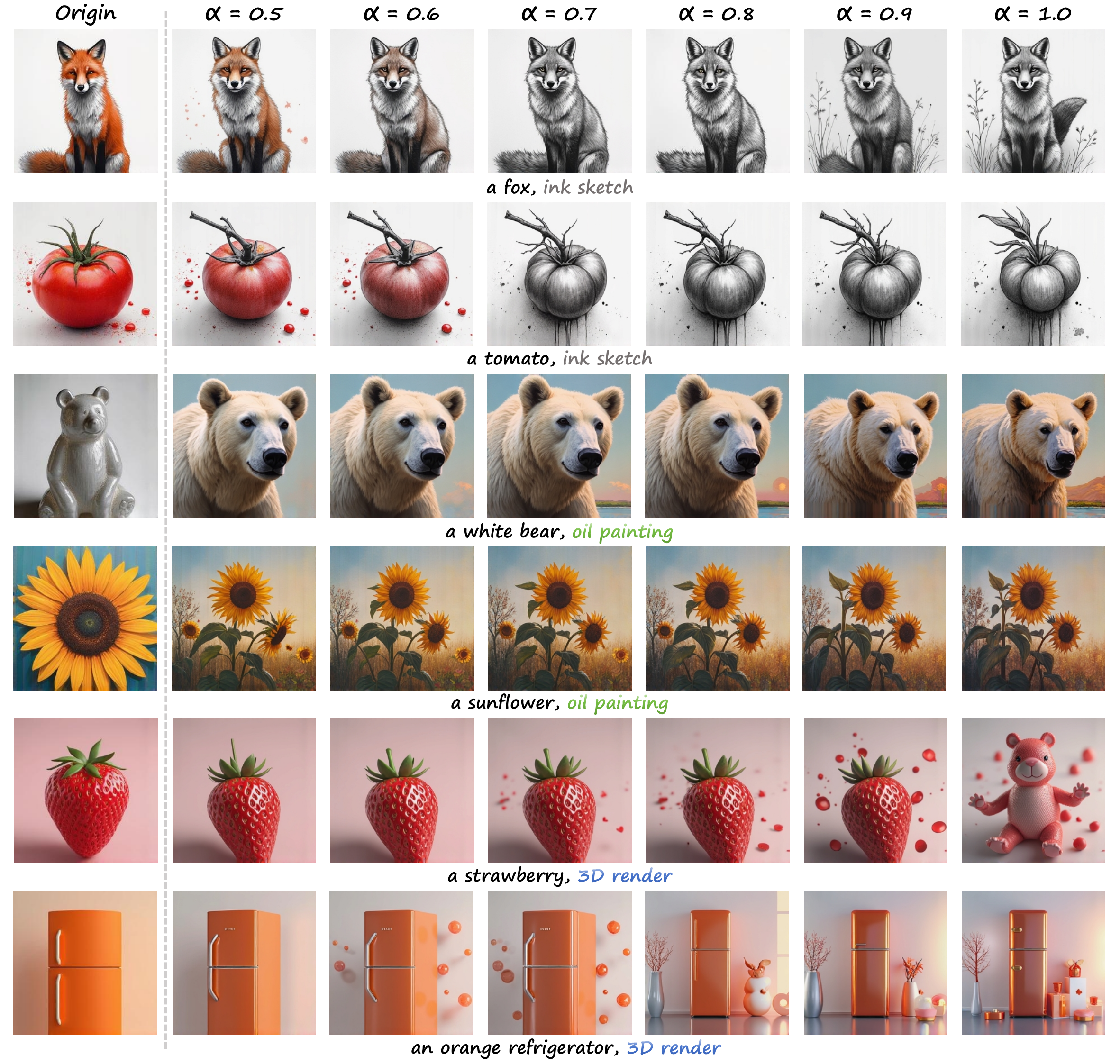}
\caption{Qualitative sensitivity to the intervention coefficient $\alpha$ on representative style-failure cases. From left to right, each row shows the original generation and the results obtained with $\bm{\alpha \in \{0.5,0.6,0.7,0.8,0.9,1.0\}}$. As $\bm{\alpha}$ increases, the target style is enforced more strongly: sketch, oil-painting, and 3D-render cues become progressively clearer. However, excessively large values may also introduce over-correction, including changes in scene layout, additional background structure, or semantic drift. Across these examples, moderate values around $\bm{\alpha=0.8}$ provide the most favorable balance between concept repair and content preservation.}
\label{fig:alpha1}
\end{figure*}
\begin{figure*}[!htbp]
\centering
\includegraphics[width=0.8\linewidth]{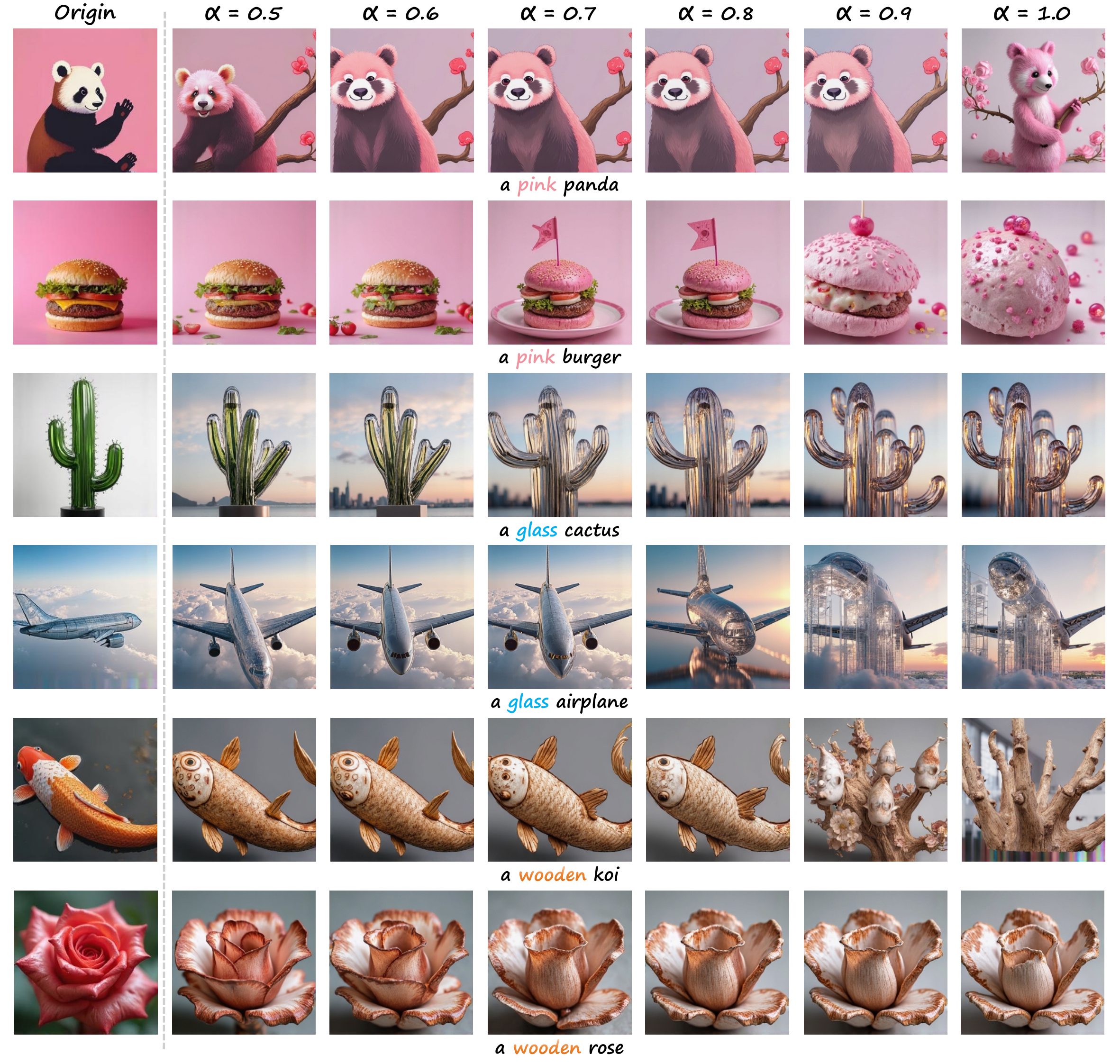}
\caption{Qualitative sensitivity to the intervention coefficient $\alpha$ on representative attribute-failure cases. From left to right, each row shows the original generation and the results obtained with $\bm{\alpha \in \{0.5,0.6,0.7,0.8,0.9,1.0\}}$. Increasing $\bm{\alpha}$ strengthens the target attribute, including color and material cues such as pink, glass, and wooden. At the same time, overly strong intervention may distort object identity or produce visually implausible structures, especially at $\bm{\alpha=1.0}$. These examples further support the conclusion that moderate intervention strengths, particularly around $\bm{\alpha=0.8}$, achieve a better trade-off between reliable concept correction and preservation of the original object content.}
\label{fig:alpha2}
\end{figure*}
\subsection{Sensitivity to the Intervention Coefficient $\alpha$}

We further analyze the sensitivity of the intervention coefficient $\alpha$ in Eq.~\eqref{eq:interpolate}. In our framework, $\alpha$ controls the interpolation strength between the current sparse code of the test sample and the target class prototype. A smaller $\alpha$ changes the original denoising trajectory more conservatively and is therefore more favorable for preserving the object structure and content details that have already formed. However, such weak guidance may not be sufficient to compensate for missing, weakened, or shifted concept evidence in failed samples. A larger $\alpha$ pushes the current denoising update more strongly toward the class prototype and therefore enforces the target style or attribute more aggressively. Yet if the intervention is too strong, it may overwrite content information that has already formed in the original trajectory and lead to over-correction, including changes in object appearance, abnormal local structure, or even semantic drift. The choice of $\alpha$ therefore controls the trade-off between concept-repair strength and content-preservation ability.

To examine this trade-off, we qualitatively compare representative failed prompts under $\alpha\in\{1.0,0.9,0.8,0.7,0.6,0.5\}$ while keeping all other sampling settings unchanged. We focus on two questions. The first is whether a given value of $\alpha$ can effectively repair the original concept failure. The second is whether the repaired result still preserves the original object identity and overall content structure. Figures~\ref{fig:alpha1} and~\ref{fig:alpha2} summarize the corresponding qualitative comparisons for style failures and attribute failures, respectively.

The qualitative results show a clear pattern. When $\alpha$ is too small, the intervention signal is weak and often produces only limited concept correction, so some originally failed samples remain insufficiently repaired. This effect is particularly visible in the style examples of Figure~\ref{fig:alpha1}. For prompts such as ``a fox, ink sketch'' and ``a tomato, ink sketch,'' smaller values such as $\alpha=0.5$ or $\alpha=0.6$ only partially suppress the original color appearance and do not yet yield a sufficiently clear sketch-like rendering. A similar tendency appears for ``a white bear, oil painting'' and ``a sunflower, oil painting,'' where weak intervention improves the result only marginally and does not fully establish the painterly texture and atmosphere associated with the target style.

As $\alpha$ increases, expression of the target style or attribute becomes more apparent. In Figure~\ref{fig:alpha1}, values around $\alpha=0.7$ and $\alpha=0.8$ generally provide a favorable balance: the sketch examples become substantially more consistent with the target monochrome line-based style, the oil-painting examples acquire a more coherent painterly appearance, and the 3D-render examples show a clearer synthetic rendered look. The same trend is observed in the attribute examples of Figure~\ref{fig:alpha2}. For ``a pink panda'' and ``a pink burger,'' increasing $\alpha$ gradually strengthens the target color cue, while for ``a glass cactus'' and ``a glass airplane,'' the intended material property becomes progressively more explicit. Likewise, the wooden texture in ``a wooden koi'' and ``a wooden rose'' becomes clearer as the intervention strength increases from weak to moderate values.

However, overly large interpolation also makes the result more likely to deviate from the original content. This effect becomes evident at $\alpha=0.9$ and is most pronounced at $\alpha=1.0$. In Figure~\ref{fig:alpha1}, aggressive intervention may introduce substantial scene changes or semantic drift. For example, the 3D-render case of ``a strawberry, 3D render'' is over-corrected at $\alpha=1.0$ and no longer preserves the original object identity, while ``an orange refrigerator, 3D render'' acquires additional scene elements and a noticeably altered composition under stronger intervention. In Figure~\ref{fig:alpha2}, the same phenomenon is visible in cases such as ``a pink panda,'' ``a pink burger,'' and especially ``a glass airplane'' and ``a wooden koi,'' where the target attribute is strengthened but the object itself becomes increasingly distorted or partially replaced by visually implausible structures. In other words, too small a value of $\alpha$ fails to repair the concept reliably, whereas too large a value of $\alpha$ tends to repair it too aggressively.

Taken together, these observations suggest that $\alpha=0.8$ is a suitable operating point. Relative to smaller values, it provides sufficiently strong concept guidance and repairs a substantial portion of style and attribute failures. Relative to more aggressive values, it is less likely to damage object content or disrupt the original layout. We therefore use $\alpha=0.8$ as the default setting in all main experiments.

\section{Robustness Analyses}
\label{app:robust}

\subsection{Robustness Across Random Seeds}

The main text reports quantitative results under a fixed test seed of $2026$. To assess whether the observed improvement is specific to a particular noise initialization, we further evaluate FLUX.1-dev on the style-generation task under four different random seeds, namely $2026$, $512$, $331$, and $42$. The original model and our method are compared under exactly the same experimental settings.

Table~\ref{tab:seed_robustness} shows that the proposed intervention yields a consistent improvement over the original model across all tested seeds. More importantly, this consistency is observed simultaneously on CLIP-Image, CLIP-Text, and Style Alignment, which indicates that the gain is not restricted to a single aspect of generation quality. The relative ranking between the original model and our method remains unchanged under different noise initializations, and the improvement margins are also stable across seeds.

\begin{table}[t]
\centering
\caption{Robustness across random seeds on the FLUX.1-dev style-generation benchmark.}
\label{tab:seed_robustness}
\resizebox{\columnwidth}{!}{
\begin{tabular}{lcccccccc}
\toprule
& \multicolumn{2}{c}{2026} & \multicolumn{2}{c}{512} & \multicolumn{2}{c}{331} & \multicolumn{2}{c}{42} \\
\cmidrule(lr){2-3}\cmidrule(lr){4-5}\cmidrule(lr){6-7}\cmidrule(lr){8-9}
Metric & Origin & Ours & Origin & Ours & Origin & Ours & Origin & Ours \\
\midrule
CLIP-Image & 0.598 & 0.623 & 0.577 & 0.618 & 0.578 & 0.618 & 0.582 & 0.622 \\
CLIP-Text & 0.273 & 0.288 & 0.275 & 0.288 & 0.277 & 0.293 & 0.274 & 0.290 \\
Style Alignment & 0.220 & 0.230 & 0.219 & 0.230 & 0.221 & 0.231 & 0.219 & 0.230 \\
\bottomrule
\end{tabular}}
\end{table}

These results support two conclusions. First, the benefit of the proposed sparse-prior intervention does not depend on a particular favorable sampling trajectory. Second, the intervention improves style realization in a manner that is robust to seed variation, rather than merely shifting performance under isolated cases. From the perspective of interpretability, this stability is important because it suggests that the repaired concept evidence reflects a persistent property of the denoising process, instead of an incidental effect tied to one specific random initialization.

Overall, the multi-seed evaluation strengthens the main claim of the paper. The proposed method provides a reproducible improvement in style-related generation quality, and its effect remains stable under different sampling seeds.

\subsection{Step-wise SAE Activation Comparison Across Guided Timesteps}

The main text visualizes the difference between successful and failed samples in SAE space using a single representative timestep. Since our framework trains a separate SAE for each denoising step, it is also important to examine whether the same contrast remains visible throughout the full guided window rather than only at one isolated step. To this end, we further visualize the top-10 most different latent dimensions for four representative prompts over the first ten denoising steps, including two successful cases, namely ``a dog, ink sketch'' and ``a car, ink sketch,'' and two failed cases, namely ``an orange, ink sketch'' and ``juice, ink sketch.''

\begin{figure*}[t]
\centering
\includegraphics[width=0.75\linewidth]{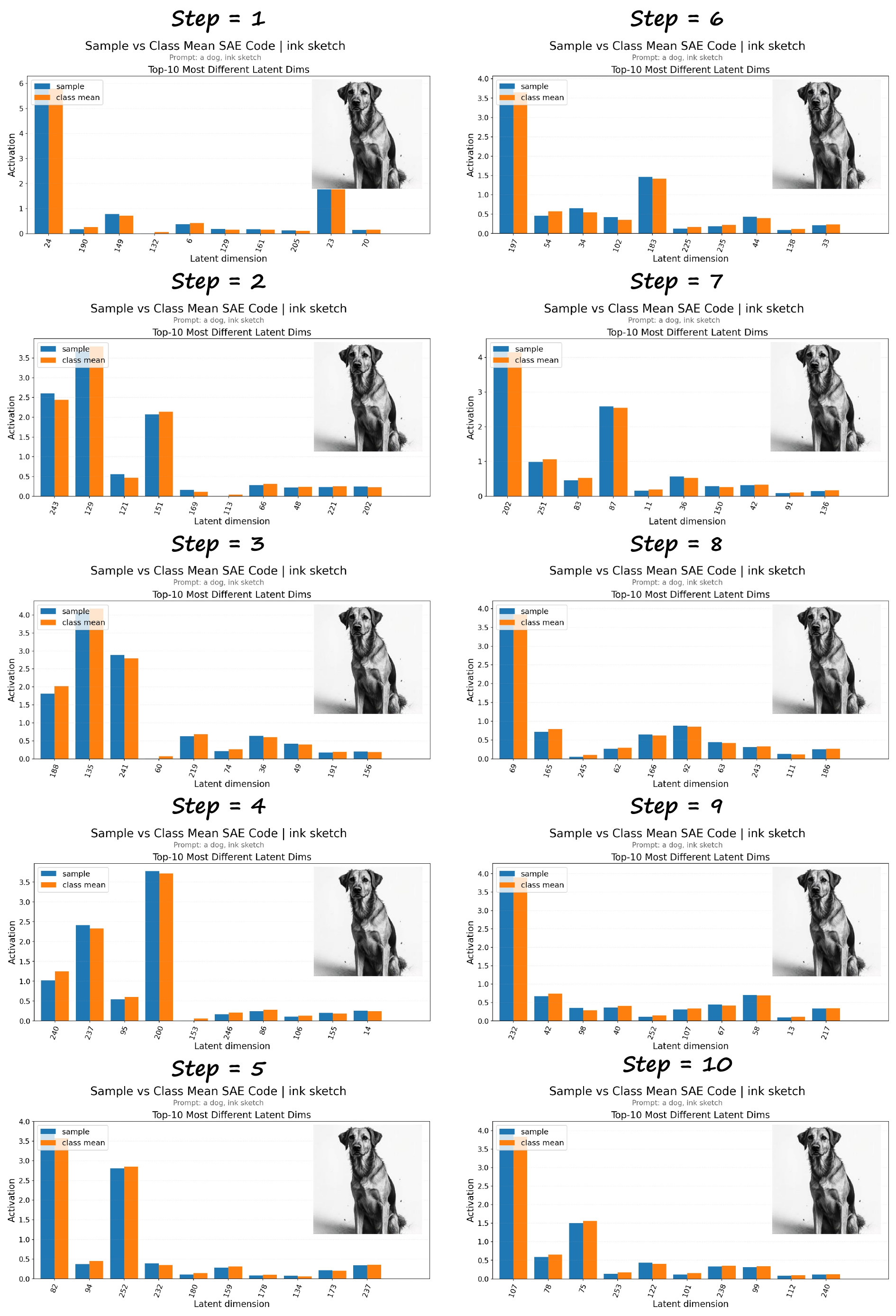}
\caption{Step-wise comparison between the sample SAE code and the class-mean SAE code for the successful prompt ``a dog, ink sketch'' over the first ten denoising steps. Across most steps, the dominant activation peaks of the sample remain closely aligned with the class mean, and the overall sparse activation pattern is stable over time. This example illustrates that successful generations follow a condition-consistent trajectory in step-wise SAE space rather than matching the target concept only at the final image level.}
\label{fig:sae1}
\end{figure*}

\begin{figure*}[t]
\centering
\includegraphics[width=0.75\linewidth]{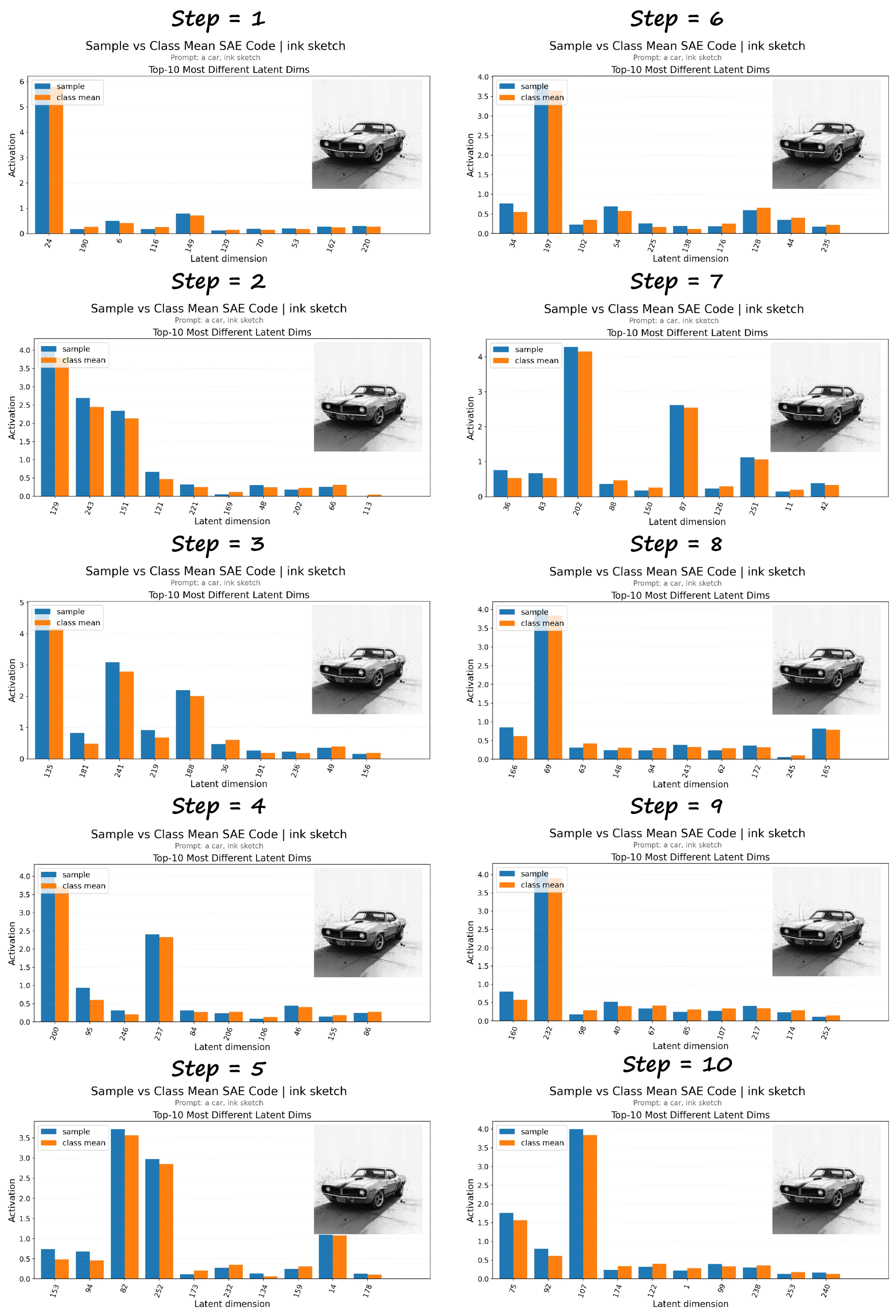}
\caption{Step-wise comparison between the sample SAE code and the class-mean SAE code for the successful prompt ``a car, ink sketch'' over the first ten denoising steps. Similar to Fig.~\ref{fig:sae1}, the sample activations remain broadly consistent with the class mean across the guided window, with only limited local fluctuations. This supports the view that successful concept realization corresponds to a temporally stable sparse activation structure.}
\label{fig:sae2}
\end{figure*}

\begin{figure*}[t]
\centering
\includegraphics[width=0.75\linewidth]{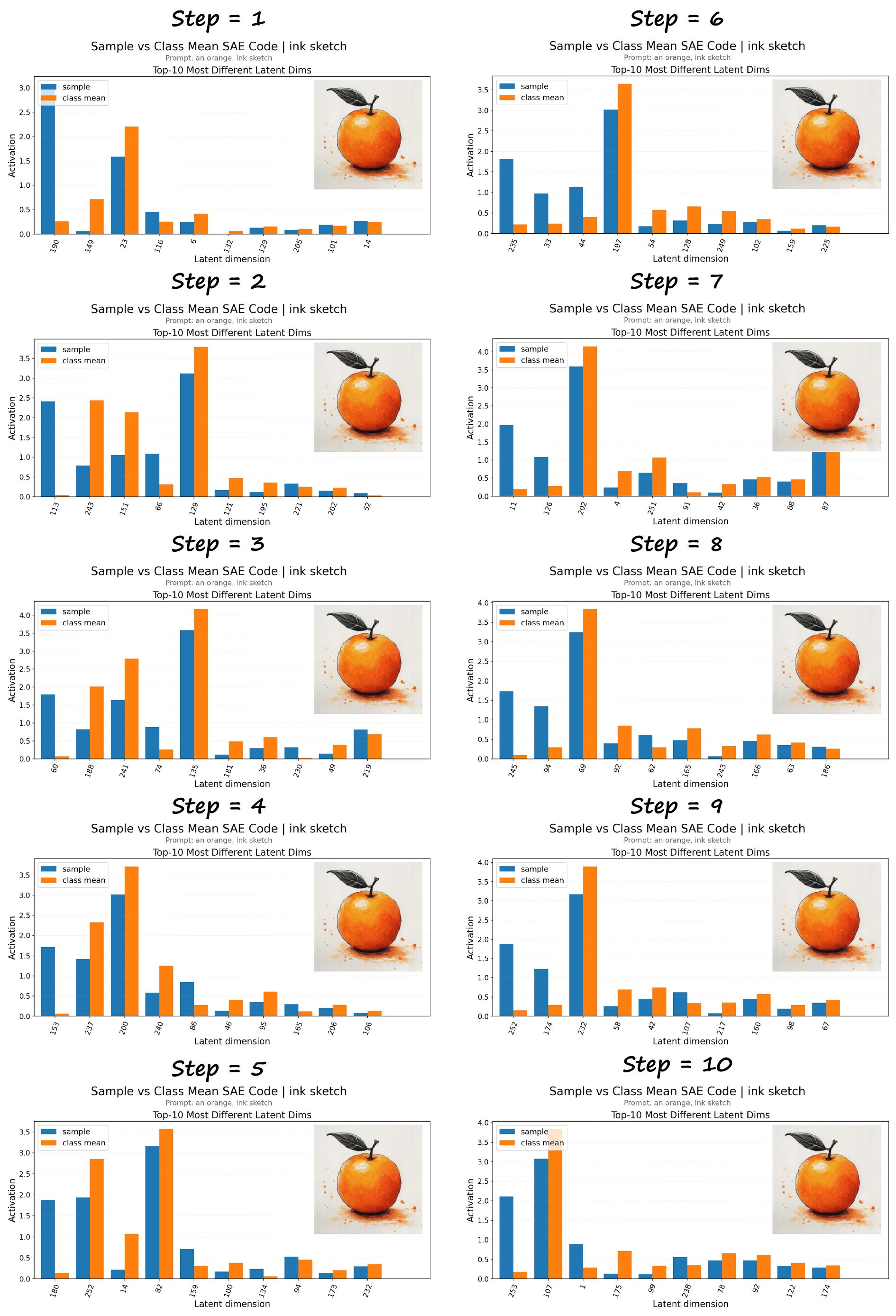}
\caption{Step-wise comparison between the sample SAE code and the class-mean SAE code for the failed prompt ``an orange, ink sketch'' over the first ten denoising steps. In contrast to successful samples, the activation pattern shows repeated mismatches with the class mean across multiple steps, especially on a small subset of dominant sparse dimensions. The discrepancy is already visible in the early denoising phase and persists throughout the guided window, indicating a trajectory-level failure of concept realization.}
\label{fig:sae3}
\end{figure*}

\begin{figure*}[t]
\centering
\includegraphics[width=0.75\linewidth]{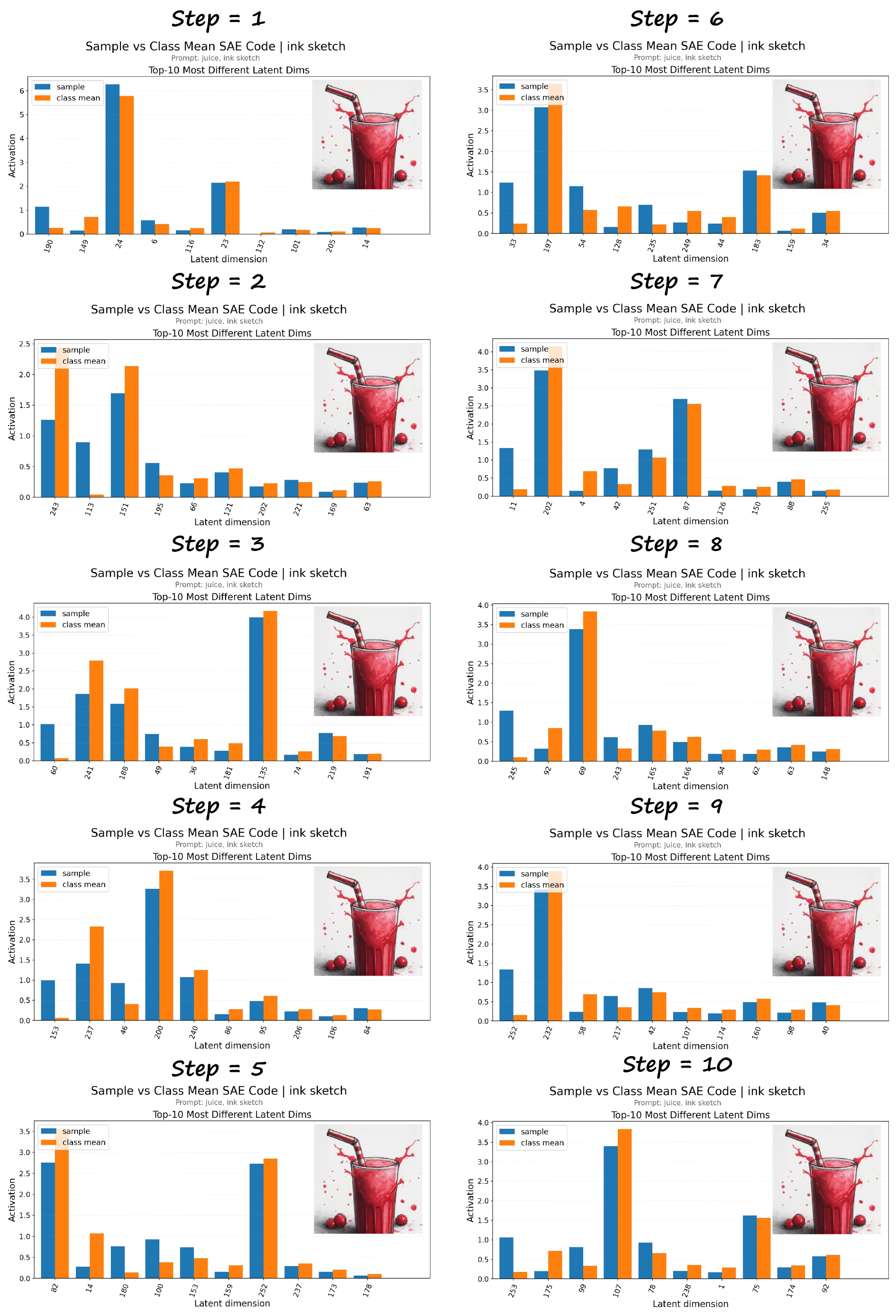}
\caption{Step-wise comparison between the sample SAE code and the class-mean SAE code for the failed prompt ``juice, ink sketch'' over the first ten denoising steps. The sample exhibits persistent deviations from the class-consistent sparse profile, with larger gaps on several highly activated latent dimensions than those observed in successful cases. This figure further suggests that concept brittleness is associated with a structured and temporally persistent mismatch in SAE space rather than with an isolated late-stage perturbation.}
\label{fig:sae4}
\end{figure*}

Figures~\ref{fig:sae1}--\ref{fig:sae4} reveal a clear temporal distinction between successful and failed trajectories. For the successful samples, the sample-wise SAE activations remain close to the corresponding class mean across most of the first ten steps. Although small fluctuations are still present, the dominant activation peaks generally appear on similar latent dimensions and with similar magnitudes. This indicates that successful generations do not merely satisfy the target concept at the final image level, but also follow a relatively stable condition-consistent trajectory in step-wise sparse space.

By contrast, the failed samples exhibit larger and more persistent deviations from the class mean across multiple denoising steps. The mismatch is not uniformly distributed across the full latent code. Instead, it is concentrated on a limited subset of highly activated sparse dimensions, which is consistent with the main-text observation that concept brittleness appears as a structured sparse discrepancy rather than as diffuse noise. In particular, for both ``an orange, ink sketch'' and ``juice, ink sketch,'' noticeable gaps between the sample activation and the class mean are already visible at the earliest steps and remain observable throughout much of the guided window. This suggests that these failures are not caused by a purely late-stage drift, but reflect a trajectory-level mismatch that emerges early and persists over time.

Another notable pattern is that the identity of the top deviating latent dimensions changes from step to step. This behavior further justifies the timestep-specific SAE design adopted in our framework. Since denoising features play different semantic roles at different stages, a shared latent space would blur this temporal structure, whereas step-wise SAE representations preserve the local statistics of each stage and make cross-sample comparison more faithful. The present visualizations therefore provide additional qualitative support for the use of a timestep-specific sparse representation.



\end{document}